\documentclass{article} 
\usepackage{iclr2016_conference,times}

\usepackage{amsmath}
\usepackage{amssymb}
\usepackage{xspace}
\usepackage{multirow}
\usepackage{array}
\usepackage{tikz}
\usepackage{multirow}
\usepackage{lettrine}
\usepackage{hyperref}

\begin{document}

\title{Particular object retrieval with integral max-pooling of CNN activations}

\author{Giorgos Tolias \thanks{Research partially conducted while G. Tolias and H. J\'egou were at Inria. We would like to thank Florent Perronnin for his valuable feedback. This work was partly supported by MSMT LL1303 ERC-CZ grant.}
\\
Center for Machine Perception\\
FEE CTU Prague\\
\And
Ronan Sicre\\
Irisa Rennes \\
\And
Herv\'e J\'egou\\
Facebook AI Research \\
}

\iclrfinalcopy 

\maketitle
\def\gfv{MAC\xspace}
\def\rfv{R-MAC\xspace}
\def\deeploc{AML\xspace}

\def\ie{\emph{i.e.}~}
\def\eg{\emph{e.g.}~}
\def\wrt{\emph{w.r.t.}~}
\def\etal{\emph{et al.}~}

\newcommand{\real}{\mathbb{R}}
\newcommand{\realnn}{{\mathbb{R}^{+}_{0}}}
\newcommand{\nat}{\mathbb{N}}
\newcommand{\natzero}{{\mathbb{N}_{0}}}

\newcommand{\cX}{\mathcal{X}}
\newcommand{\cY}{\mathcal{Y}}
\newcommand{\cB}{\mathcal{B}}
\newcommand{\cR}{\mathcal{R}}

\newcommand{\bB}{\mathbf{B}}

\newcommand{\vf}{\mathbf{f}}
\newcommand{\f}{\mathrm{f}}
\newcommand{\vg}{\mathbf{g}}
\newcommand{\vq}{\mathbf{q}}
\newcommand{\tvf}{\tilde{\mathbf{f}}}
\newcommand{\tf}{\tilde{\mathrm{f}}}

\def\l2{$\ell_2$}

\renewcommand{\paragraph}[1]{{\medskip \noindent \bf #1}}
\newcommand{\pari}[1]{{\medskip \noindent \it #1}}
\newcommand{\equ}[1]{Equation~(\ref{#1})\xspace}

\newcommand{\comm}[1]{#1}
\newcommand{\alert}[1]{#1}

\def\sssp{\hspace{1pt}}
\def\ssp{\hspace{3pt}}
\def\msp{\hspace{5pt}}
\def\bsp{\hspace{8pt}}

\def\reg{\mathcal R}

\vspace{-1ex}
\begin{abstract}
Recently, image representation built upon Convolutional Neural Network (CNN) has been shown to provide effective descriptors for image search, outperforming pre-CNN features as short-vector representations.
Yet such models are not compatible with geometry-aware re-ranking methods and still outperformed, on some particular object retrieval benchmarks, by traditional image search systems relying on precise descriptor matching, geometric re-ranking, or query expansion.
This work revisits both retrieval stages, namely initial search and re-ranking, by employing the same primitive information derived from the CNN.
We build compact feature vectors that encode several image regions without the need to feed multiple inputs to the network.
Furthermore, we extend integral images to handle max-pooling on convolutional layer activations, allowing us to efficiently localize matching objects.
The resulting bounding box is finally used for image re-ranking.
As a result, this paper significantly improves existing CNN-based recognition pipeline: We report for the first time results competing with traditional methods on the challenging Oxford5k and Paris6k datasets.
\end{abstract}

\section{Introduction}
\label{sec:intro}
\lettrine{C}{ontent} based image retrieval has received a sustained attention over the last decade, leading to mature systems for tasks like visual instance retrieval.
Current state-of-the-art approaches are derived from the Bag-of-Words model of~\cite{SZ03} and mainly owe their success to locally invariant features~\citep{L04} and large visual codebooks~\citep{PCISZ07}.
These methods are typically composed of an initial \emph{filtering} stage where all database images are ranked in terms of similarity to a query image and a second \emph{re-ranking} stage, which refines the search results of the top-ranked elements.
The filtering stage is improved in several ways, such as incorporating weak geometric information~\citep{JDS10a}, employing compact approximations of the local descriptors~\citep{JDS10a}, or learning smart codebooks~\citep{MPCM13,AK12}.
In such cases, local descriptors are individually matched and selective matching functions~\citep{TAJ15,TGSS14} improve the search quality.
Geometric matching models~\citep{PCISZ07,AT14} are typically applied in a pairwise manner during the re-ranking stage of a short-list of images.
Query expansion approaches significantly increase the performance~\citep{CMPM11}, at the cost of larger query times.

The recent advances achieved by Convolutional Neural Networks (CNN) and the use of intermediate layer activations as feature vectors~\citep{DJVO+13} create opportunities for representations that are competitive for  image or particular object retrieval, and not only classification tasks.
Several works have already investigated this research direction, such as global or local representations based on either fully connected~\citep{BSCL14,GWGL14} or convolutional layers~\citep{RSMC14,ARSM+14,BL15}.
The performance of CNN-based features has rapidly improved to the point of competing and even outperforming pre-CNN works that aggregate local features~\citep{JPDSPS11,RJC15}.
In particular, activations of convolutional layers followed by a global max-pooling operation~\citep{ARSM+14} produce highly competitive compact image representations.
One limitation is that such approaches are not compatible with the geometric-aware models involved in the final re-ranking stages.

\begin{figure}[t]
\vspace{-10pt}
\centering
 \raisebox{1ex}{\includegraphics[width=0.21\textwidth]{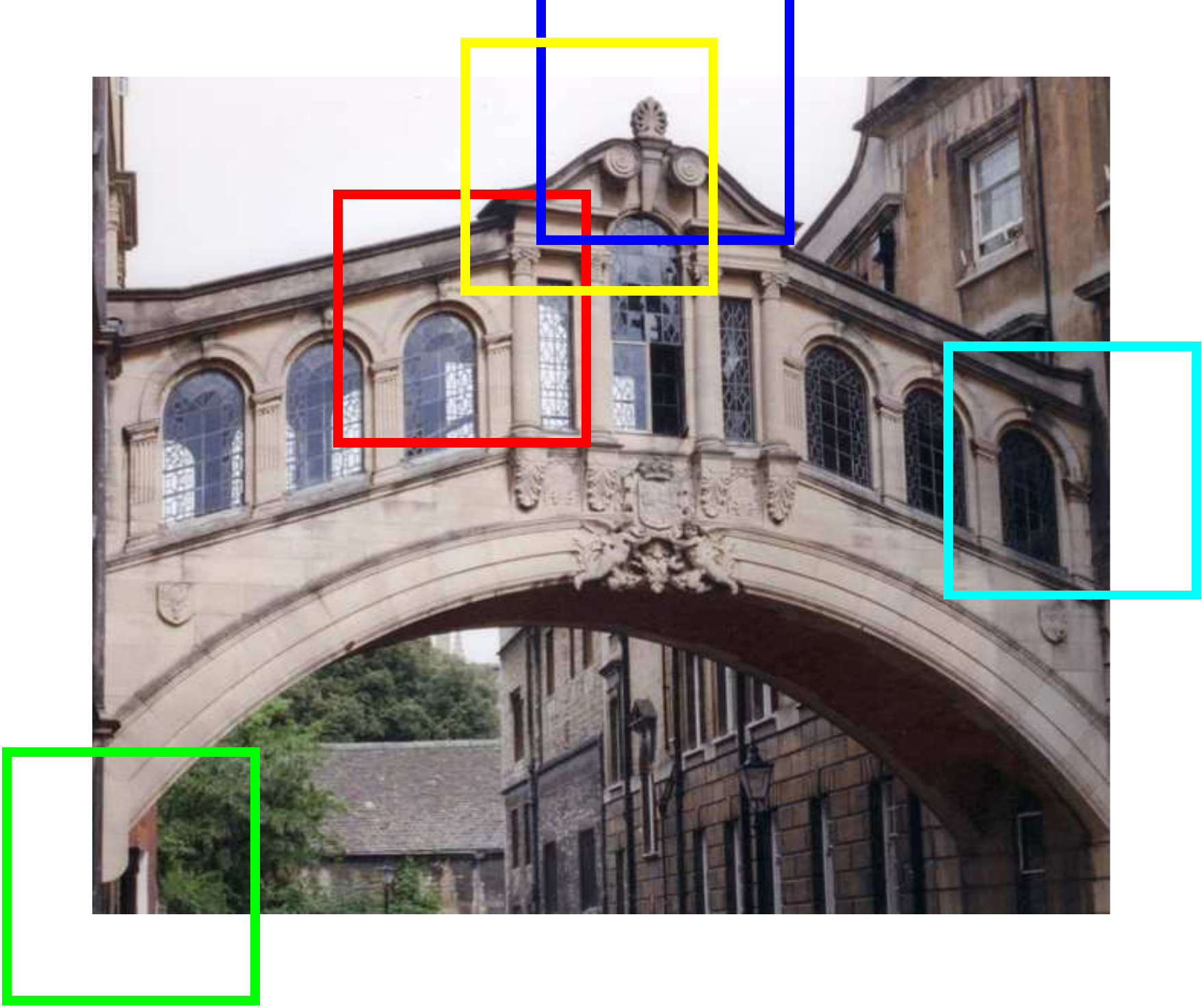}} \hfill
\includegraphics[height=0.17\textwidth, trim=0 0 0 50]{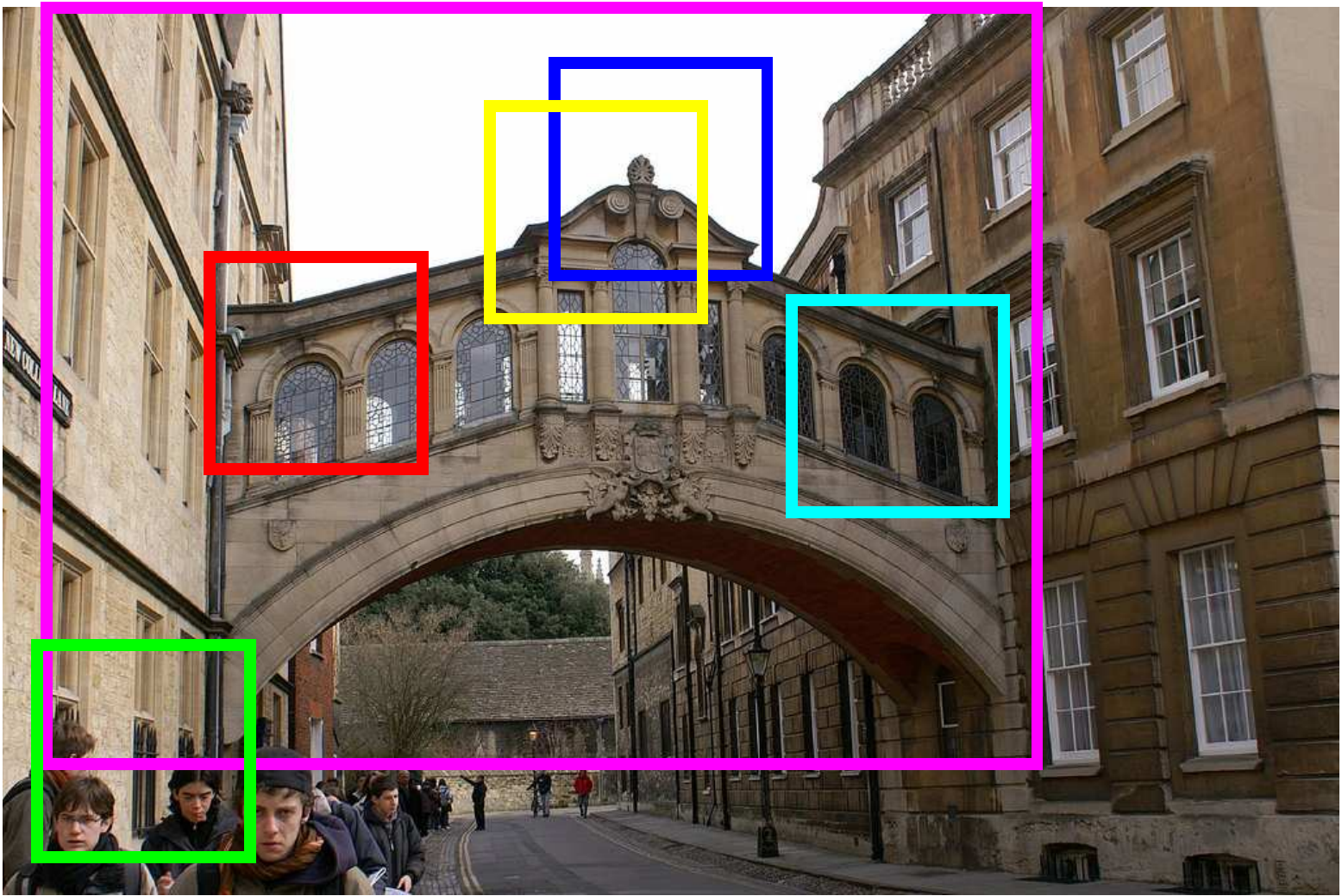}
\hfill
 \raisebox{0ex}{\includegraphics[width=0.19\textwidth]{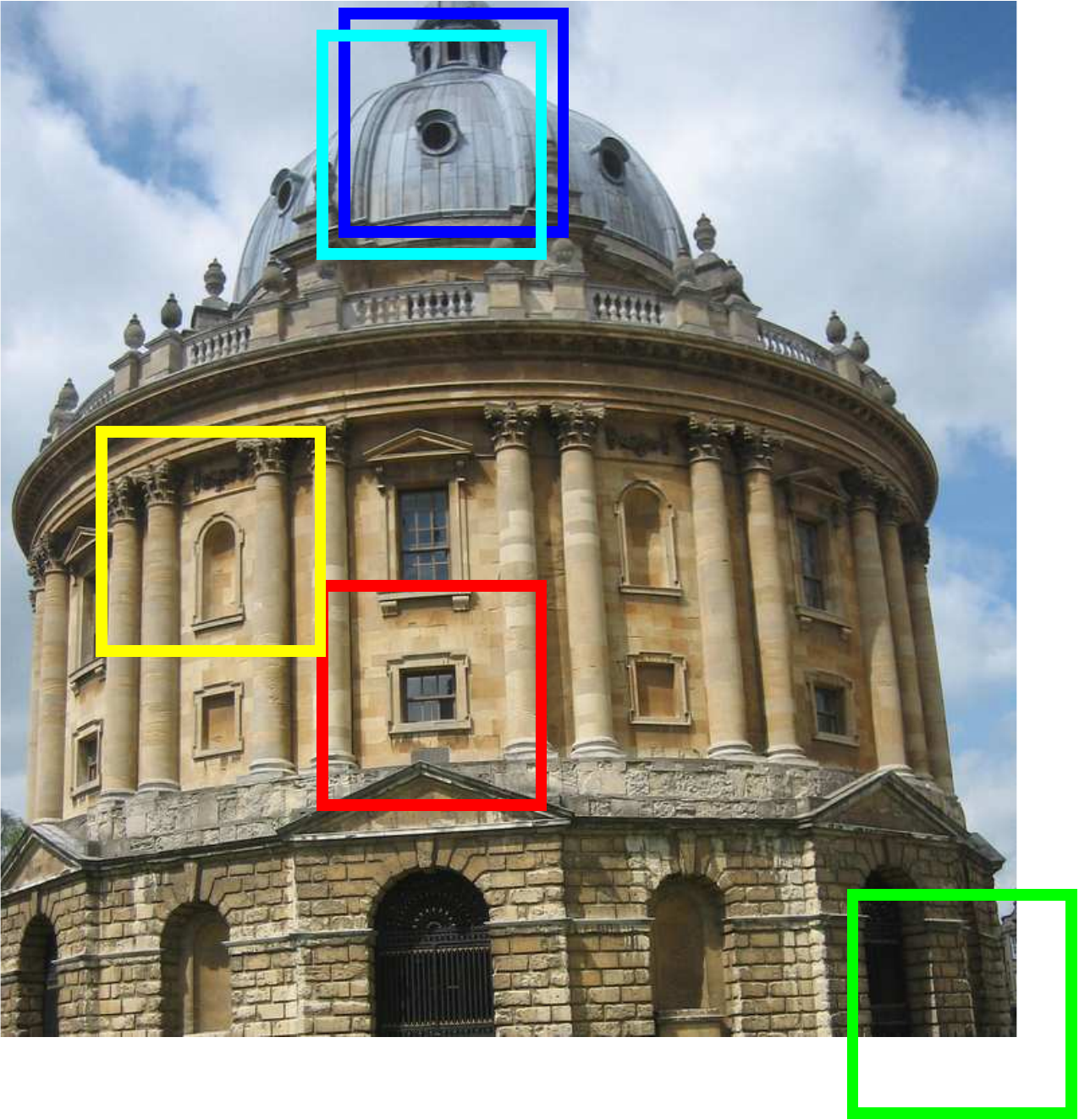}} \hfill
\includegraphics[height=0.18\textwidth, trim=0 0 0 50]{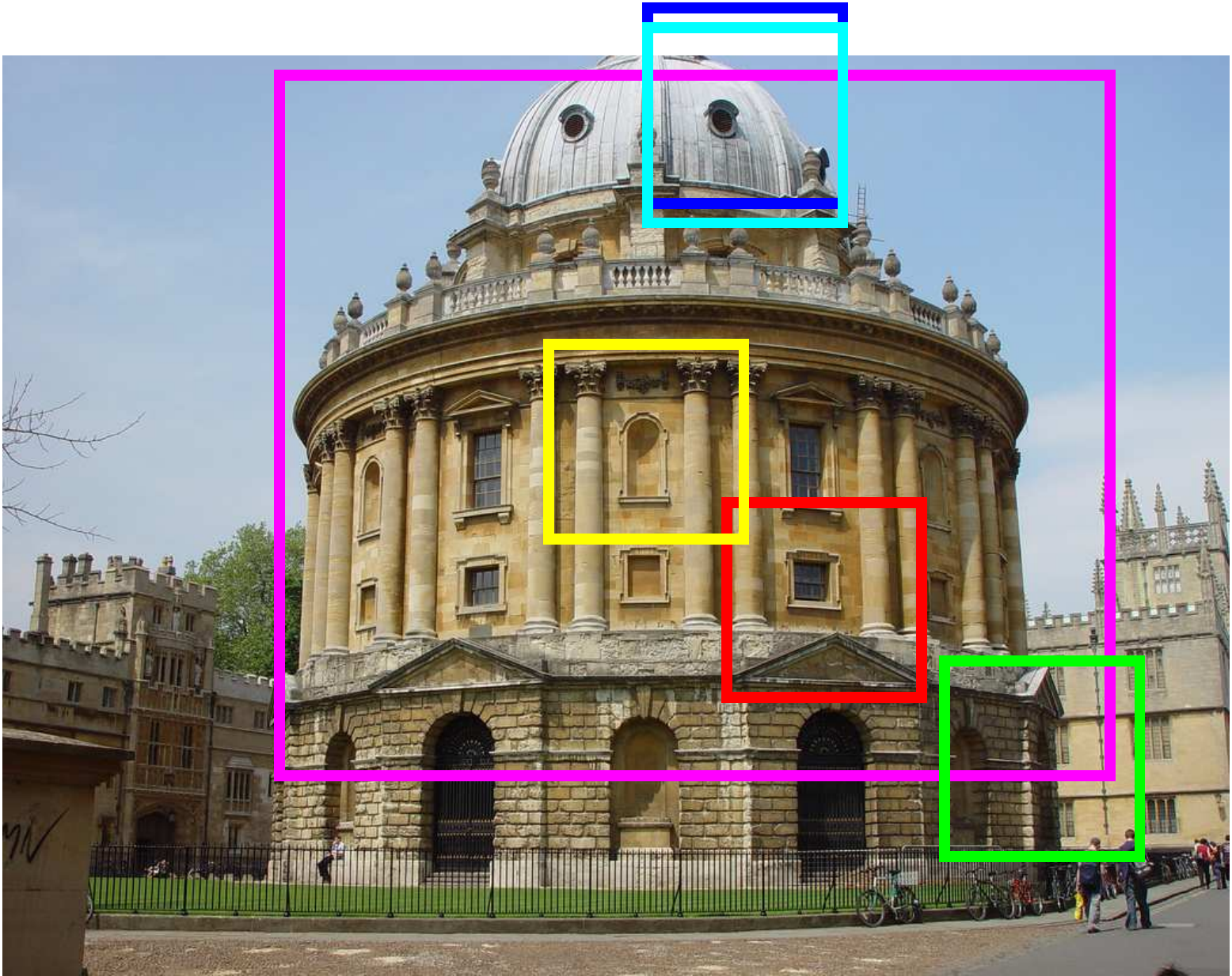}
 \vspace{-8pt}
\caption{Query objects (left) and the corresponding localization in another image (right) are shown.
We visualize the patches that contribute the highest to the image similarity score.
Displayed patches correspond to the receptive field of CNN activations.
Object localization is displayed in magenta, while different colors are used for patches in correspondence.
\label{fig:intro}}
\vspace{-10pt}
\end{figure}

This work revisits both filtering and re-ranking stages with CNN-based features. We make the three following contributions.
\begin{itemize}
\item First, we propose a compact image representation derived from the convolutional layer activations that encodes multiple image regions without the need to re-feed multiple inputs to the network, in  spirit of recent Fast-RCNN~\citep{Girshick15} and Faster-RCNN~\citep{RHGS15} methods but here targeting particular object retrieval.
The underlying primitive representation is used in all stages (initial retrieval and re-ranking).
\item Second, we employ the generalized mean~\citep{DTPB09} to enable the use of integral images along with max-pooling. This efficient method
is exploited for particular object localization (see Figure~\ref{fig:intro}) directly in the 2D maps of CNN activations.
\item Third, our localization approach is used for image re-ranking and leads us to define a simple yet effective query expansion method.
\end{itemize}
These approaches are complementary and, when combined, produce for the first time a system which compete on the Oxford and Paris building benchmarks with state-of-the-art re-ranking approaches based on local features. Our approach outperforms by a large margin previous methods based on CNN, while being more efficient in practice.

\vspace{-4pt}
\section{Related work}
\vspace{-4pt}
\label{sec:related}
\textbf{CNN based representation.}
A typical CNN consists of several convolutional layers, followed by fully connected layers and ends with a softmax layer producing a distribution over the training classes.
Instead of using this inherent classifier, one can consider the activations of the intermediate layers to train a classifier.
In particular, the activations of the fully connected layers have been shown to be very effective and capable of adaptation to various domains~\citep{OBLS14}, such as scene recognition~\citep{DJVO+13,SJ15}, object detection~\citep{IMKG+14}, and semantic segmentation~\citep{GDDM14}.
In the case of image retrieval, fully connected layers are used as global descriptors followed by dimensionality reduction~\citep{BSCL14}.
They are also employed as region descriptors to be compared to database descriptors~\citep{RASC14} or aggregated in a VLAD manner~\citep{GWGL14}. 

Recent works derive visual representations from the activations of the convolutional layers. This is achieved either by stacking activations~\citep{GDDM14} or by performing spatial max-pooling~\citep{ARSM+14} or sum-pooling~\citep{BL15} for each feature channel.
According to~\cite{ARSM+14} such representation offers better generalization properties for test data that are far from the source (training) data.
Noticeably, higher performance in particular object or scene retrieval is obtained by using convolutional layers rather than fully connected ones.
The very recent work of~\cite{BL15} shows that sum-pooling performs better than max-pooling when the image representation is whitened.
In addition to be a costly choice, we will show that this is not optimal in our context of object localization (see Section~\ref{sec:experiments}). Finally, \cite{KMO15} propose spatial and feature channel weighting that significantly improves performance. Their approach is complementary to what we propose for the filtering and the re-ranking stage.

Recent examples utilize information from fully connected layers to perform generic object detection~\citep{IMKG+14,PKS14}.
Such approaches are prohibitive for the re-ranking purposes of large scale image retrieval. They have high computational cost and the inherent features are not optimal for particular object matching.

\textbf{Localization.}
In the recent years, the sliding window principle has been quite successful for many object localization methods.
Due to the large number of possible windows, exhaustive search is extremely costly.
However, integral images~\citep{VJ01} offer a constant cost solution to the evaluation of a single region.
This attractive alternative is applicable for feature vectors constructed via a sum-pooling operation.

A globally optimal solution is given by Efficient Subwindow Search (ESS) of~\cite{LBH09}, who use branch-and-bound search to avoid exhaustive search.
Their work employs integral images, which are also used in later improvements of ESS~\citep{APLV09}. 
\cite{APLV09} formalize localization as a maximum sub-array problem and similarly to~\cite{CSFD+13} they employ Bentley's algorithm~\citep{bentley99}.
Integral images facilitate the evaluation of many region candidates~\citep{UVGS13} based on VLAD or Fisher vectors~\citep{VSS14}.
All aforementioned approaches take advantage of integral images due to the inherent sum-pooling operation in the given representation.
In this paper, we extend integral images to perform max-pooling over CNN activation maps, which is shown to be a better choice for describing regions (as opposed to the entire image).

Several object localization techniques have been proposed in the context of image retrieval as well. ~\cite{Lam09} propose a two layer branch-and-bound method that alternates between regions and images.
Integral images offer a significant speed-up in the work of~\cite{LB10} to  perform localization through Bag-of-Words.
The overall idea bears similarities with our work. However, we differentiate by employing CNN-based representation with max-pooling.
Some approaches~\citep{TGSS14,SLBW14} individually index local features for localization.
In our case, the localization method is built on top of a compact representation, initially used for the filtering stage.
Finally,~\cite{AZ13} propose a localization strategy based on VLAD, where similarity is computed for multiple image regions, giving a more precise localization via regression.

\section{Background}
\vspace{4pt}
\label{sec:background}
\begin{figure*}[t]
\centering
\includegraphics[height=0.17\textwidth]{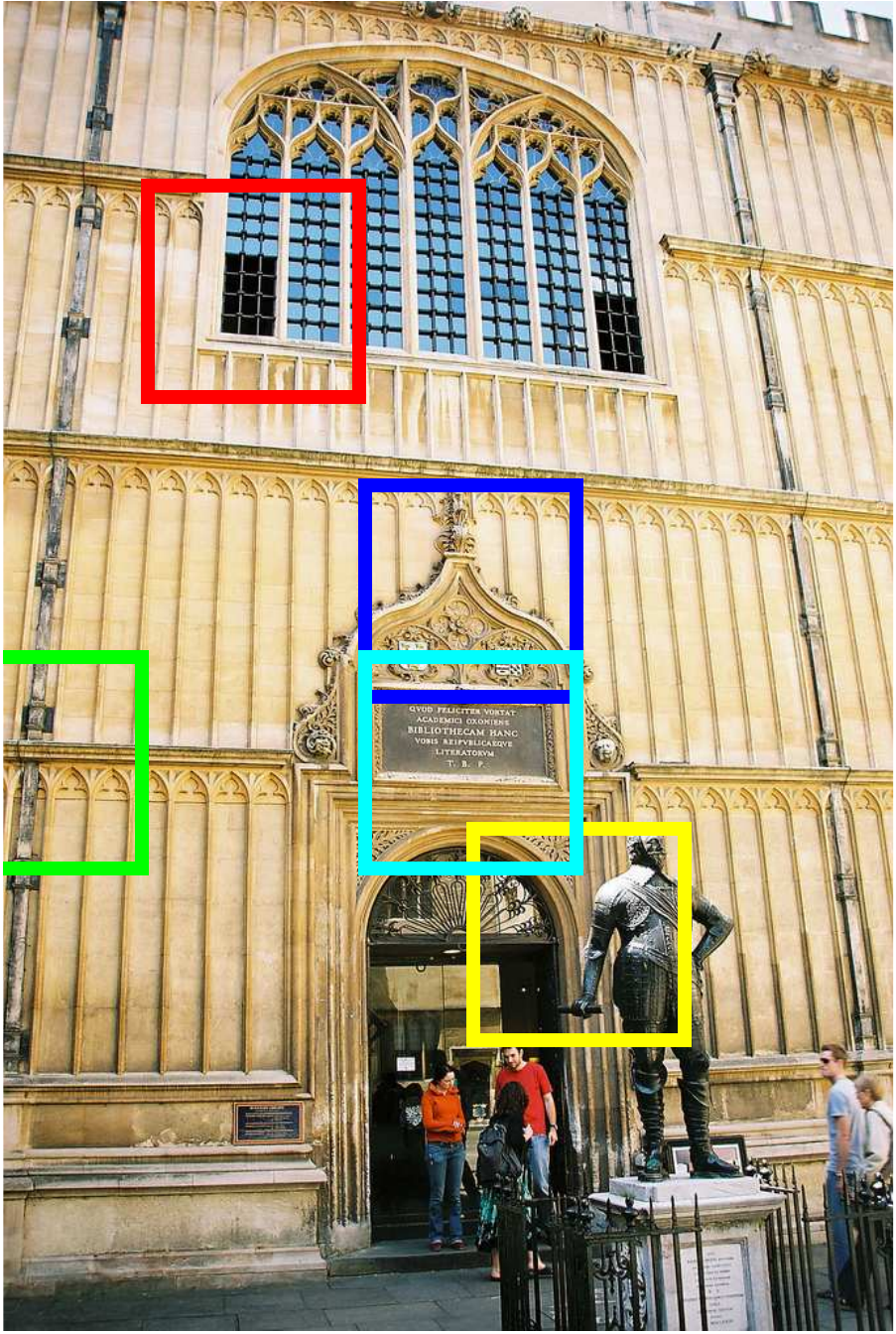}  \
\includegraphics[height=0.17\textwidth]{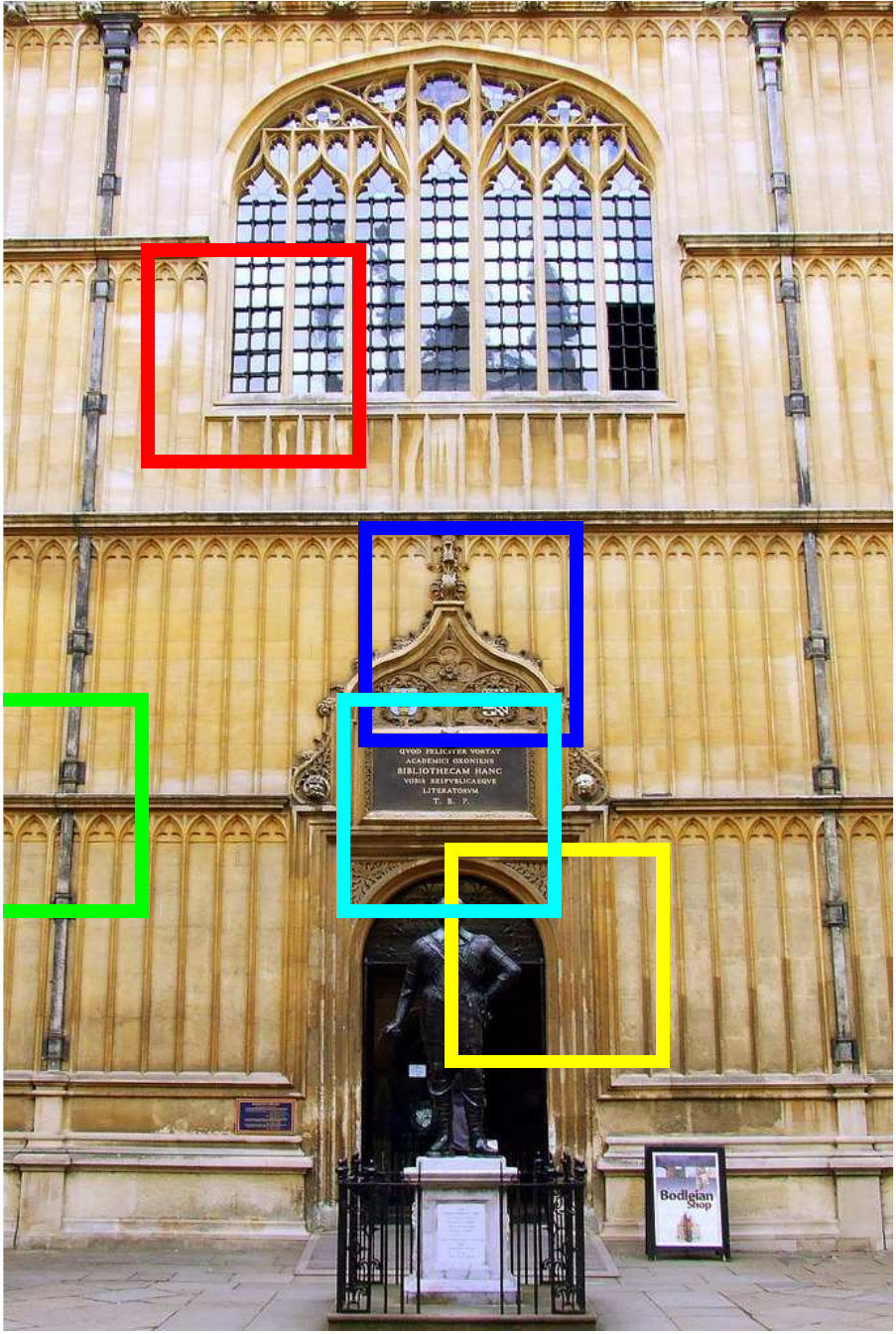}  \hspace{1.5ex}
\includegraphics[height=0.14\textwidth]{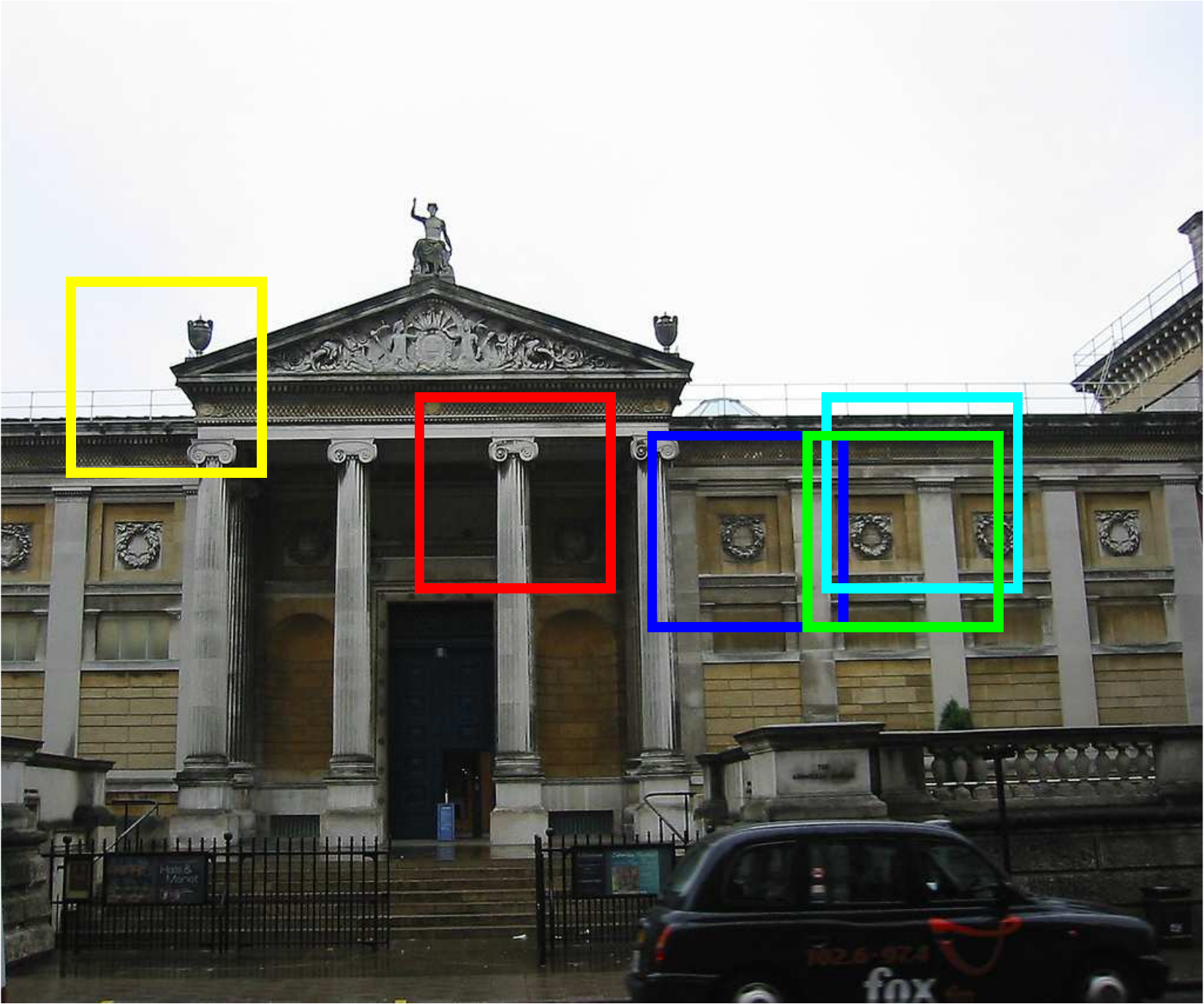}  \
\includegraphics[height=0.14\textwidth]{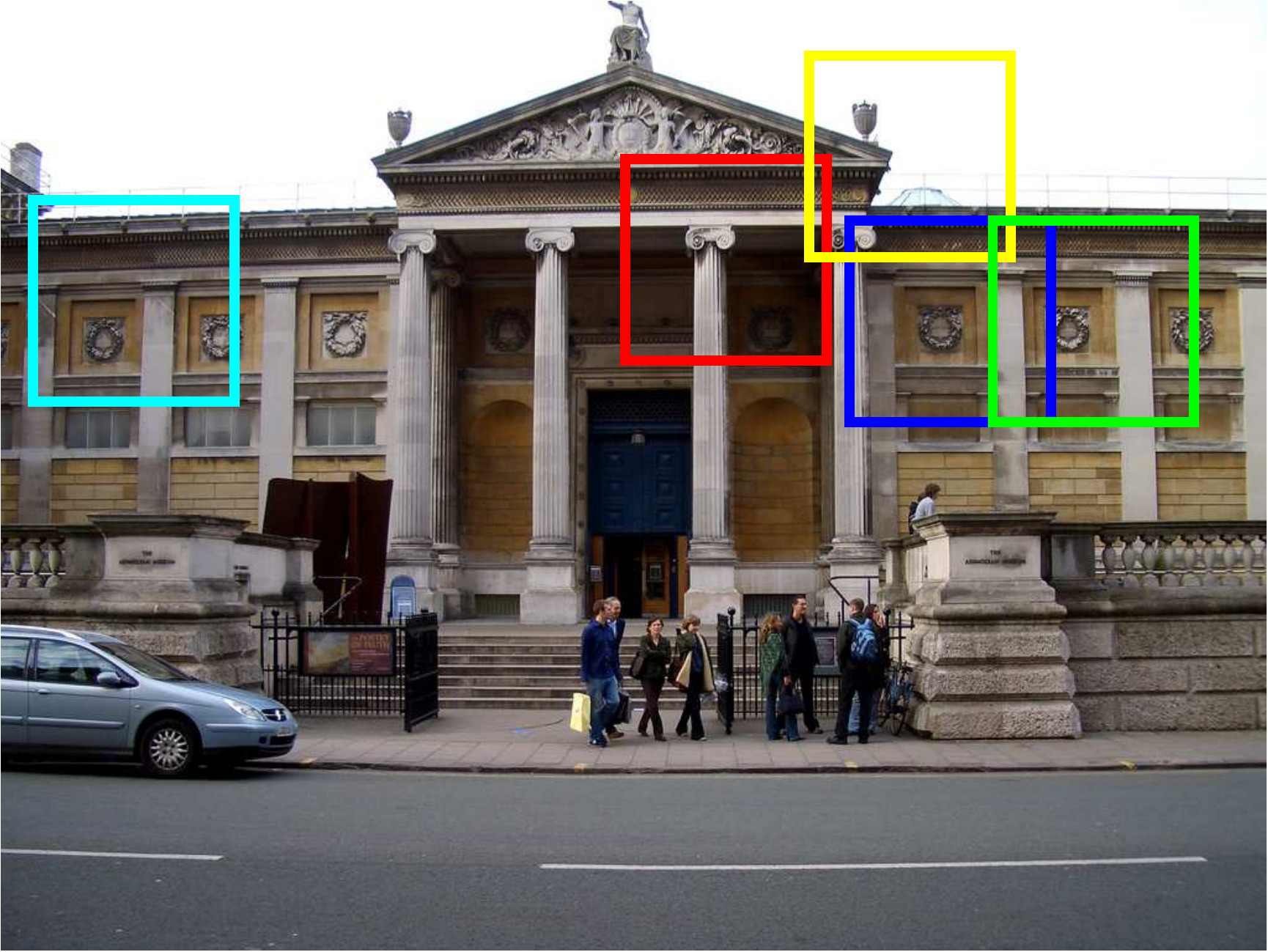}  \hspace{1.5ex}
\includegraphics[height=0.17\textwidth]{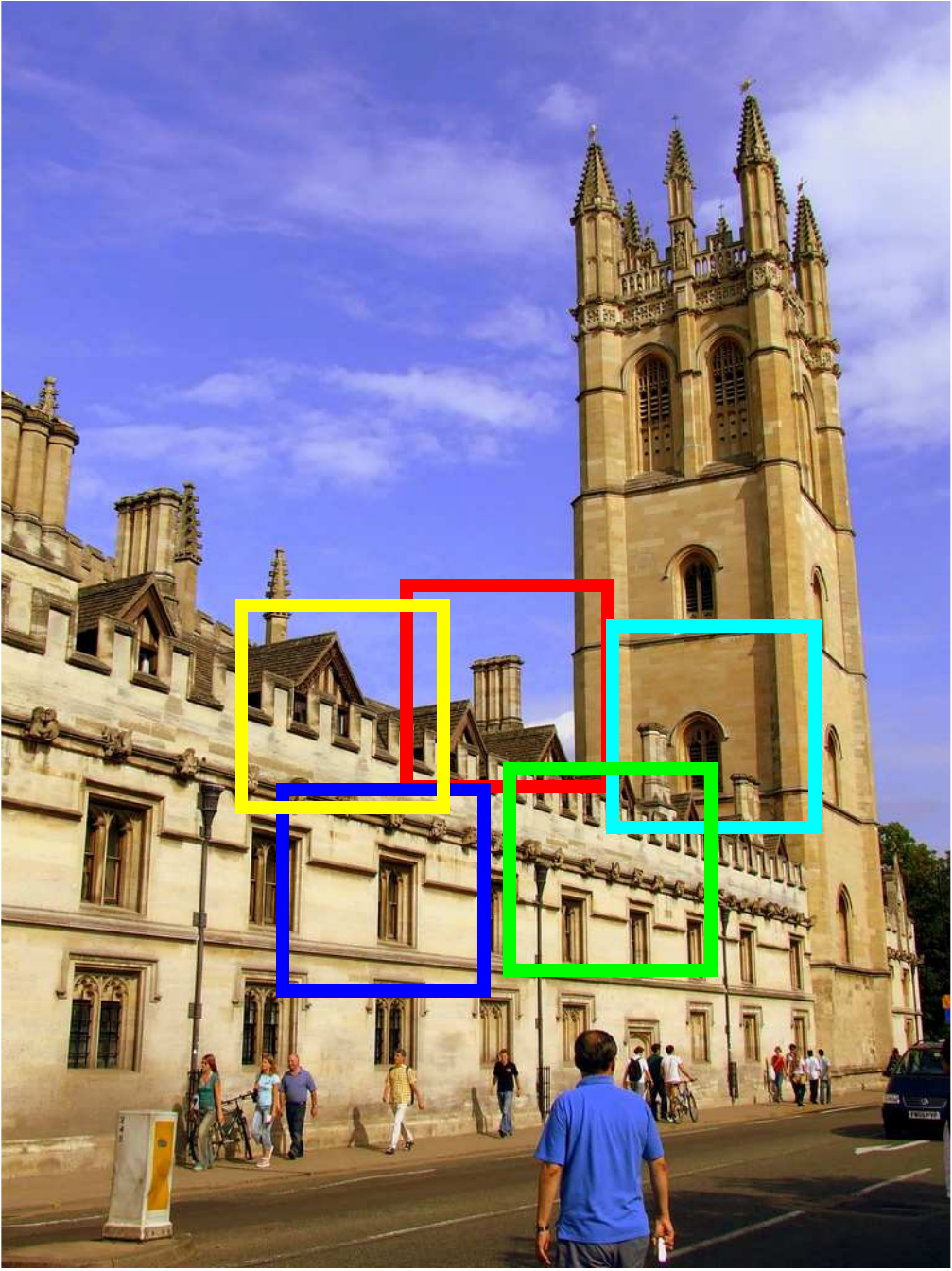}  \
\includegraphics[height=0.17\textwidth]{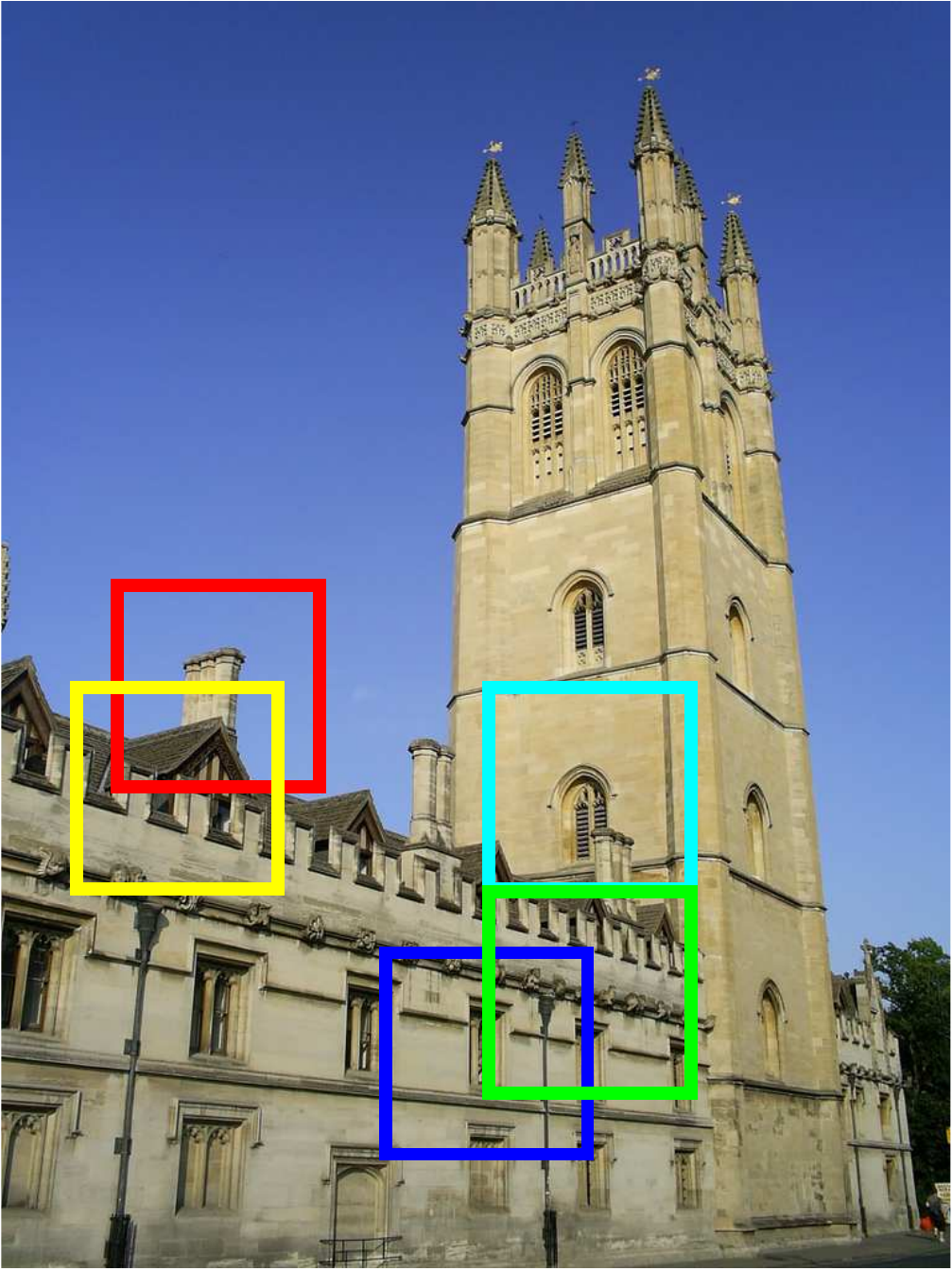}
\vspace{-10pt}\caption{We visualize the receptive fields related to the 5 \gfv components that contribute the most to the image similarity.
Each displayed receptive field corresponds to the maximum response of a feature channel.
A different color is used for each feature channel, while different feature channels are shown for each image pair.
\label{fig:maxmatches}}
\vspace{5pt}
\end{figure*}
We consider a pre-trained CNN and discard all the fully connected layers.
Given an input image $I$ of size $W_I \times H_I$, the activations (responses) of a convolutional layer form a 3D tensor of $W \times H \times K$ dimensions, where $K$ is the number of output feature channels, i.e. multi-dimensional filters.
The spatial resolution $W \times H$ depends on the network architecture, the layer examined, and the input image resolution.
We assume that Rectified Linear Units (ReLU) are applied as a last step, guaranteeing that all elements are non-negative.

We represent this 3D tensor of responses as a set of 2D feature channel responses $\cX = \{\cX_i\}, i = 1 \ldots K$,
where $\cX_i$ is the 2D tensor representing the responses of the $i$\textsuperscript{th} feature channel over the set $\Omega$ of valid spatial locations, and $\cX_i(p)$ is the response at a particular position $p$. 
Therefore, the feature vector constructed by a spatial max-pooling over all locations~\citep{ARSM+14} is given by
\begin{equation}
\vf_{\Omega} = [\f_{\Omega,1} \ldots \f_{\Omega,i} \ldots \f_{\Omega,K}]^\top \text{,~with~} \f_{\Omega,i} = \max_{p \in \Omega} \cX_i(p).
\end{equation}

\textbf{Maximum activations of convolutions (\gfv).}
Two images are compared with the cosine similarity of the $K$-dimensional vectors produced as described above.
This representation, referred to as \gfv, does not encode the location of the activations (unlike activations of fully connected layers), due to the max-pooling operated over a single region of size $W\times H$.
It encodes the maximum ``local'' response of each of the convolutional filters and is therefore translation invariant.
In all the following, we consider the last convolutional layer of the examined networks.

Figure~\ref{fig:maxmatches} visualizes the patches that contribute the most to the image similarity. They correspond either to the same object part or similar parts due to repeated structures.
We extract \gfv from input images of any resolution or aspect ratio by simply subtracting the mean pixel value~\citep{IMKG+14} from the input images. No crop or change of aspect ratio is required~\citep{ARSM+14}.

The max pooling operation that is performed over a single cell offers translation invariance to the resulting representation. 
This is in contrast to representation derived from the fully connected layers that requires objects to be aligned. In our case, we assume that objects are up-right and we simply benefit from the rotation tolerance provided by the CNN due to the training data used. The same stands for the tolerance to scale changes.
\section{Encoding regions into short vectors}
\label{sec:vector}
This section describes how we exploit the activations of the CNN convolutional layers to derive representations for image regions.
Region vectors are aggregated to produce a short signature used in the filtering stage of image retrieval.

\textbf{Region feature vector.}
The feature vector $\vf_\Omega$ described in Section~\ref{sec:background} is a representation for the whole image $I$.
Now, we consider a rectangular region $\reg \subseteq \Omega = [1,W]\times[1,H]$,
and define the regional feature vector
\vspace{1ex}
\begin{equation}
\vf_\reg = [\f_{\reg,1} \ldots \f_{\reg,i} \ldots \f_{\reg,K}]^\top
\end{equation}
\vspace{1ex}
where
$\f_{\reg,i} = \max_{p\in \reg} \cX_i(p)$ is the maximum activation of the $i$\textsuperscript{th} channel on the considered region.
The regions $\reg$ are defined on the space $\Omega$ of all valid positions for the considered feature map (and not on the input image plane).
A region of size 1 corresponds to a feature vector consisting of a single activation at a particular location.
We are now able to construct a representation for multiple regions without re-feeding additional input to the CNN, similarly to recent RNN variants~\citep{RHGS15,Girshick15}, which drastically reduces the processing cost.

Now assume a linear mapping of a given region $\reg$ back to the original image.
The proposed region vector captures a larger image region than the back-projected one, due to the large receptive field.
A similar effect occurs in the context of object detection~\citep{IMKG+14}, where fully connected layers are applied in a sliding window fashion.

\textbf{\rfv: regional maximum activation of convolutions.}
We now consider a set of $R$ regions of different sizes.
The structure of the regions is similar to the one proposed by~\cite{RSMC14}, but we define them on the CNN response maps and not on the original image. We sample square regions at $L$ different scales.
At the largest scale ($l=1$), the region size is determined to be as large as possible, i.e., its height and width are both equal to $\min(W,H)$. The regions are sampled uniformly such that the overlap between consecutive regions is as close as possible to $40\%$.
Remark that the aspect ratio of the original image has an influence on the number $m$ of regions that we extract (1 region only if the input image is square).
At every other scale $l$ we uniformly sample $l\times (l+m-1)$ regions of width $2\min (W,H) / (l+1)$, as illustrated in Figure~\ref{fig:all} (left).

Then we calculate the feature vector associated with each region, and post-process it with \l2-normalization, PCA-whitening~\citep{JC12} and \l2-normalization.
We combine the collection of regional feature vectors into a single image vector by summing them and \l2-normalizing in the end.
This choice keeps the dimensionality low which is equal to the number of feature channels. However, we show in our experiments that the resulting representation, referred to as \rfv, offers a significant better performance than the corresponding \gfv with same dimensionality. Note, the aggregation of the region vectors can be seen as a simple kernel that cross matches all possible regions, including across different scale.

\section{Object localization}
\label{sec:detector}
In this section we propose an extension of integral images to perform approximate max-pooling over a set $\cX$ of 2D feature channel response maps, which provide a rough yet efficient localization to our CNN-based method.

\textbf{Approximate integral max-pooling.}
Noticing that the responses $\cX_i$ are non-negative, we exploit the generalized mean~\citep{DTPB09}  to approximate each feature value $\f_{\reg,i}$ associated with a given region $\reg$ by the estimate
\begin{equation}
\tf_{\reg,i} = \left(\sum_{p\in \reg} \cX_i(p)^\alpha\right)^{\frac{1}{\alpha}}
\approx \max_{p\in \reg} \cX_i(p) = \f_{\reg,i},
\label{equ:boxmaxapprox}
\end{equation}
where the parameter $\alpha>1$ is such that $\tf_i \to \f_i$ when $\alpha \to +\infty$.

\begin{figure*}
\vspace{-10pt}
\centering
\raisebox{18pt}{
  \includegraphics[height=0.15\textwidth]{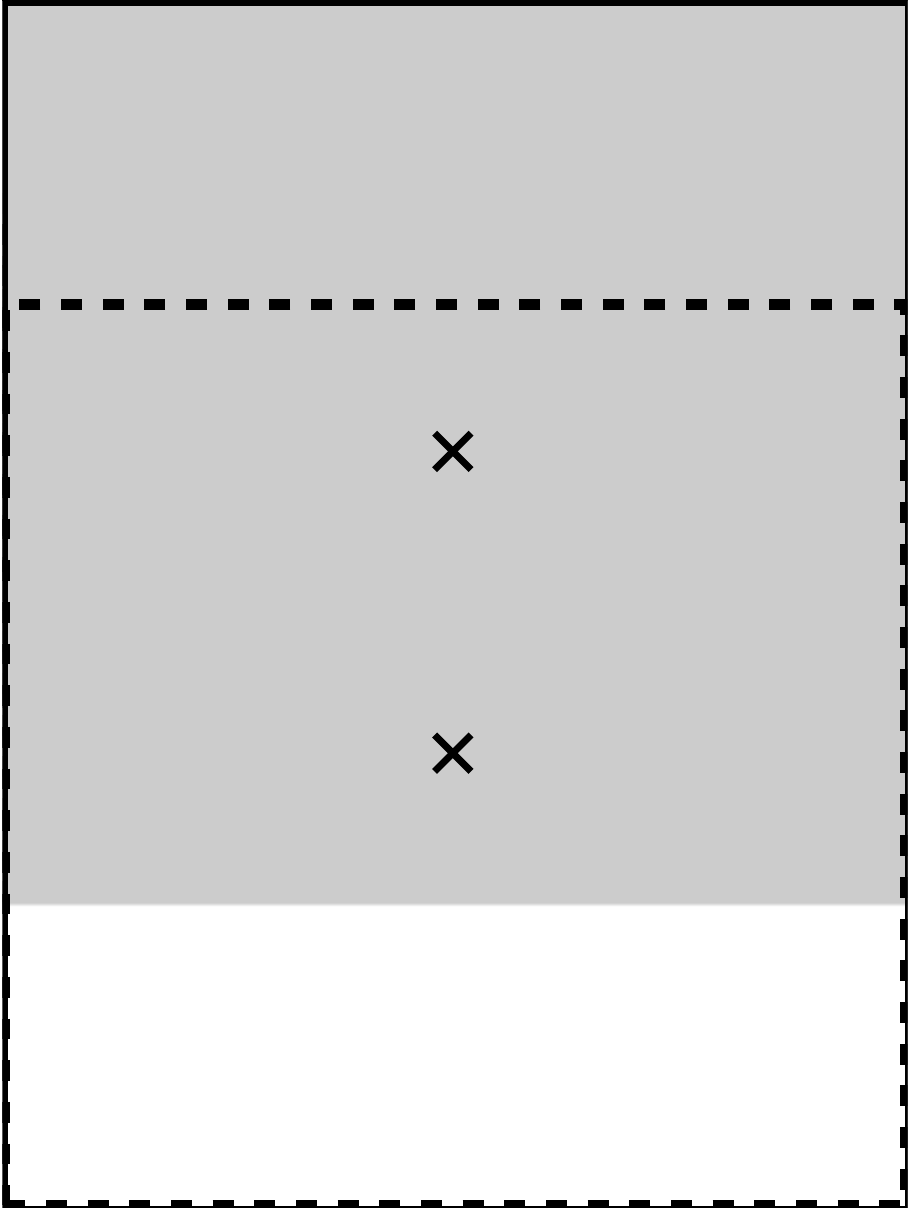}
  \includegraphics[height=0.15\textwidth]{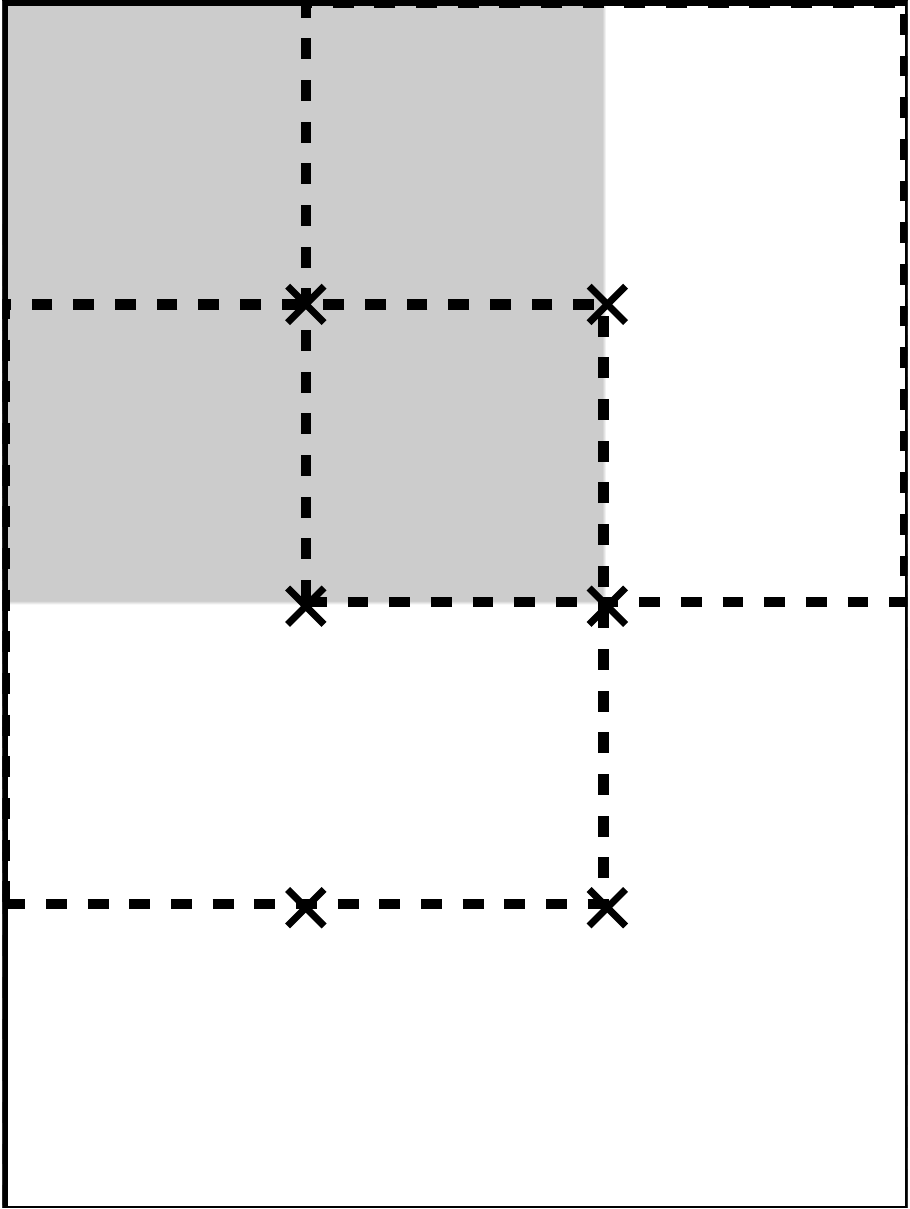}
  \includegraphics[height=0.15\textwidth]{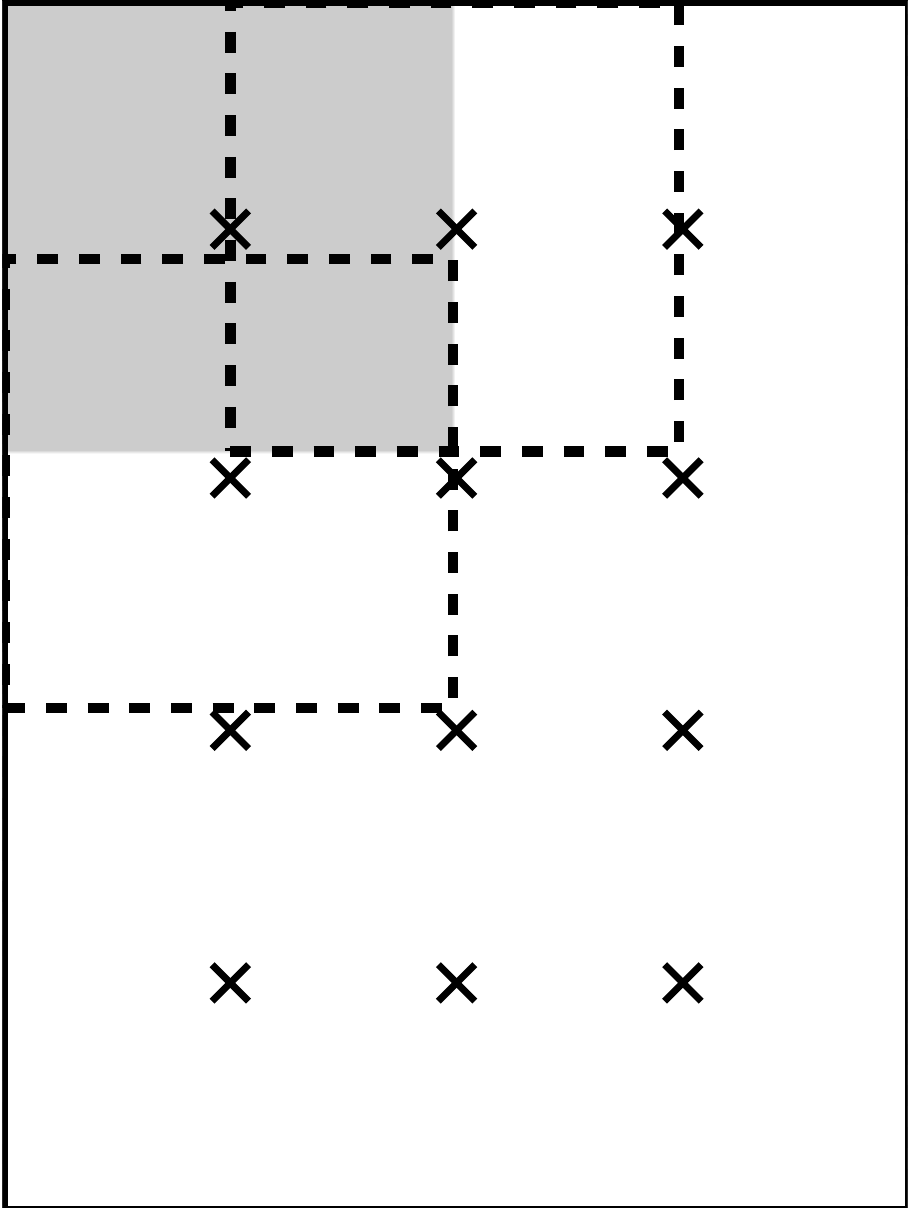}
}
\hfill
\includegraphics[height=0.2\textwidth]{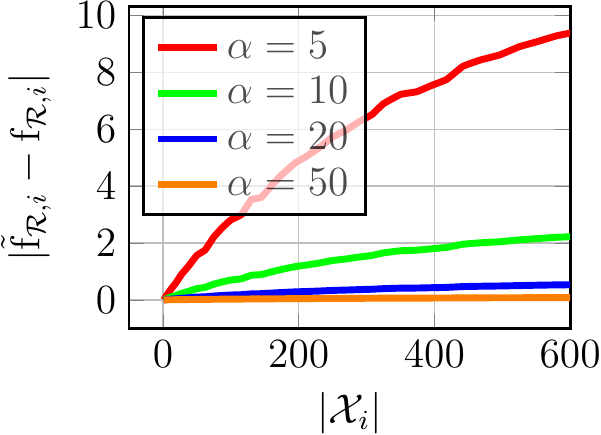}
\hfill
\includegraphics[height=0.2\textwidth]{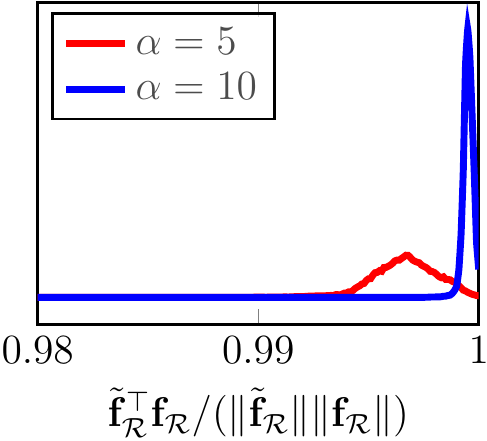}
\vspace{0pt}\caption{Left: Sample regions extracted at 3 different scales ($l=1\ldots 3$). We show the top-left region of each scale (gray colored region) and its neighboring regions towards each direction (dashed borders). We depict the centers of all regions with a cross. Middle: Approximation error of the maximum value versus the size of the response set for different values of exponent $\alpha$. Measurements are performed on 10 randomly selected images by evaluating all possible regions. The responses for this set of images take values in $[0, 151]$. Right: Empirical distribution of the cosine similarity value between the exact vector $\vf_\reg$ and its approximation $\tvf_\reg$. Measurements are collected by constructing the exact and approximate vectors of all possible regions on 10 randomly sampled images.
\label{fig:all}}
\vspace{10pt}
\end{figure*}

Figure~\ref{fig:all} (middle) shows the average approximation error $|\tf_{\reg,i}-\f_{\reg,i}|$ estimated over several image regions.
We report the approximation error as a function of the size of the corresponding response set on which the maximum value is computed.
The various sizes of response sets are an outcome of using all possible regions.
A high value of the exponent $\alpha$ leads to a better approximation, while applying on more elements makes the approximation less precise.

By approximating the maximum in this manner,
we can now use integral images~\citep{VJ01} to approximate the regional feature vector $\vf_\reg$ defined on any rectangular region $\reg$.
For each channel, we construct the integral image of the 2D tensor whose value at position $p$ is equal to $\cX_i(p)^\alpha$, $p\in \reg$.
Then, the sum of \equ{equ:boxmaxapprox} is simply given by the sum of 4 terms~\citep{VJ01}.
This allow us to efficiently compute max-pooling for many regions and therefore to construct the corresponding feature vectors. This is in contrast to the explicit construction of many regions with representation derived from fully connected layers, which is prohibitive due to the need to resize/crop and re-feed each region to the network.

We evaluate the approximation quality by measuring the cosine similarity between the exact vector and its approximate counterpart.
The distribution of this similarity is presented in Figure~\ref{fig:all} (right) and is measured on all possible regions of 10 randomly selected images.
The proposed approximation is very precise even for moderate values of $\alpha$.
We set $\alpha$ equal to 10 in all of our experiments.

\textbf{Window detection.}
Let us now assume that there is another image $Q$ depicting a single object, \ie cropped via a bounding box defining the object of interest.
We denote as $\vq$ the corresponding \gfv feature vector.
The 2D region, defined on the CNN activations $\cX$ of image $I$, that maximizes the similarity to $\vq$ is computed as
\begin{equation}
\hat{\reg} = \arg\max_{\reg \subseteq \Omega}  \frac{\tvf_\reg^\top \vq}{\|\tvf_\reg\| \|\vq\|}.
\end{equation}
The region $\hat{\reg}$ maximizing the similarity is mapped back to the original image $I$ with a precision of $(\frac{W}{W_I}, \frac{H}{H_I})$ pixels, providing a rough localization of the object depicted in $Q$.
The corresponding similarity does not take into account all the visual content of image $I$ and is therefore free from the influence of background clutter.
The brute-force detection of the optimal region by exhaustive search is expensive, as the number of possible regions is in ${\mathcal O}(W^2 H^2)$.
In preliminary tests, we have evaluated a globally optimal solution based on \emph{branch and bound} search, as in ESS~\citep{LBH09}.
The necessary bounds are trivially derived for our representation.
The search is not significantly sped up in our case:
The maxima are not distinct enough and a large number of regions are considered, while the overhead of maintaining the priority queue is high.

\textbf{\deeploc: approximate max-pooling localization.}
Instead, we restrict the number of regions that we evaluate and locally refine the best ones with simple heuristics. Candidate regions
are uniformly sampled with a \emph{search step} equal to $t$. In addition, regions having an aspect ratio larger than $s$ times that of the query region are discarded. The parameters of the best region are refined in a coordinate descent manner, while allowing a maximum change of 3 units. The refinement process is repeated up to 5 times. Experiments show that the overlap of the detected region to the optimal one is high.

\vspace{-1.5ex}
\section{Retrieval, localization and re-ranking}
\label{sec:retrieval}
\textbf{Initial retrieval}.
The \gfv or \rfv feature vector is computed for all databases images.
Similarly, at query time we process the query image and extract the corresponding feature vector.
During the filtering stage we directly evaluate cosine similarity between the query and all the database vectors.
Therefore, we obtain the initial ranking based on the similarity of \gfv or \rfv vectors.

\textbf{Re-ranking}.
We consider a second re-ranking stage, as typically performed in spatial verification~\citep{PCISZ07} with local features.
A short-list of $N$ top-ranked images is considered and \deeploc, as described in Section~\ref{sec:detector}, is applied on pairs of query and database images.
Note that the query is now represented by the \gfv vector, since this is used in \deeploc, while the database image is represented by $\cX$.
For each re-ranked image we obtain a score given by the region that maximizes the similarity to the query.
This similarity is used to re-rank the elements of the short-list.
Furthermore, a rough localization of the query object is available.

\noindent \emph{Remarks}: At the filtering stage, whitened \gfv (whitening as described in Section~\ref{sec:experiments}) or \rfv can be used, while the localization procedure employs similarity with respect to \l2-normalized \gfv.
However, once the query object is localized, then, similarity between the query and the detected region is computed via whitened \gfv or \rfv, depending on the chosen filtering method. This similarity score is used to perform re-ranking.
The required representation is constructed on query time only for the detected region and is acquired efficiently with integral images.

\textbf{Query expansion} (QE). Re-ranking brings positive images at the very top ranked positions.
Then, we collect the 5 top-ranked images, merge them with the query vector, and compute their mean.
Finally, the similarity to this mean vector is adopted to re-rank once more the top $N$ images.
\smallskip
\vspace{-1.5ex}
\section{Implementation details}
\label{sec:implementation}
We observe that thresholding the response values of $\cX$ which are larger than 128 ($0.001\%$ of all responses) and mapping each value to the closest smaller integer (floor operation) leads to insignificant losses.
This allows the computation of $\alpha$-th power with a lookup table and speeds-up the construction of integral images.
Moreover, we approximate the $\alpha$-th root of~\equ{equ:boxmaxapprox} by performing binary search on the same lookup table of $\alpha$-th power.
This process allows the optimal window search to be more efficient.

The response maps represented by $\cX$ are sparse~\citep{AGM14}.
In particular, using the network of~\cite{KSH12} on Oxford buildings dataset~\citep{PCISZ07} results in $81\%$ of response values being zero, which is convenient for storage purpose.
We further decrease the memory requirements by uniformly quantizing the responses into $8$ values.
This results in more elements mapped to the same value. Therefore, we store the positions of non-zeros values with delta coding and use only 1 byte per non-zero element.
Note that an image of resolution equal to $1024\times 768$ corresponds to feature channel response maps of size $30 \times 22$ using the same network.
Finally, an image requires around 32\,kB of memory. At re-ranking time we construct one integral image at a time and use double precision (8 bytes) for its elements.
\bigskip
\section{Experiments}
\label{sec:experiments}
This section presents the results of our compact representation for image retrieval, evaluate the localization accuracy \deeploc, and finally employ it for retrieval re-ranking.

\textbf{Experimental setup.}
We evaluate the proposed methods on Oxford Buildings dataset~\citep{PCISZ07} and Paris dataset~\citep{PCISZ08}, which are composed of 5063 and 6412 images, respectively.
We refer to these datasets as Oxford5k and Paris6k.
We additionally use 100k Flickr images~\citep{PCISZ07} to compose Oxford105k and Paris106k, respectively. A distractor set of 1 million images from Flickr~\citep{JDS10a} is additionally used to go at larger scale.
Retrieval performance is measured in terms of mean Average Precision (mAP).
We follow the standard protocol and use the bounding boxes defined on the query images\footnote{The query regions are cropped and then used as input to the CNN.}.
These bounding boxes are also employed to evaluate localization accuracy. 
PCA is learned on Paris6k when testing on Oxford5k and vice versa. In order to be fair, we directly compare our results only to previous methods that do not perform learning on the test set.

The focus of our work is not to train a CNN, but to extract visual descriptors from its convolutional layers.
We use networks widely used in the literature: AlexNet by~\cite{KSH12} and the very deep network (VGG16) by~\cite{SZ14}. 
We choose VGG16 instead of VGG19 because we observe that the latter does not always attain better performance while it has higher feature extraction cost.
Our representation is extracted from the last pooling layer, which has 256 feature channels for AlexNet and 512 for VGG16.
MatConvNet~\citep{VL14} is used to extract the features.

\begin{table}
\caption{Left: Comparison between the exhaustive sliding window and our alternative of window sampling and refinement. We report the average IoU \wrt the globally optimal window and the average percentage of windows evaluated \wrt to the exhaustive search (noted by \%W). Measurements are conducted on all pairs of Oxford5k query images and their corresponding positive images. Right: Performance (mAP) of \gfv and \rfv on Oxford5k. Resol. corresponds to the input image resolution (maximum dimension).  \label{tab:both}}
\vspace{1.5ex}
\setlength\extrarowheight{1pt}
\small
\centering
\begin{minipage}{0.49\textwidth}
\begin{tabular}{|@{\sssp}c@{\sssp}|@{\sssp}r@{\msp}r@{\sssp}|@{\sssp}r@{\msp}r@{\sssp}|@{\sssp}r@{\msp}r@{\sssp}|} \hline
\multirow{3}{*}{Search step $t$}  & \multicolumn{6}{c|}{Aspect ratio change threshold $s$}\\ \cline{2-7}
		   														&  \multicolumn{2}{c|}{1.1} & \multicolumn{2}{c|}{1.5} & \multicolumn{2}{c|}{2.0} \\  \cline{2-7}
		   														& IoU  & \%W 								& IoU  & \%W               &           IoU  & \%W     \\ \hline \hline
		 														1 & 81.8 & 8.9 								& 88.7 & 27.5 						 &           93.7 & 46.3    \\
		 											      2 & 79.9 & 0.5                & 83.8 & 2.0               &           86.6 & 3.6     \\
		 											      3 & 78.7 & 0.2                & 81.2 & 0.5               &           83.6 & 0.8     \\
		 											      4 & 77.0 & 0.1                & 79.5 & 0.2               &           81.5 & 0.3     \\
		 											      5 & 75.8 & 0.1                & 79.0 & 0.1               &           80.9 & 0.2     \\ \hline
\end{tabular}
\end{minipage}
\begin{minipage}{0.49\textwidth}
\begin{tabular}{|@{\ssp}c@{\ssp}|@{\ssp}c@{\ssp}|@{\ssp}c@{\ssp}|c@{\bsp}c@{\bsp}c@{\bsp}c|} \hline
\multirow{2}{*}{Network} & \multirow{2}{*}{Resol.}  & \multirow{2}{*}{\gfv} & \multicolumn{4}{c|}{\rfv} 														\\ \cline{4-7}
												 & 			&								   					            		&	 			$L$=1   &		 	$L$=2  & 		 $L$=3 &	 		 $L$=4 	\\ \hline \hline
\multirow{2}{*}{AlexNet} &					 1024					 & 		 								 			44.9 						&       47.9    &     54.6   &     \textbf{56.1}  &       55.6   \\
												 &            724 			   & 										       44.8            &       48.4    &     54.4   &     54.3  &       52.6   \\
\hline
\multirow{2}{*}{VGG16}   & 					 1024          &                           55.2            &       57.3    &     64.5   &     \textbf{66.9}  &       \textbf{67.4}   \\
		                     &            724          &                        52.2            &       54.8    &     58.0   &     60.9  &       60.3   \\
\hline
\end{tabular}
\end{minipage}
\vspace{2ex}
\end{table}

\textbf{Localization accuracy.}
To evaluate the accuracy of \deeploc, we employ pairs of Oxford5k query images and their corresponding positive images.
We first perform exhaustive search to detect the globally optimal window.
Then, we apply our speeded-up detector that evaluates fewer regions and in the end refines the best one.
In both cases the approximate max-pooling is used for each window evaluation.
We report Intersection over Union (IoU) with the optimal window and the percentage of windows evaluated compared to the exhaustive case.
Results are shown in Table~\ref{tab:both} (left).
We provide a large speed-up while maintaining a high overlap with the optimal detection.
Recall that our purpose is to apply this detector for fast re-ranking.
Measuring IoU provides evidence for localization accuracy, however we observed that it does not directly impact retrieval performance.
We finally set $s=1.1$ and $t=3$ for re-ranking usage.

In order to evaluate the localization accuracy with respect to ground-truth annotation we cross-match all 5 query images that exist per building.
One of them is used as a query (cropped bounding box), while for the other we compare the detected region to the ground-truth annotation.
Exhaustive evaluation achieves an IoU equal to 52.6\% (52.9\%) and the speeded-up approach achieves 51.3\% (51.4\%) on Oxford5k (Paris6k) datasets.
The accuracy loss is limited, while the localization is approximately 180 times faster. 
\deeploc provides a rough localization at low computational cost.
Such a setup results in an average re-ranking query time of 2.9 sec using AlexNet, when re-ranking 1000 images with a single threaded implementation. 
\begin{figure*}
\centering
  \includegraphics[height=0.25\textwidth]{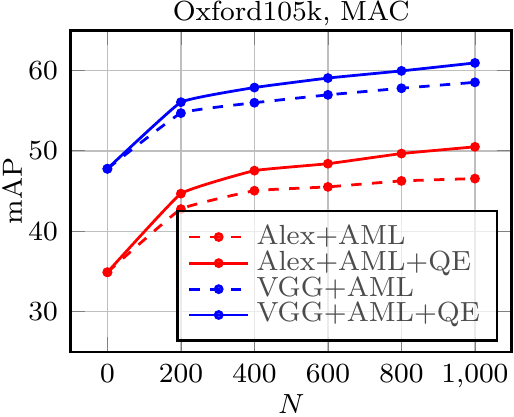}
\hfill
  \includegraphics[height=0.25\textwidth]{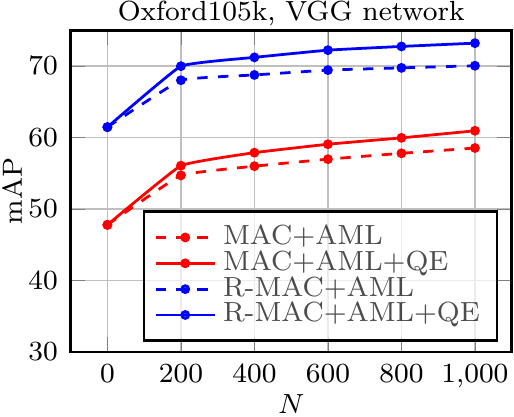}
  \hfill
 \includegraphics[height=0.25\textwidth]{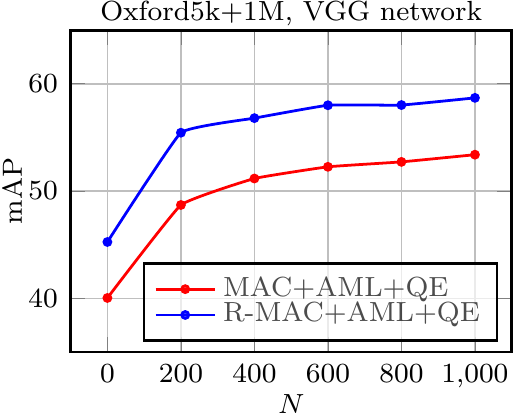}
\caption{Performance of retrieval with re-ranking by \deeploc versus number of re-ranked images on Oxford105k and Oxford5k combined with 1M distractor images.
\label{fig:rerank_map}}
\end{figure*}

\textbf{Retrieval and re-ranking.}
We evaluate retrieval performance using \gfv and \rfv compact representations.
The \gfv vectors are \l2-normalized, PCA-whitened and \l2-normalized once more, while the corresponding processing of the \rfv is as described in Section~\ref{sec:vector}. Table~\ref{tab:both} (right) presents the results on Oxford5k.
We evaluate different input image resolutions and observe that the original image size (1024) provides higher performance.
Note that \gfv is similar to the one proposed by~\cite{ARSM+14}, however their process remains constrained by  standard input size and aspect ratio.
The proposed \rfv gives a large performance improvement at no extra cost, as both feature vectors have exactly the same dimensionality.
Regions of different scales are aggregated together, meaning that $L=3$ combines regions at scales $l=1$, $l=2$, and~$l=3$.
We set $L=3$ in the following. 
In order to decompose the components of \rfv, we construct \rfv by aggregating only regions of $l=3$. It achieves mAP equal to 63.0 on Oxford5k with VGG16. 
Aggregating both regions of $l=2$ and $l=3$ improves to 65.4.
Finally, adding $l=1$ (original \rfv) performs 66.9 (see Table \ref{tab:both} right).
Filtering time is 12 ms on average for Oxford105k. 

Next, we employ \deeploc for image re-ranking and conduct performance evaluation on Oxford105k by re-ranking up to 1000 images.
The performance is consistently improved as shown in Figure~\ref{fig:rerank_map}.
\rfv brings a larger benefit and VGG16 performs better than AlexNet.
Query expansion, as described in Section~\ref{sec:retrieval}, improves the performance at low extra cost, since similarity is re-computed only for the re-ranked short-list.
Finally, we carry out experiment at larger scale with 1M distractor images and present results in Figure~\ref{fig:rerank_map}. \deeploc improves the performance by 13\% mAP.

\begin{figure*}[t]
\scriptsize
\newcommand{\figh}{10ex}
\centering

\begin{tabular}{@{\sssp}c@{\sssp}}\includegraphics[height=\figh]{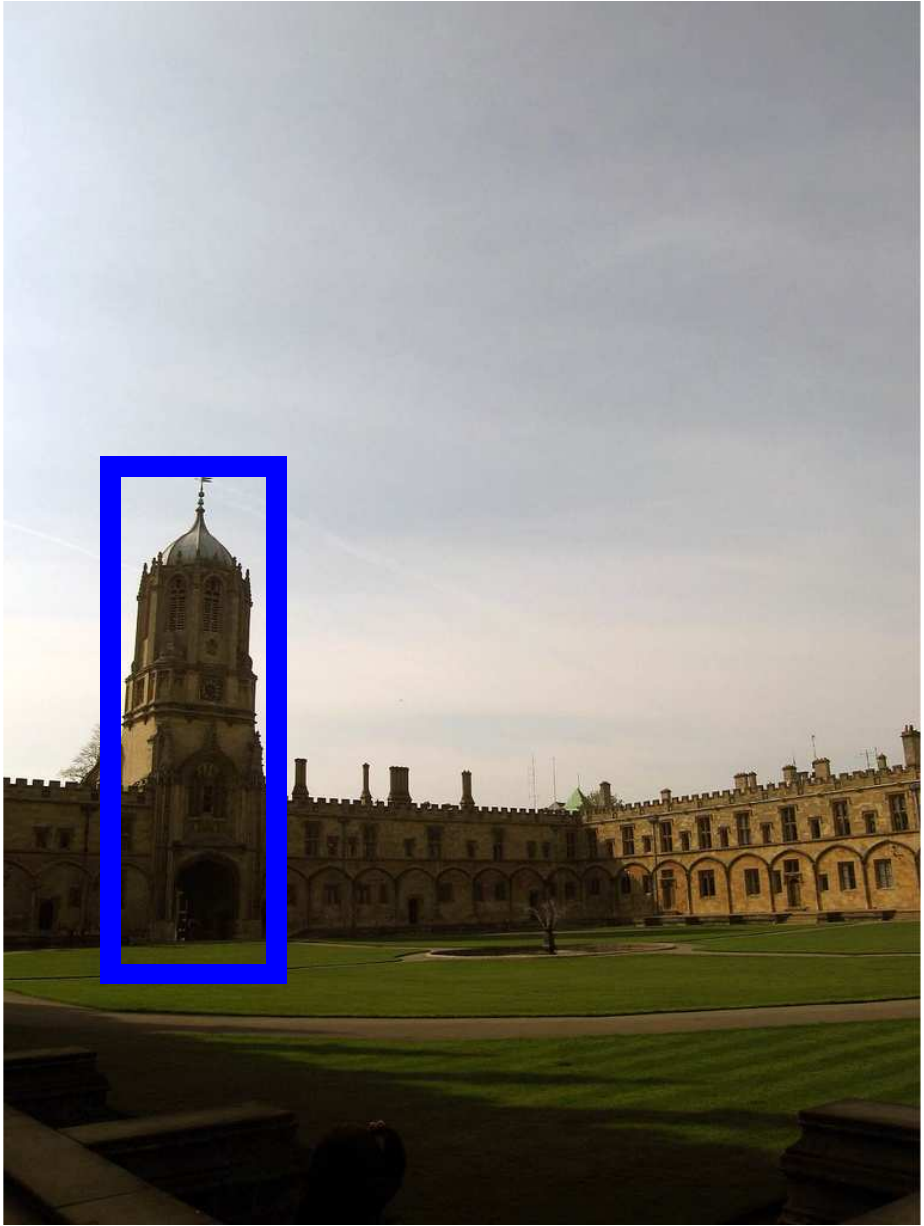}\end{tabular} 
\begin{tabular}{@{\sssp}c@{\sssp}}\includegraphics[height=\figh]{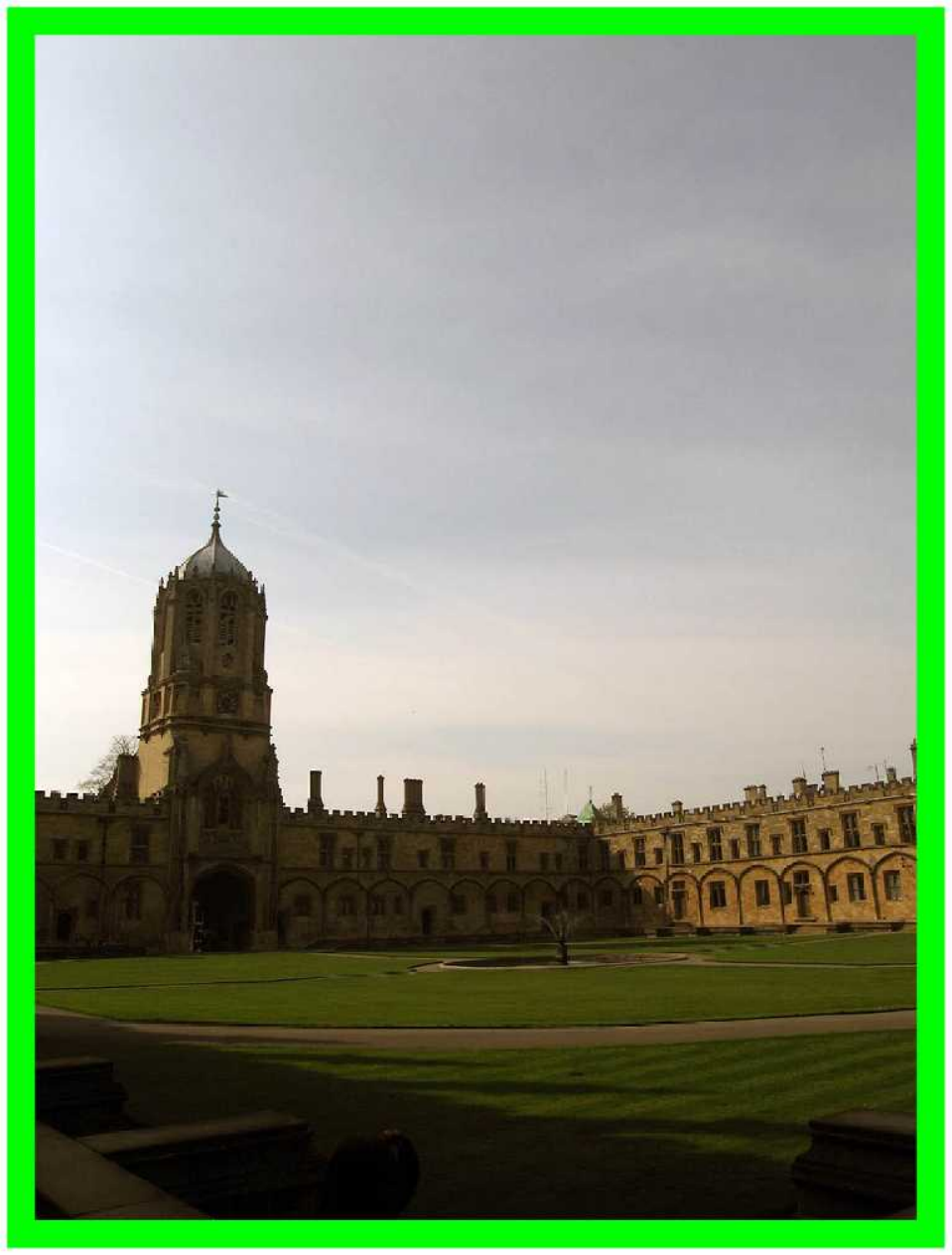}\end{tabular}
\begin{tabular}{@{\sssp}c@{\sssp}}\includegraphics[height=\figh]{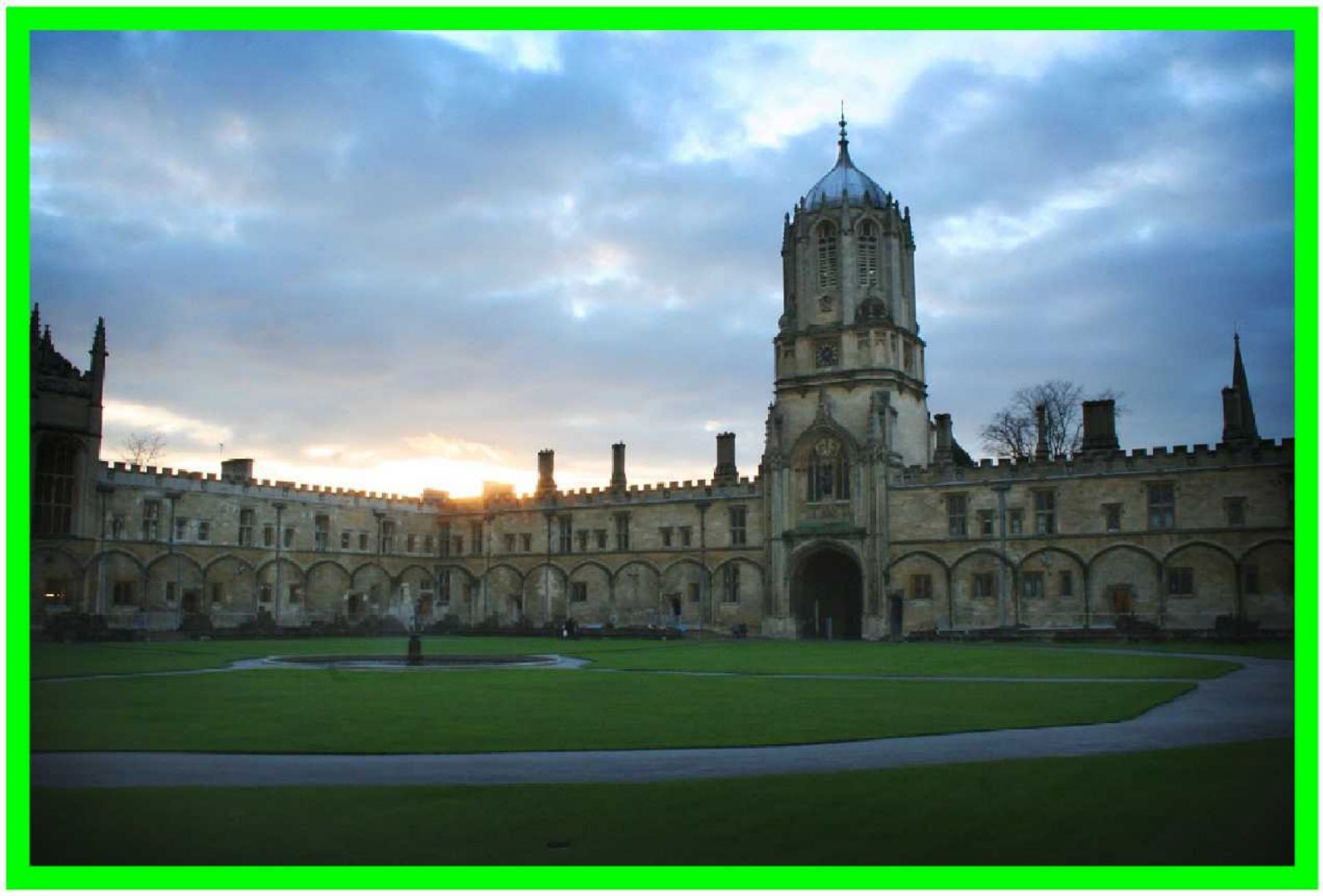}\end{tabular}
\begin{tabular}{@{\sssp}c@{\sssp}}\includegraphics[height=\figh]{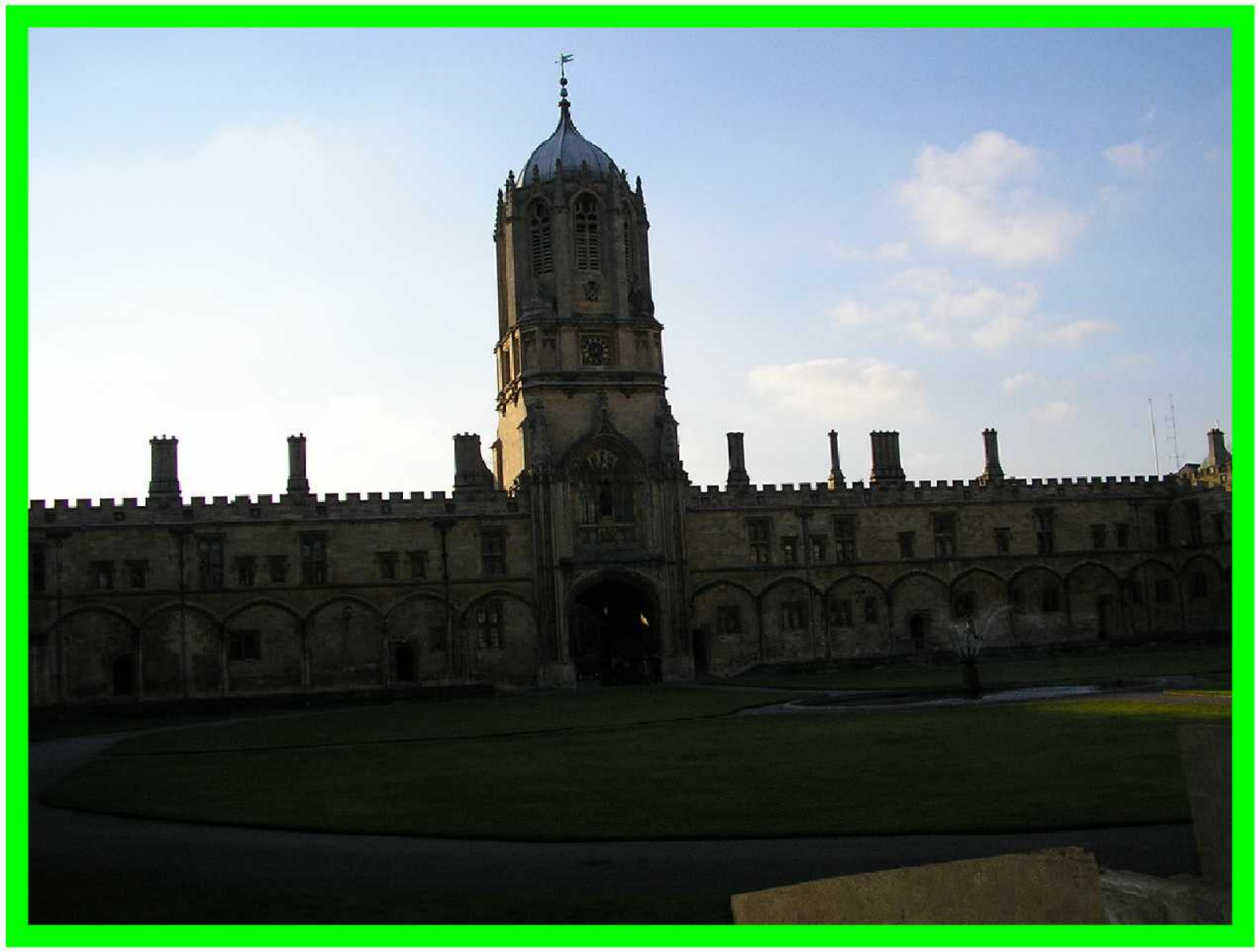}\end{tabular}
\begin{tabular}{@{\sssp}c@{\sssp}}\includegraphics[height=\figh]{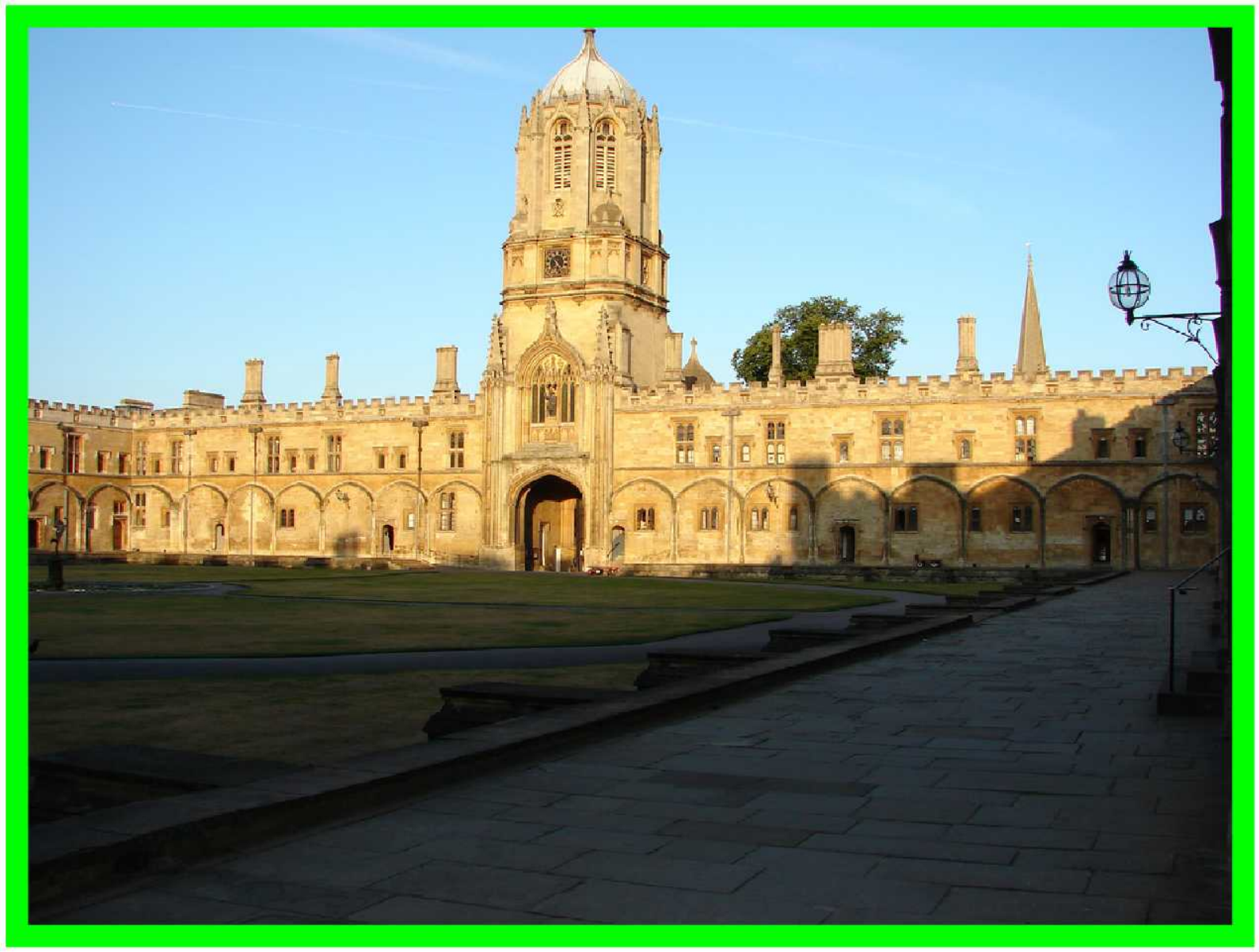}\end{tabular}
\begin{tabular}{@{\sssp}c@{\sssp}}\includegraphics[height=\figh]{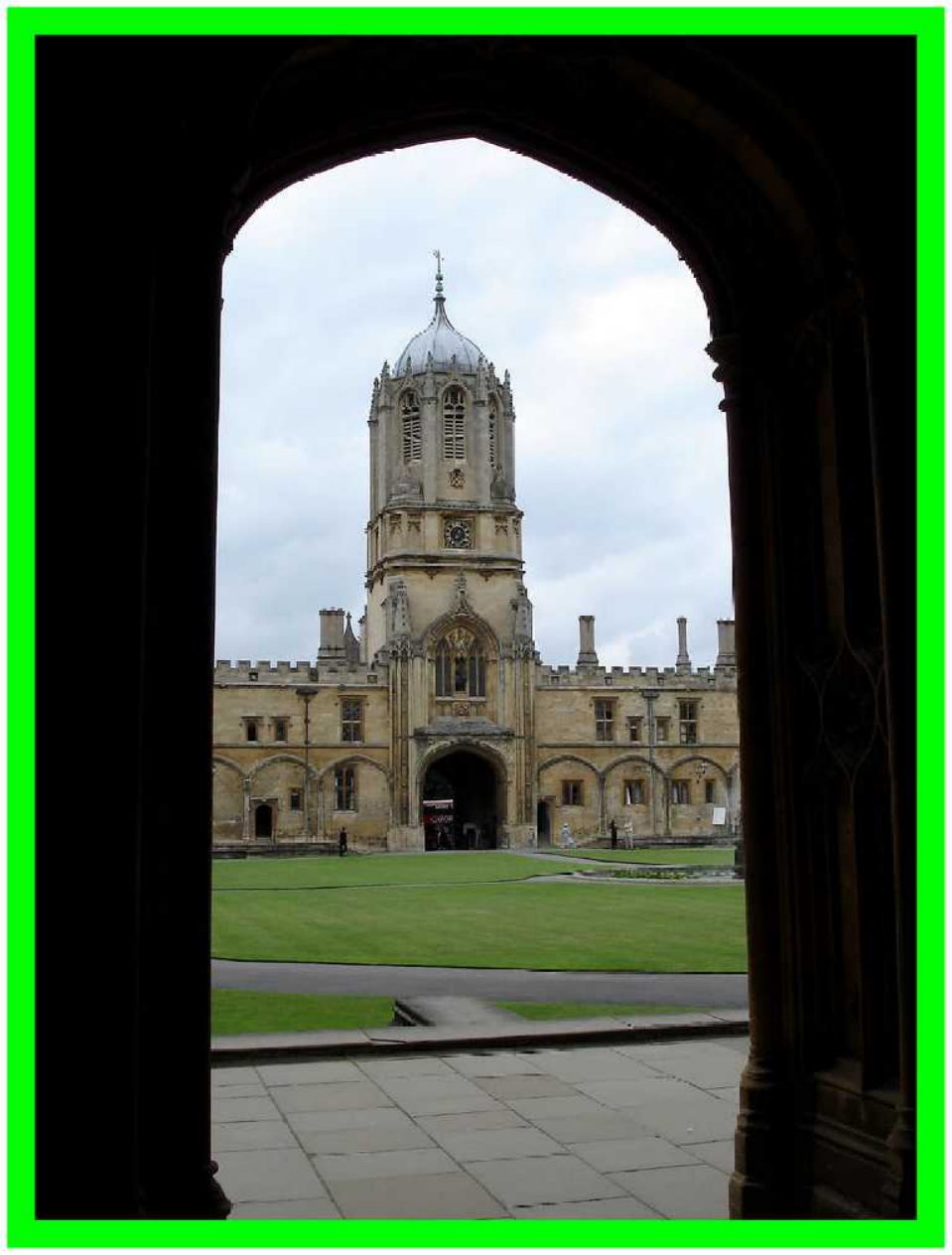}\end{tabular}
\begin{tabular}{@{\sssp}c@{\sssp}}\includegraphics[height=\figh]{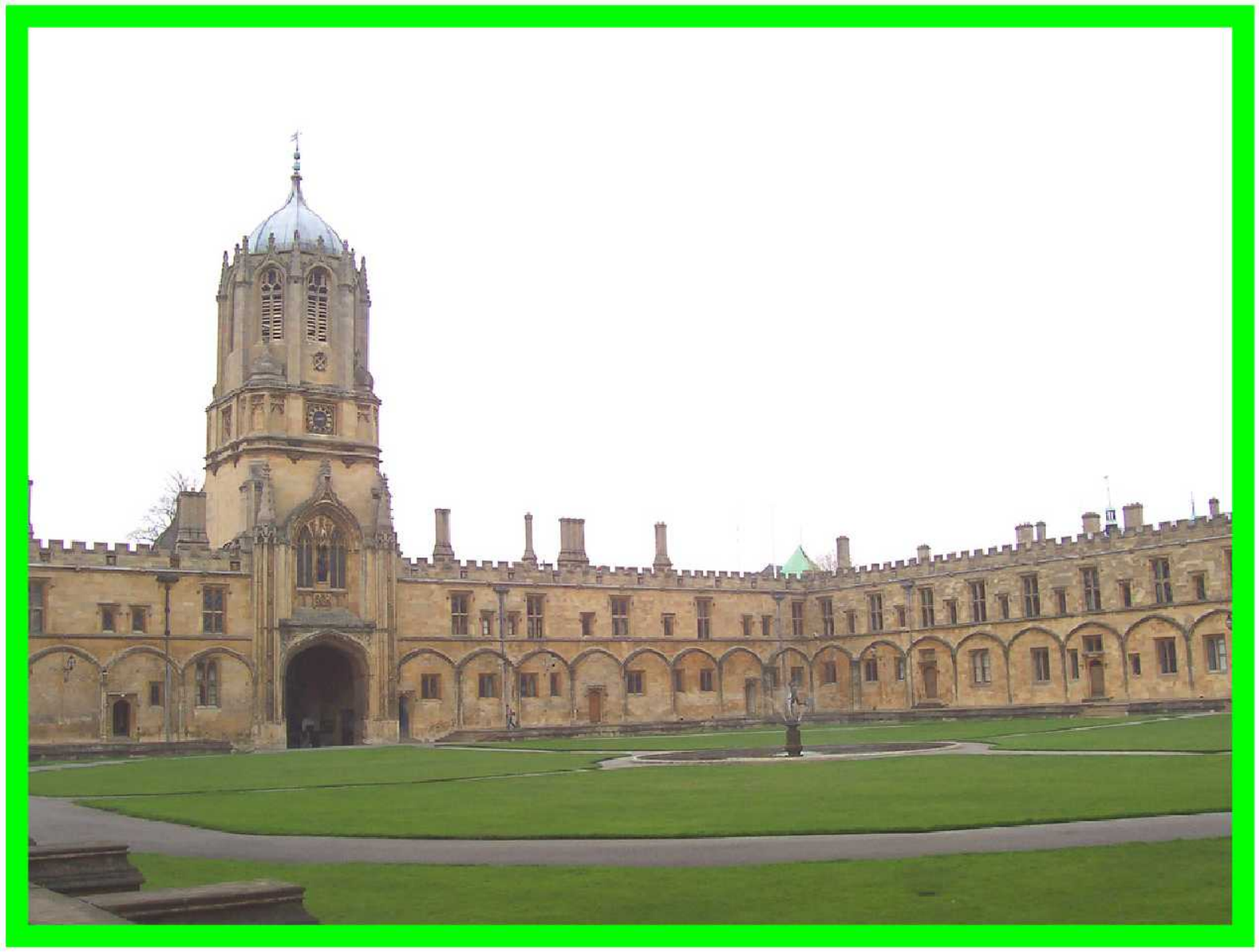}\end{tabular}
\begin{tabular}{@{\sssp}c@{\sssp}}\includegraphics[height=\figh]{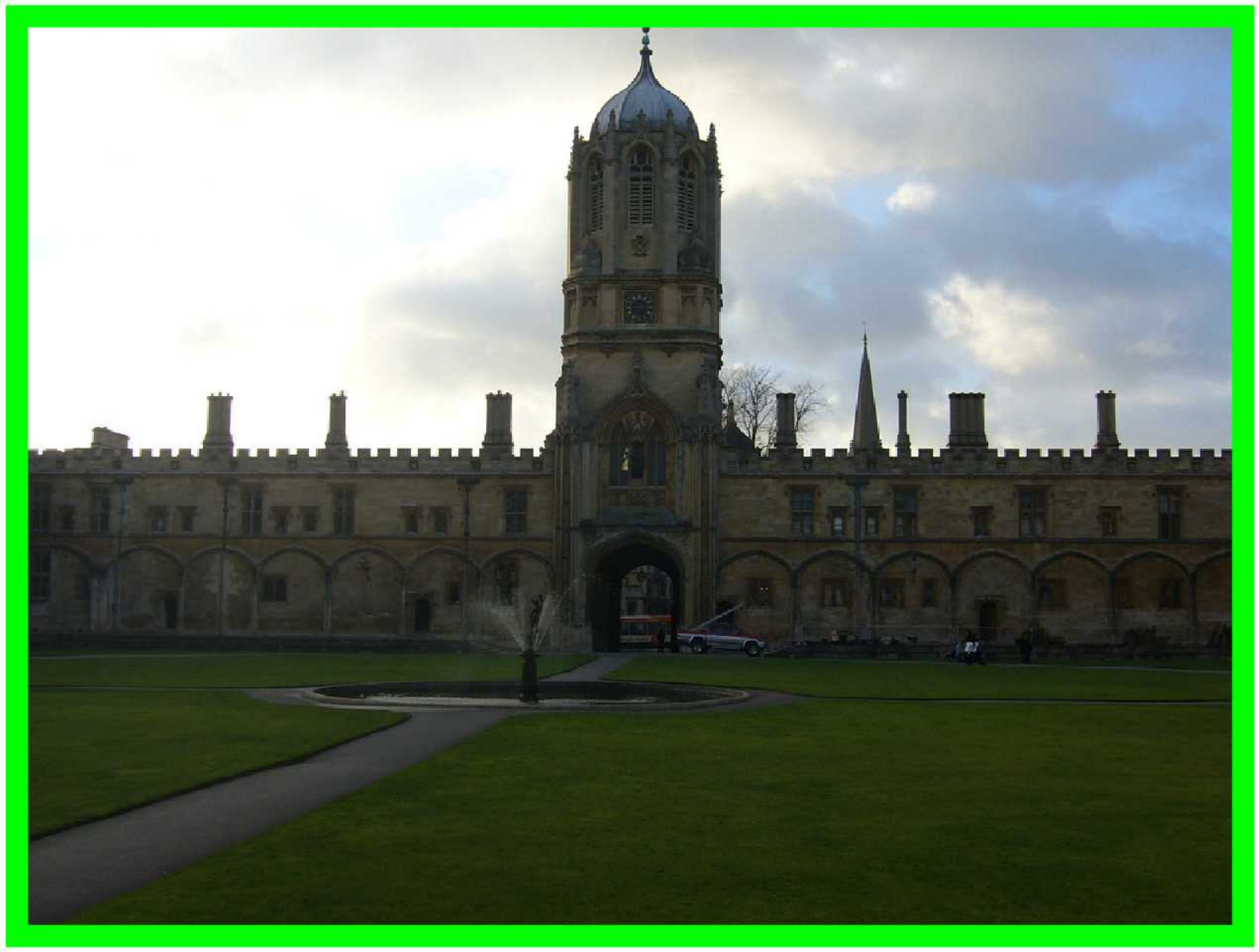}\end{tabular}
\begin{tabular}{@{\sssp}c@{\sssp}}\includegraphics[height=\figh]{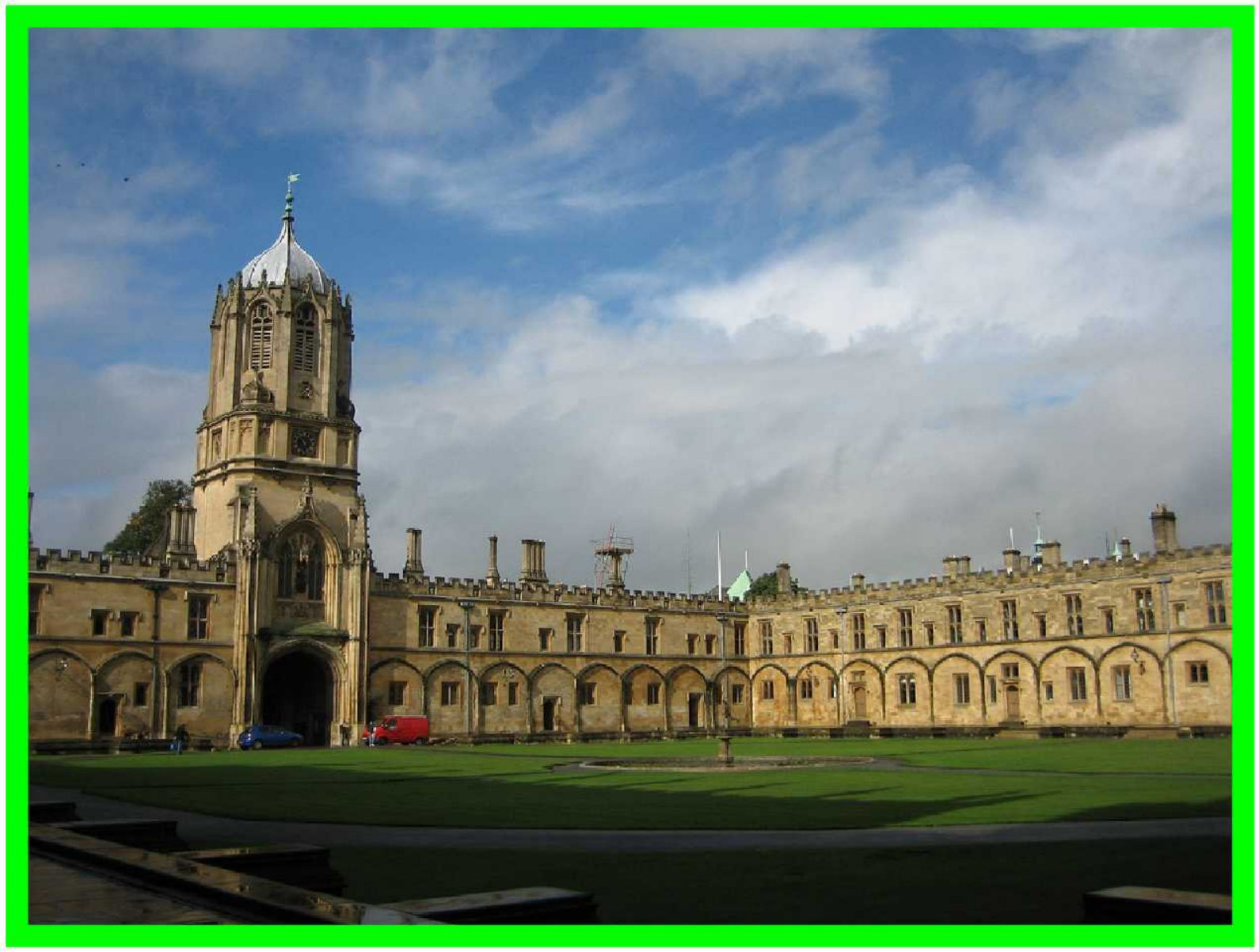}\end{tabular}

\begin{tabular}{@{\sssp}c@{\sssp}}\includegraphics[height=\figh]{figs/rerank/23//q23_5865_7105.pdf}\\Query\\ \end{tabular} 
\begin{tabular}{@{\sssp}c@{\sssp}}\includegraphics[height=\figh]{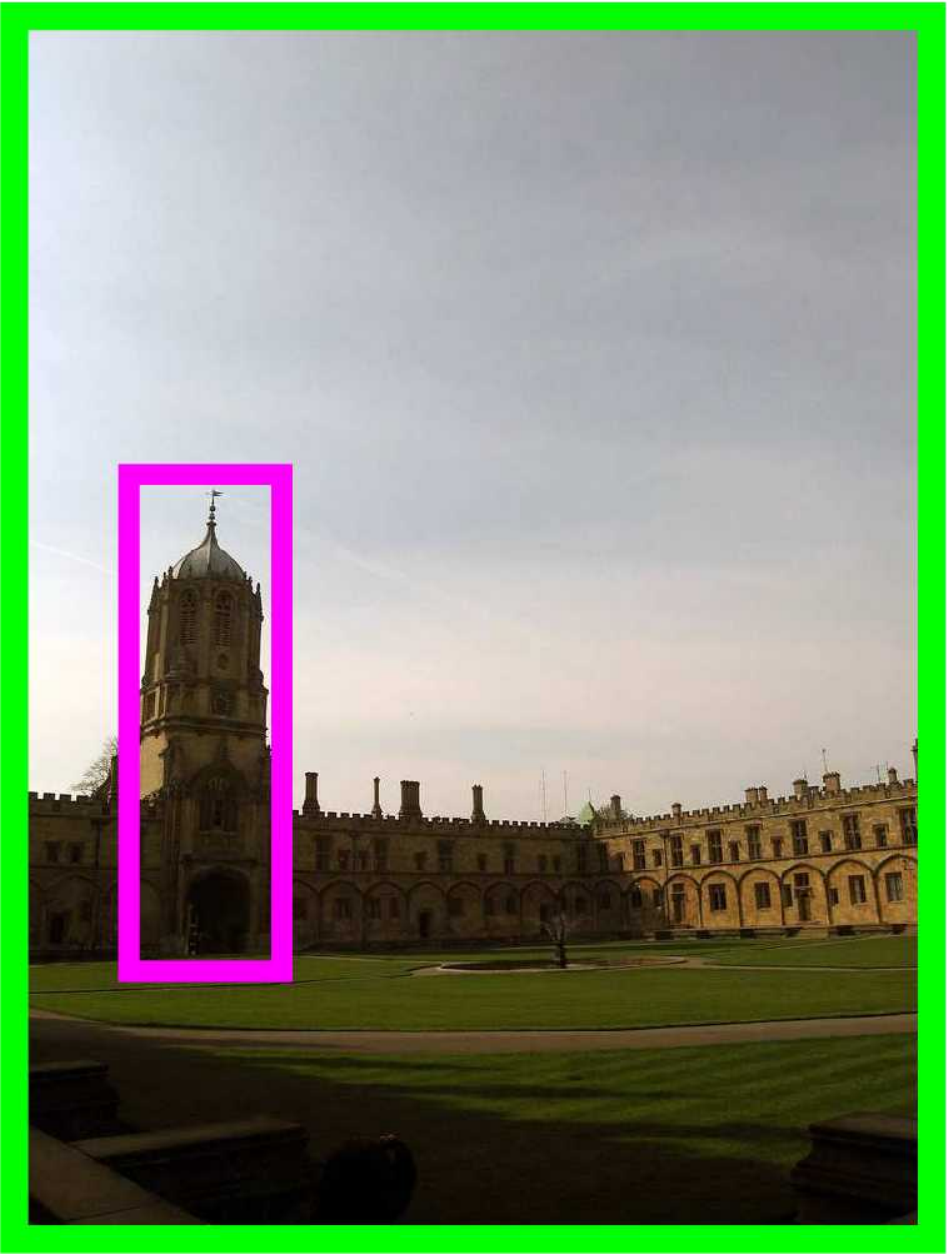}\\1 $\rightarrow$ 1\\ \end{tabular} 
\begin{tabular}{@{\sssp}c@{\sssp}}\includegraphics[height=\figh]{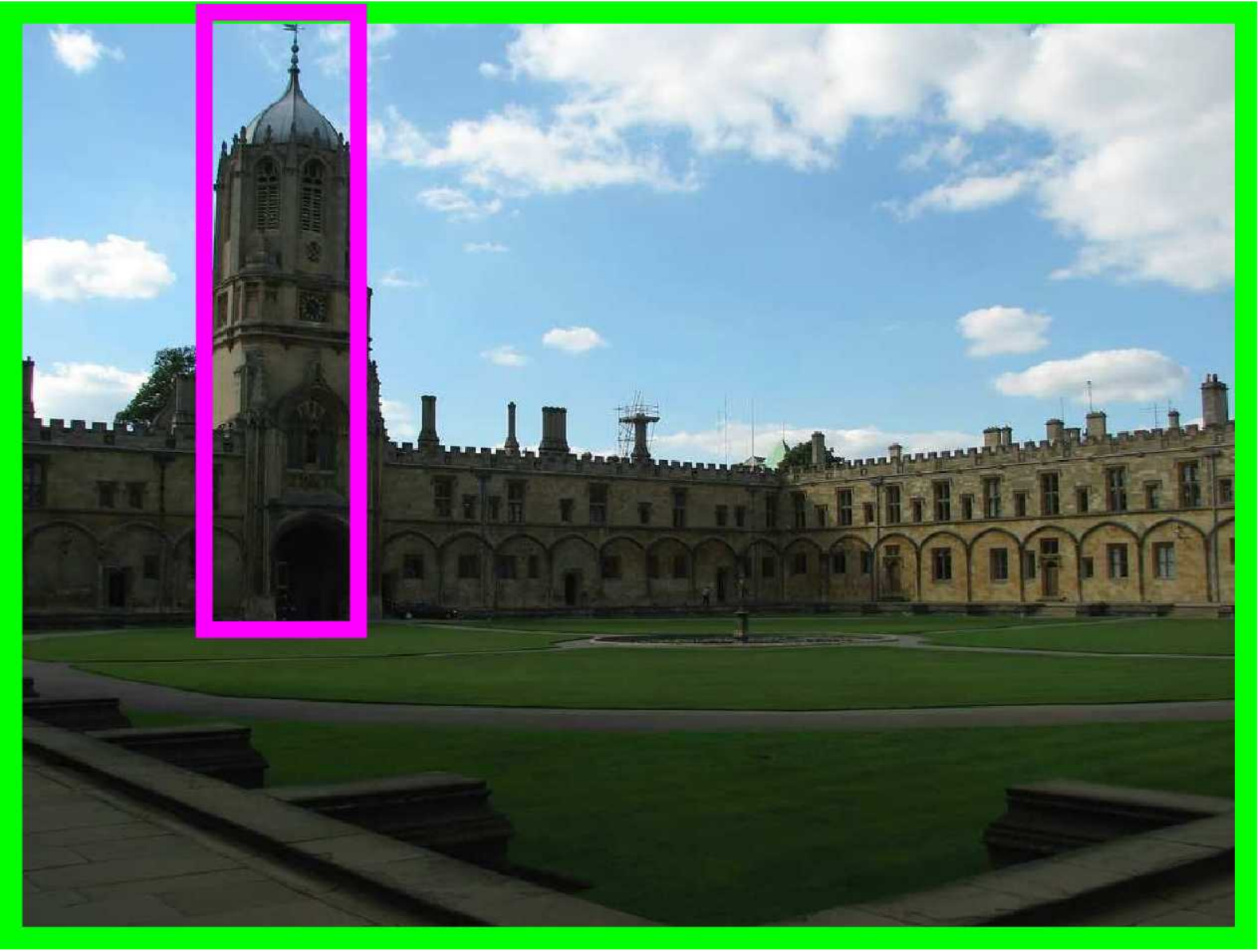}\\21 $\rightarrow$ 2\\ \end{tabular} 
\begin{tabular}{@{\sssp}c@{\sssp}}\includegraphics[height=\figh]{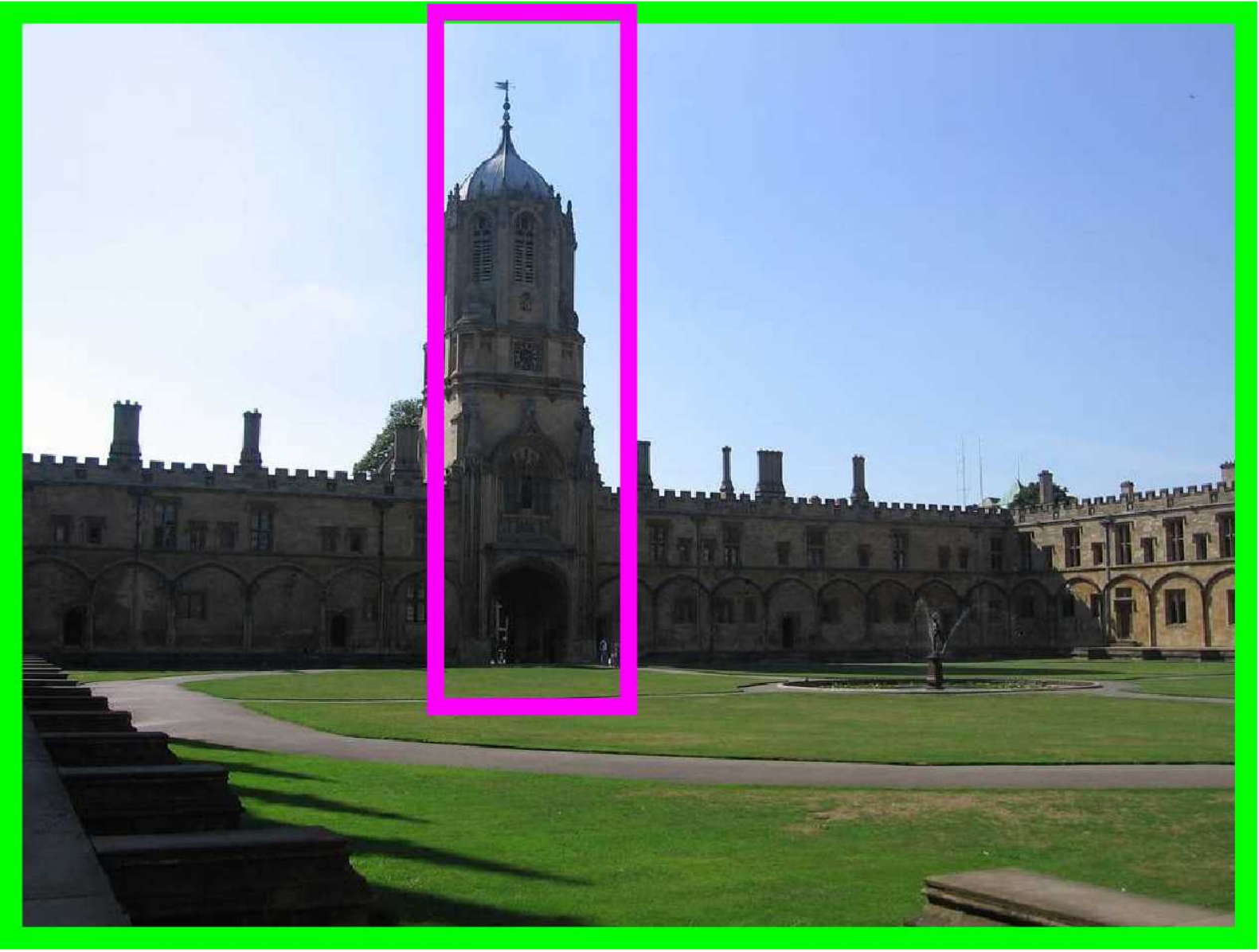}\\19 $\rightarrow$ 3\\ \end{tabular} 
\begin{tabular}{@{\sssp}c@{\sssp}}\includegraphics[height=\figh]{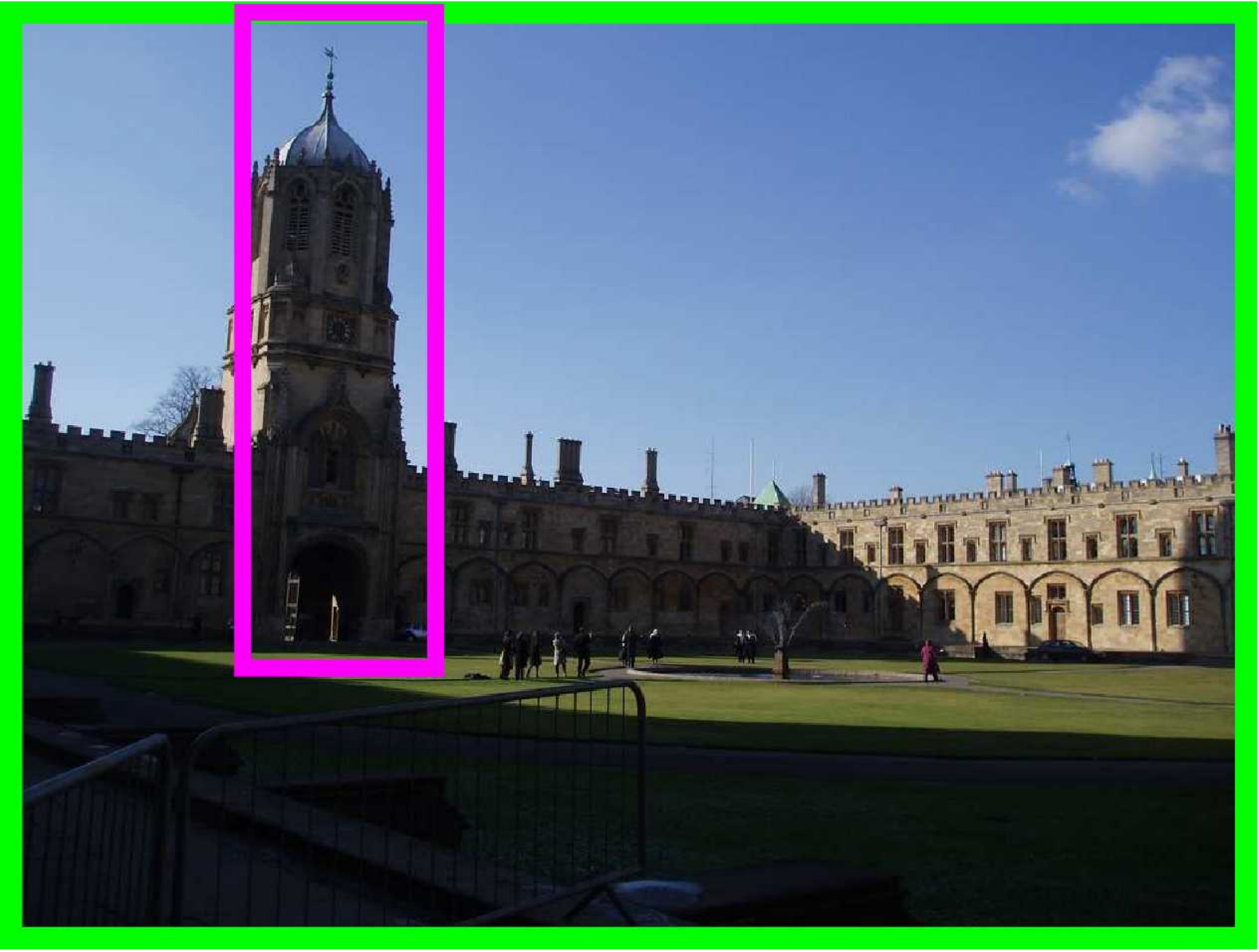}\\13 $\rightarrow$ 4\\ \end{tabular} 
\begin{tabular}{@{\sssp}c@{\sssp}}\includegraphics[height=\figh]{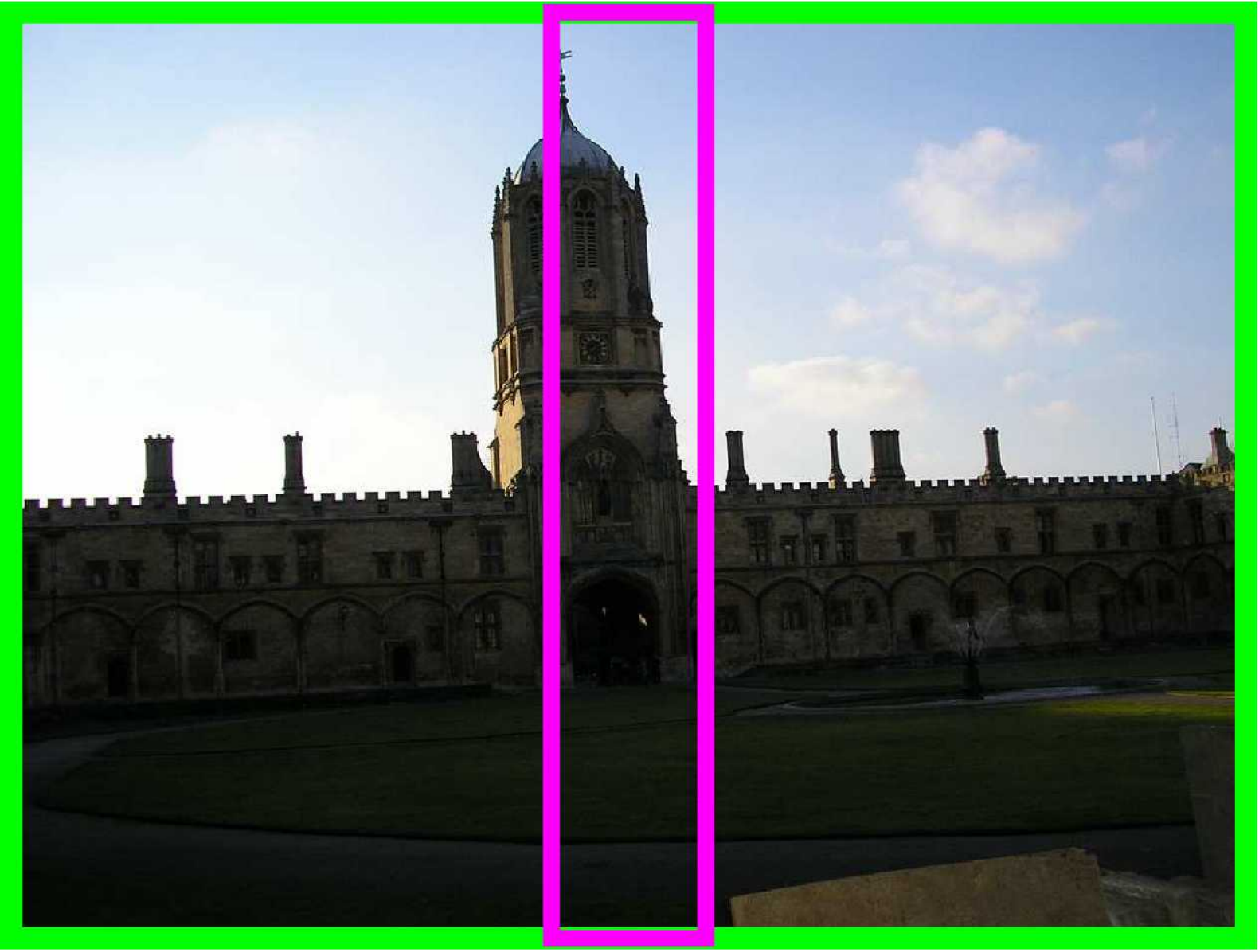}\\3 $\rightarrow$ 5\\ \end{tabular} 
\begin{tabular}{@{\sssp}c@{\sssp}}\includegraphics[height=\figh]{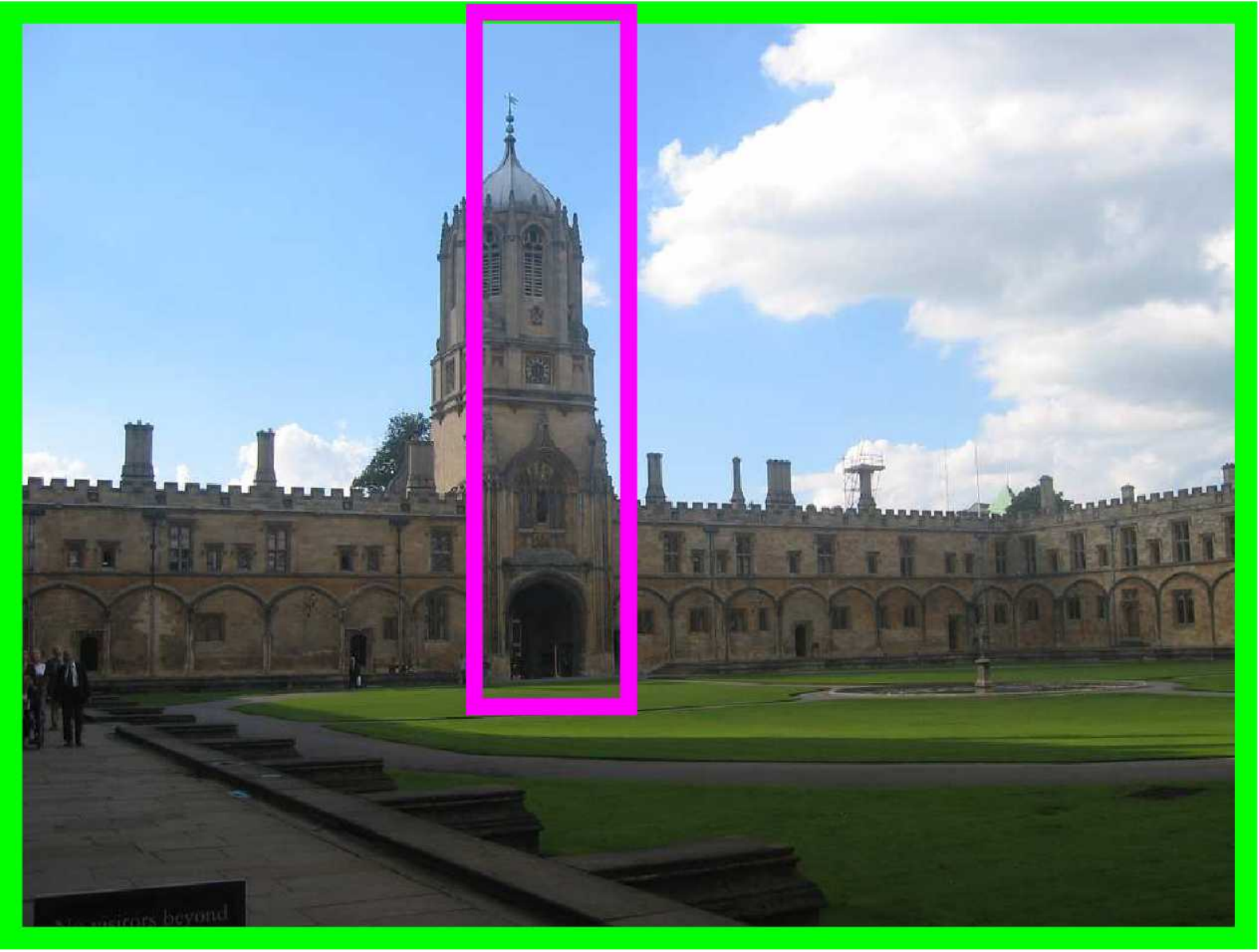}\\25 $\rightarrow$ 6\\ \end{tabular} 
\begin{tabular}{@{\sssp}c@{\sssp}}\includegraphics[height=\figh]{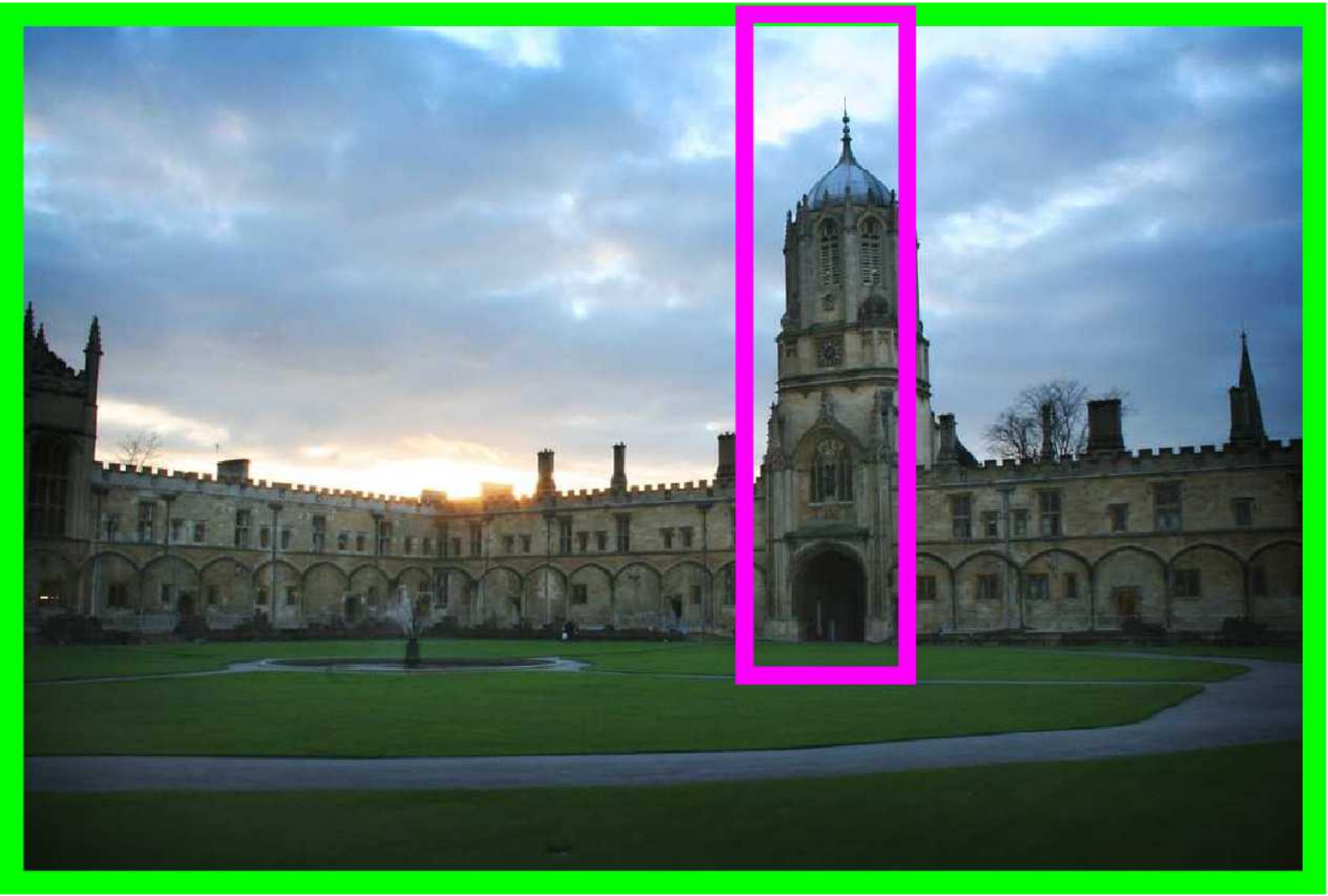}\\2 $\rightarrow$ 7\\ \end{tabular} 
\begin{tabular}{@{\sssp}c@{\sssp}}\includegraphics[height=\figh]{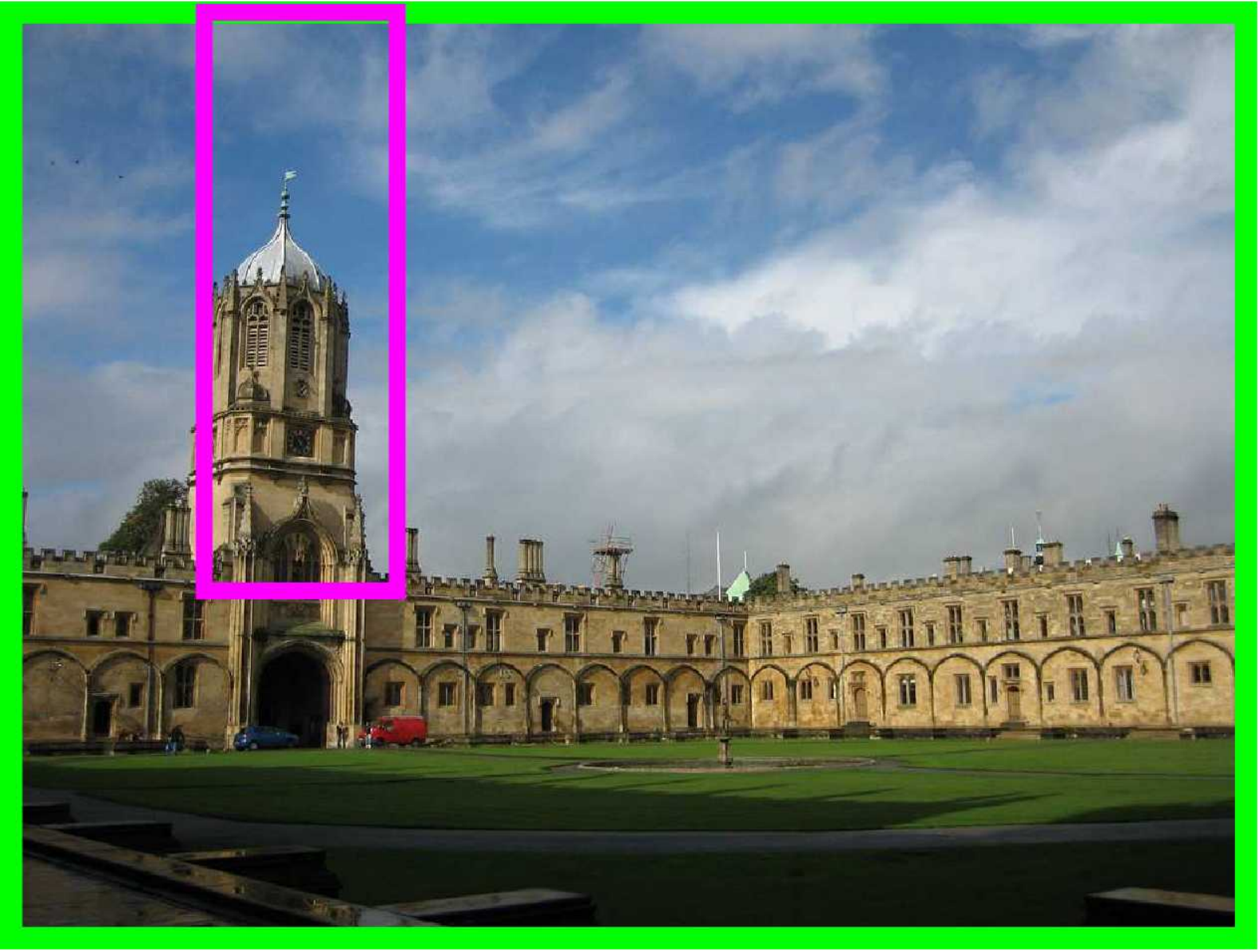}\\8 $\rightarrow$ 8\\ \end{tabular} 
  
\vspace{2ex}

\begin{tabular}{@{\sssp}c@{\sssp}}\includegraphics[height=\figh]{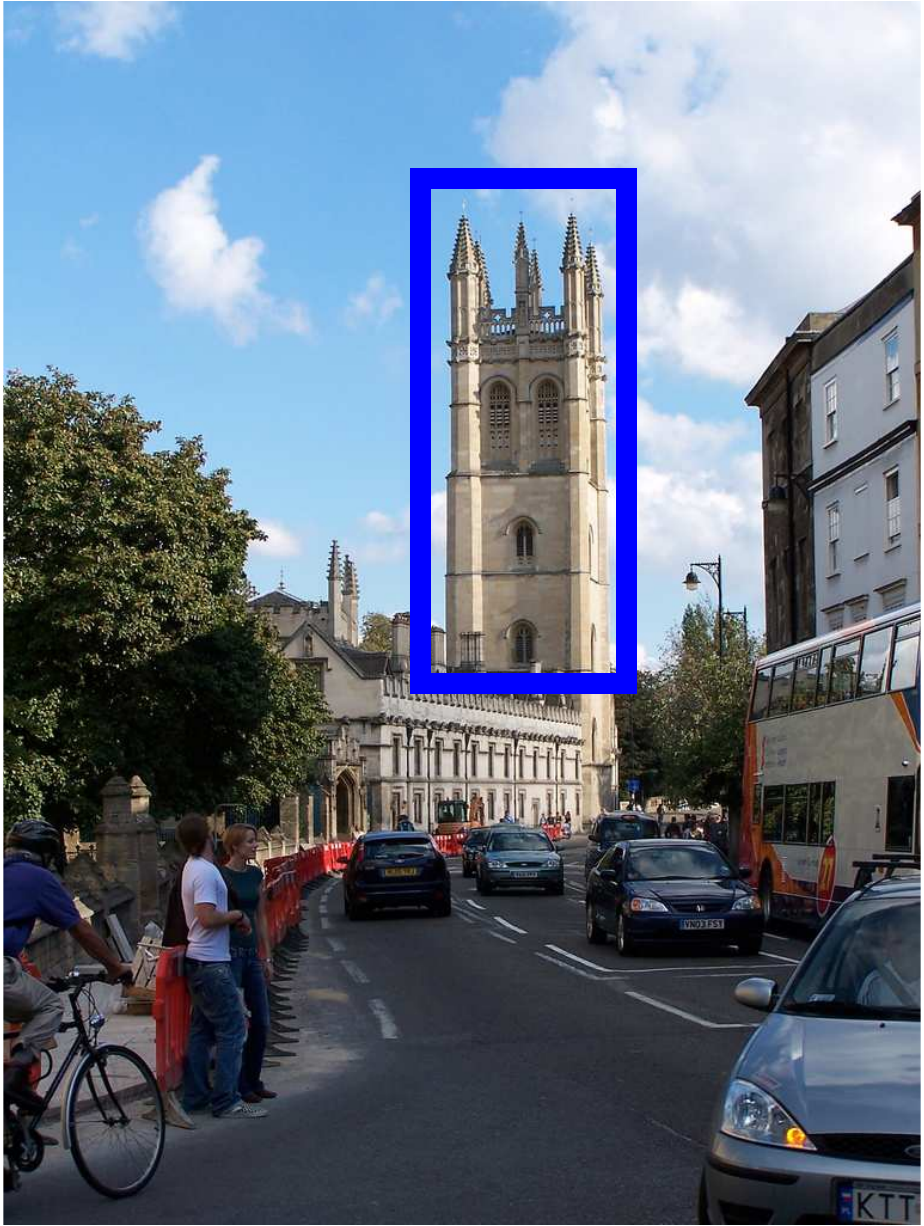}\end{tabular} 
\begin{tabular}{@{\sssp}c@{\sssp}}\includegraphics[height=\figh]{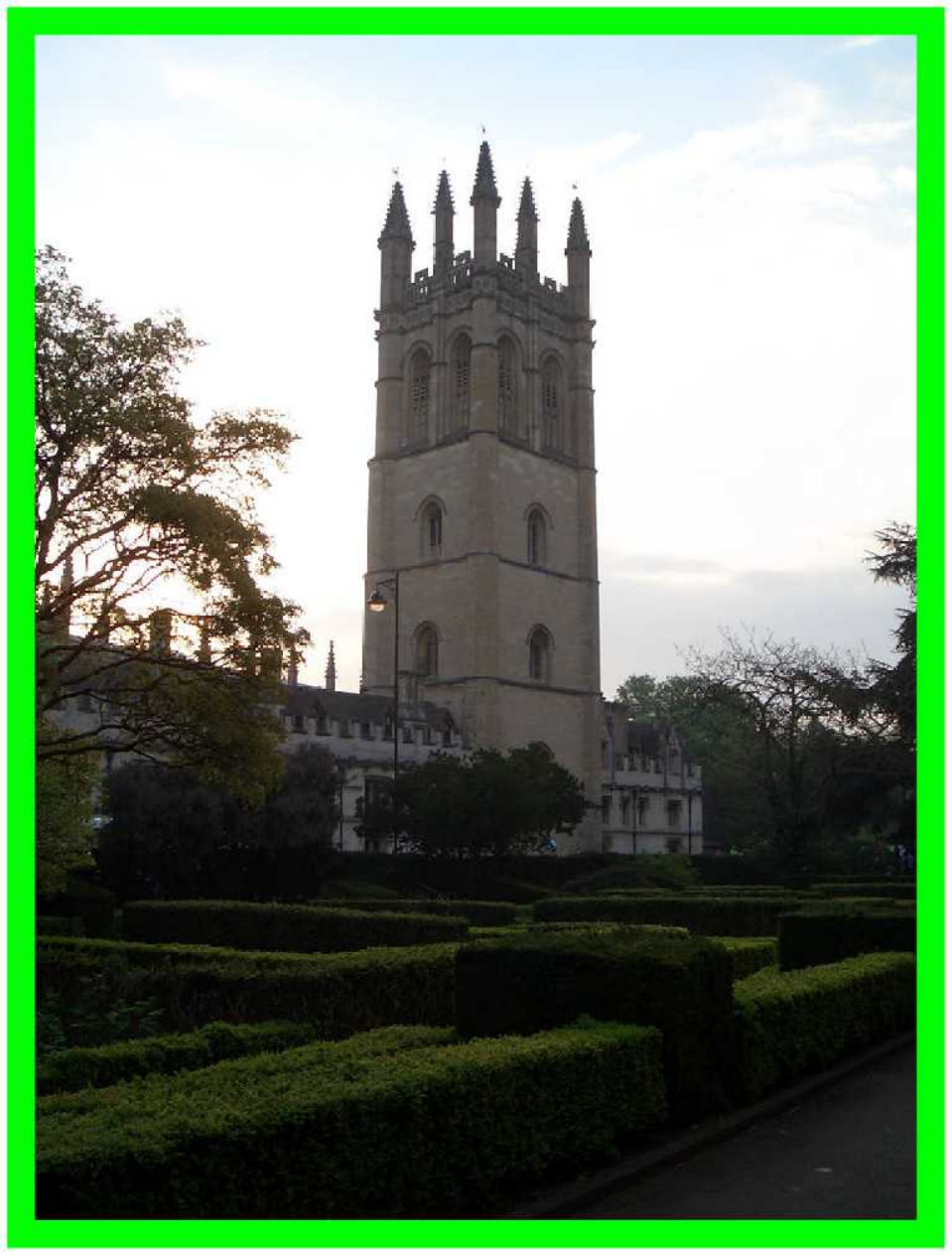}\end{tabular}
\begin{tabular}{@{\sssp}c@{\sssp}}\includegraphics[height=\figh]{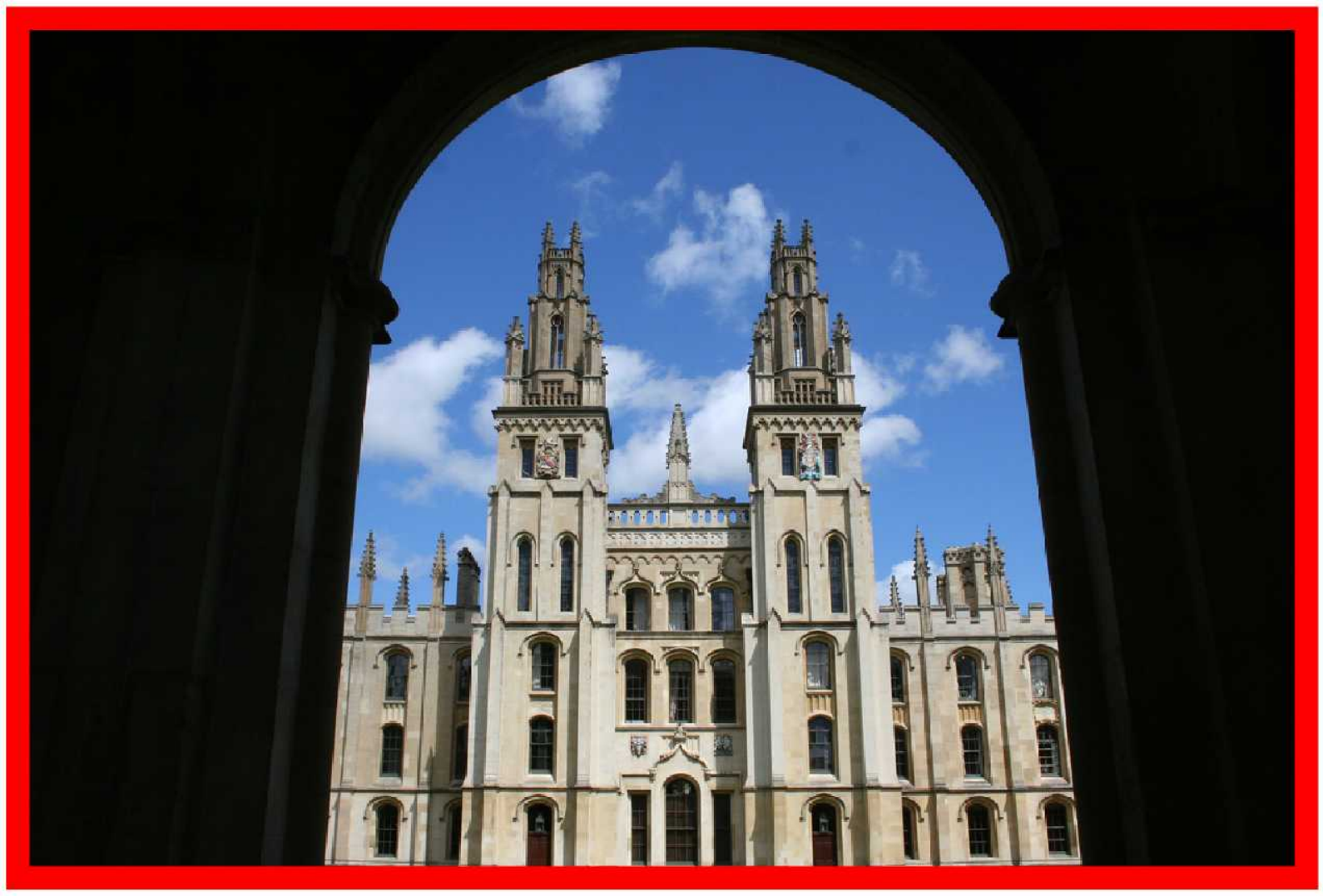}\end{tabular}
\begin{tabular}{@{\sssp}c@{\sssp}}\includegraphics[height=\figh]{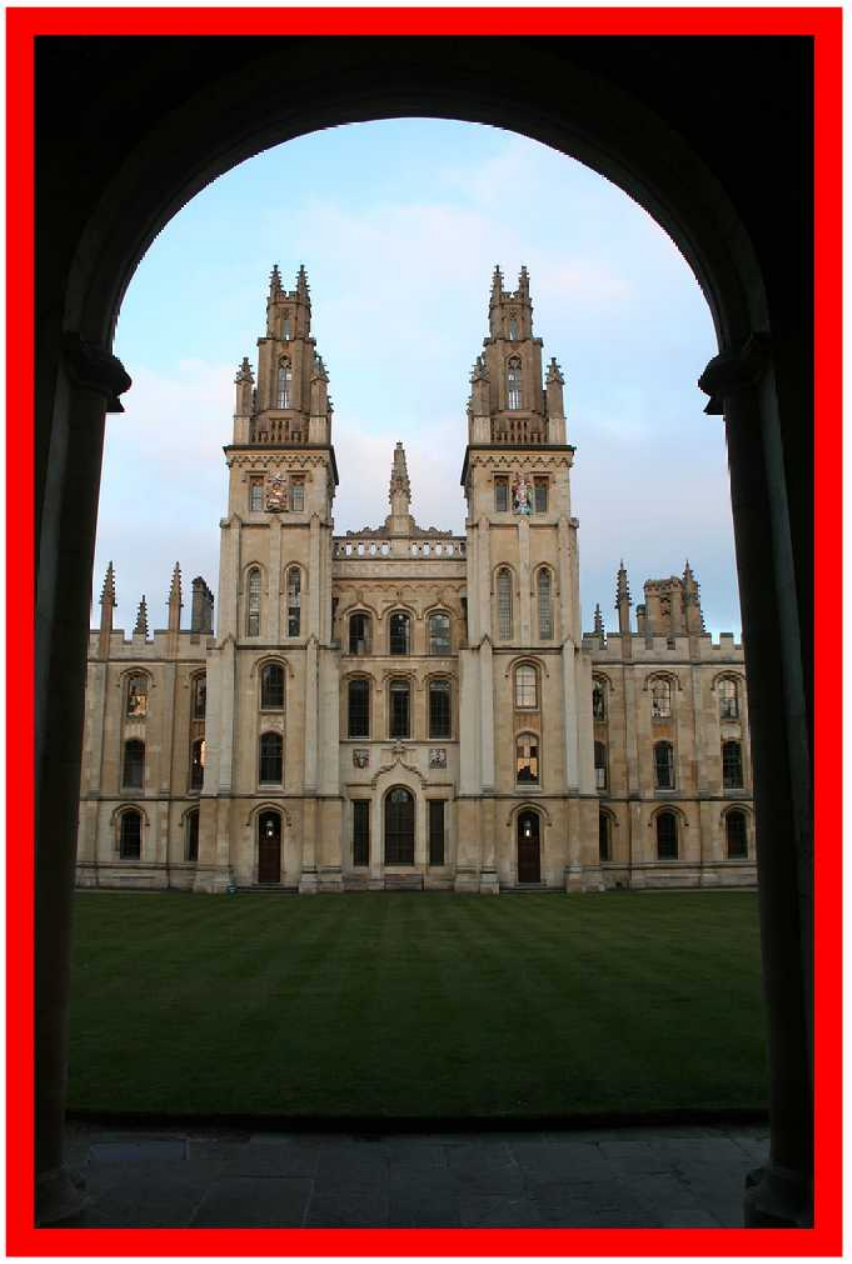}\end{tabular}
\begin{tabular}{@{\sssp}c@{\sssp}}\includegraphics[height=\figh]{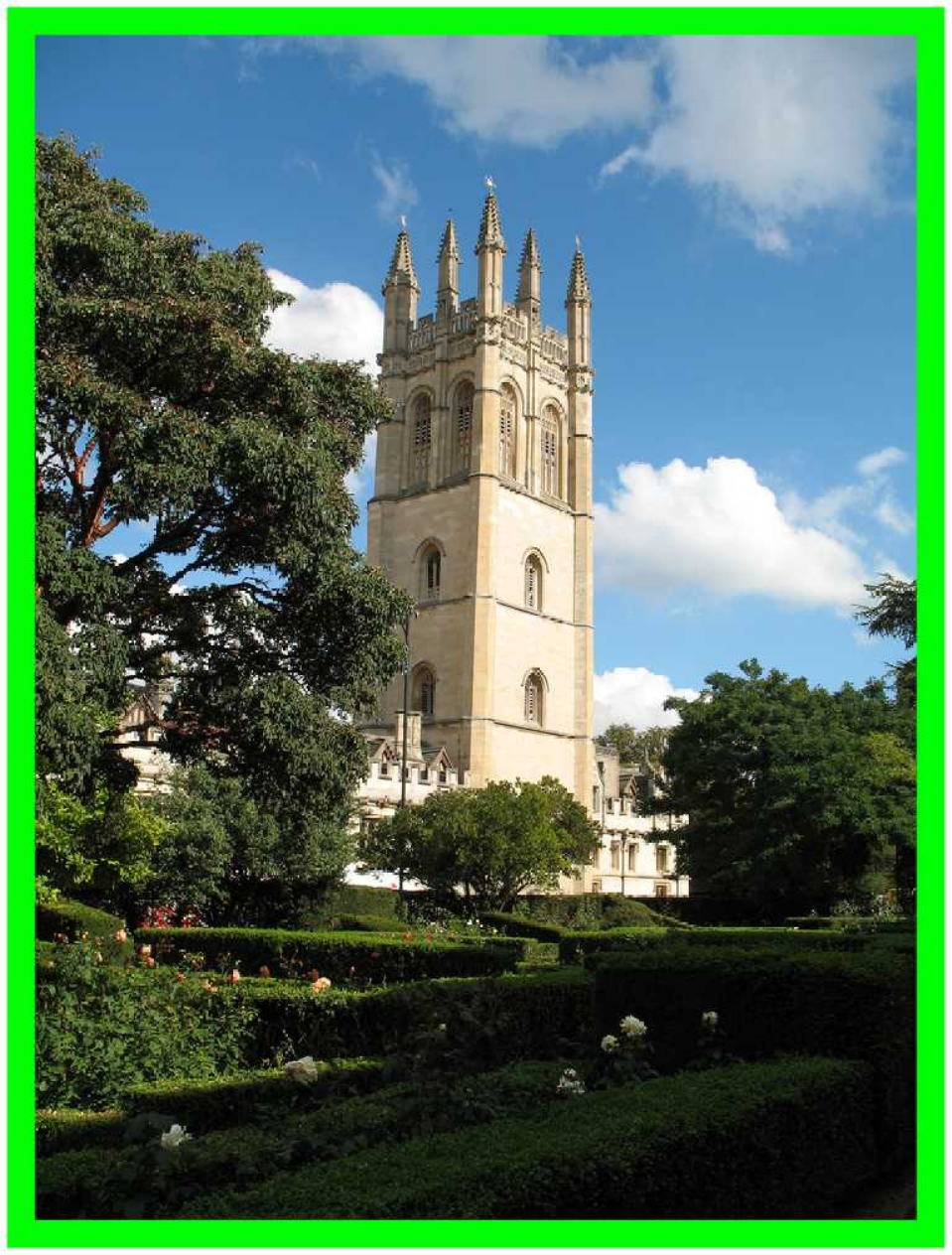}\end{tabular}
\begin{tabular}{@{\sssp}c@{\sssp}}\includegraphics[height=\figh]{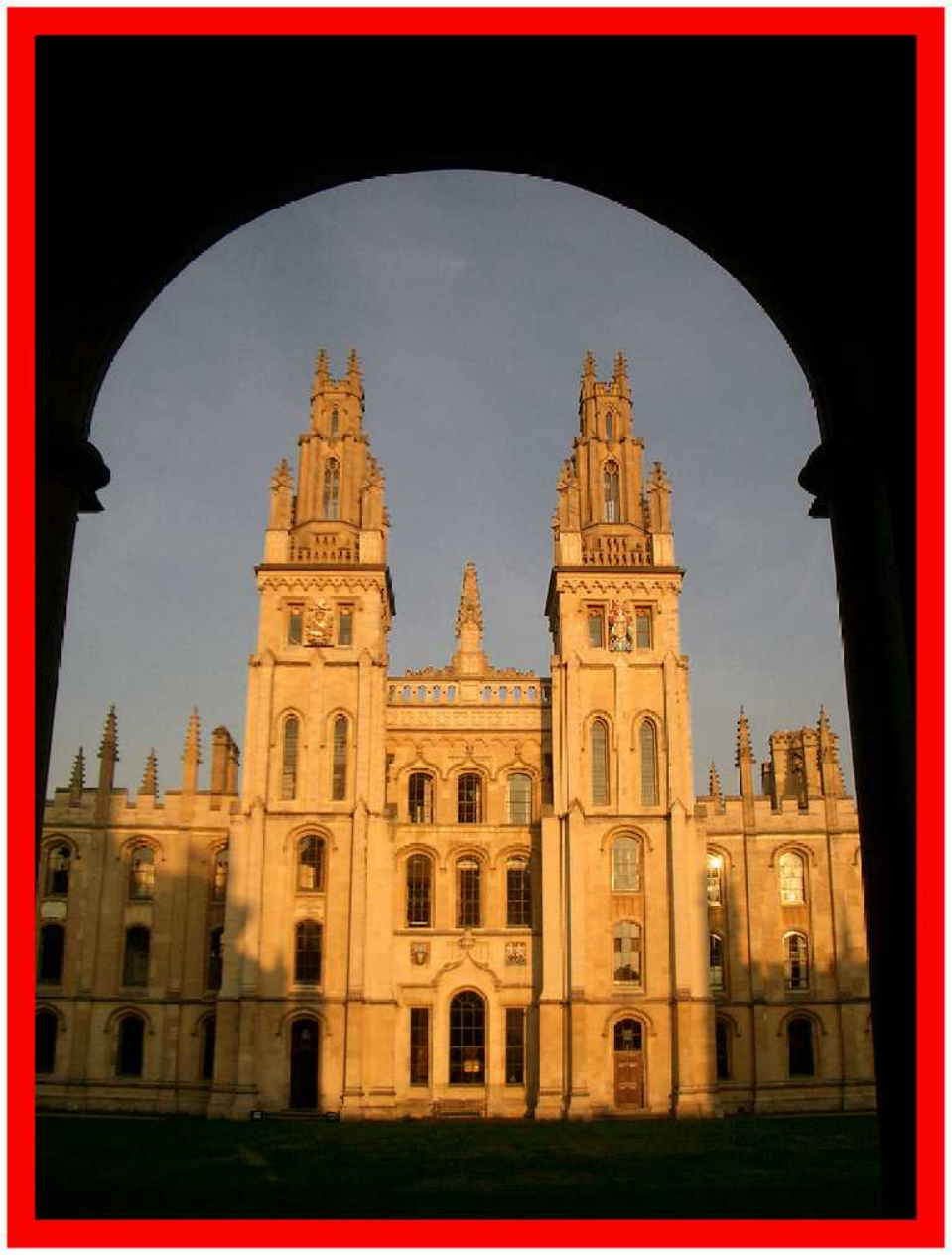}\end{tabular}
\begin{tabular}{@{\sssp}c@{\sssp}}\includegraphics[height=\figh]{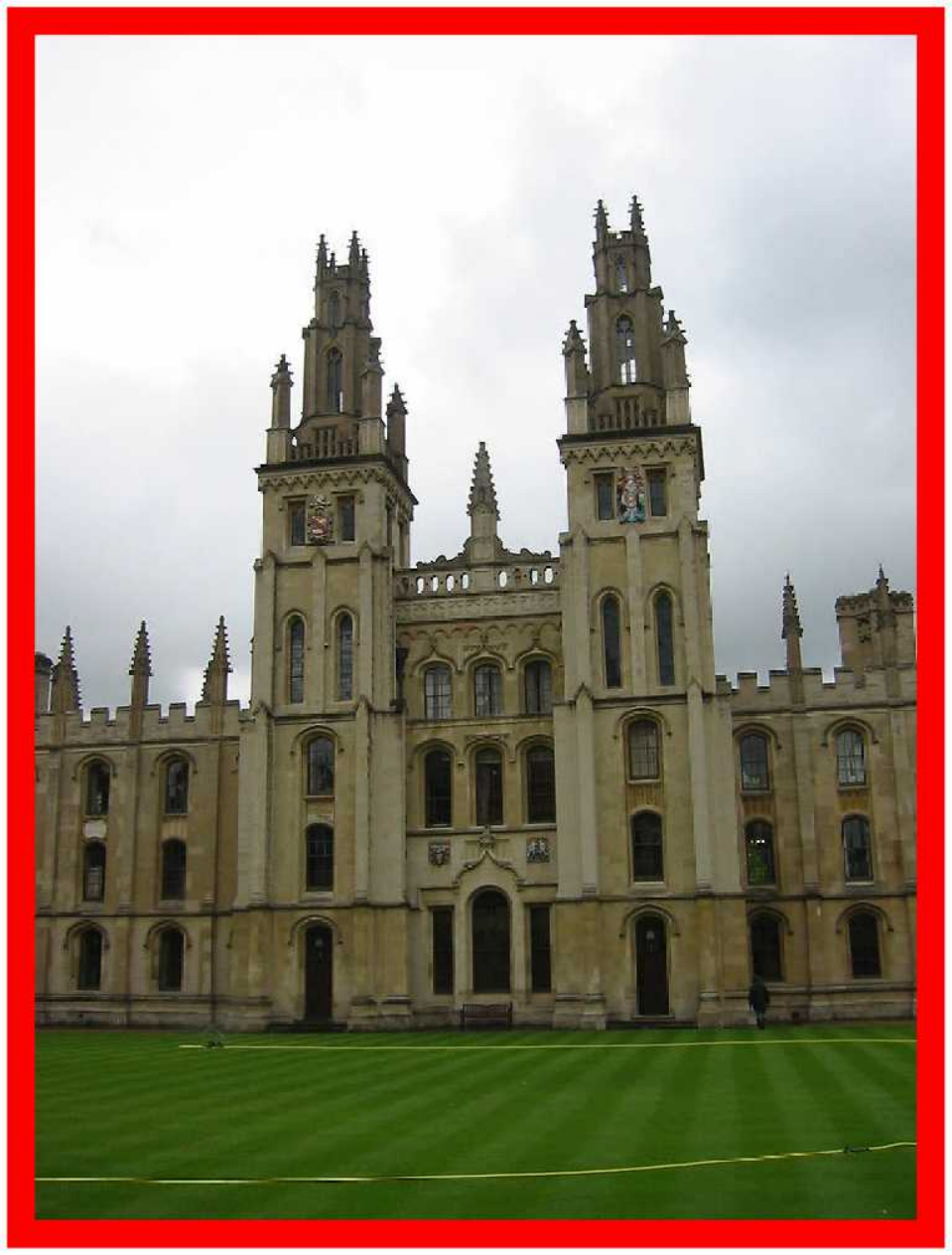}\end{tabular}
\begin{tabular}{@{\sssp}c@{\sssp}}\includegraphics[height=\figh]{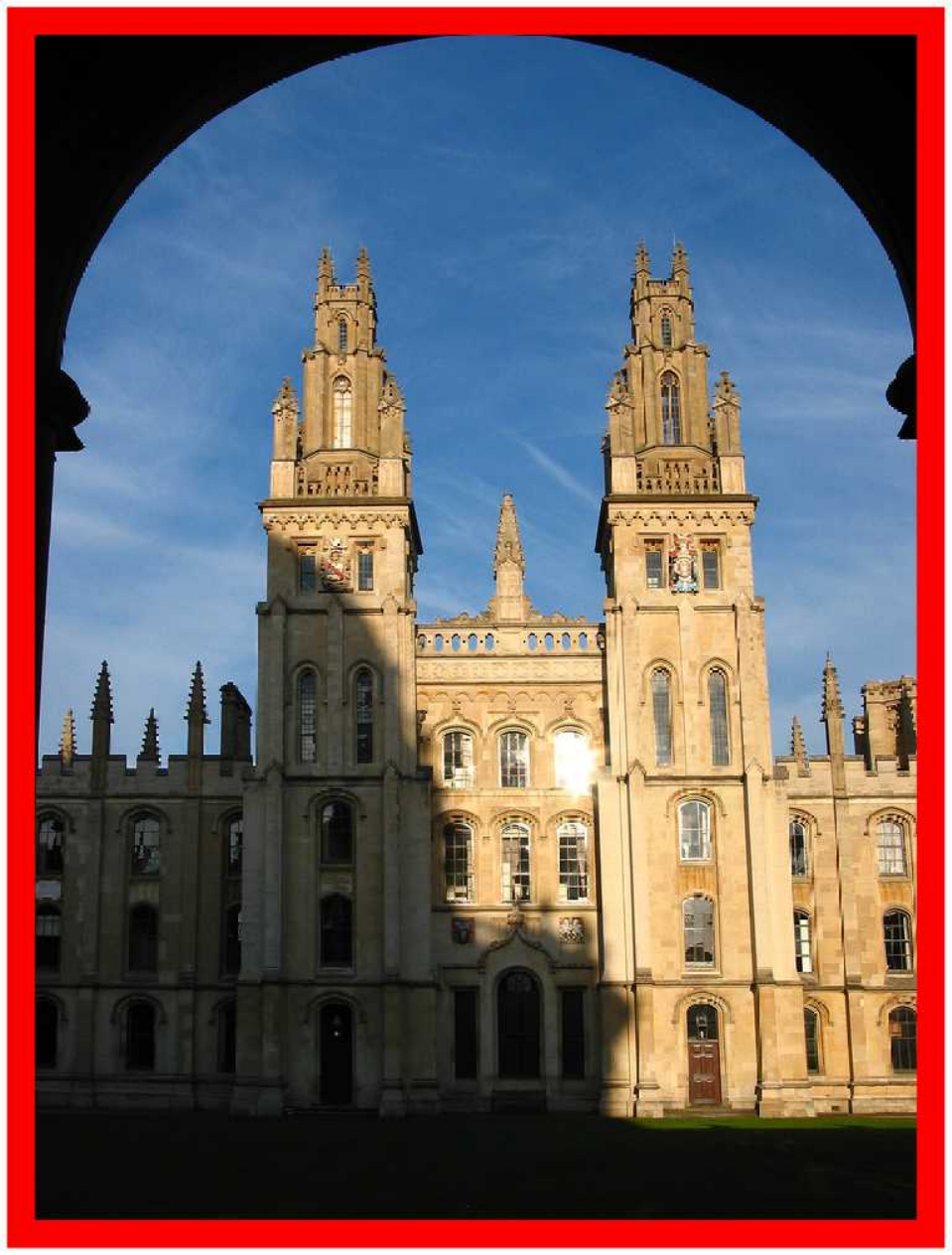}\end{tabular}
\begin{tabular}{@{\sssp}c@{\sssp}}\includegraphics[height=\figh]{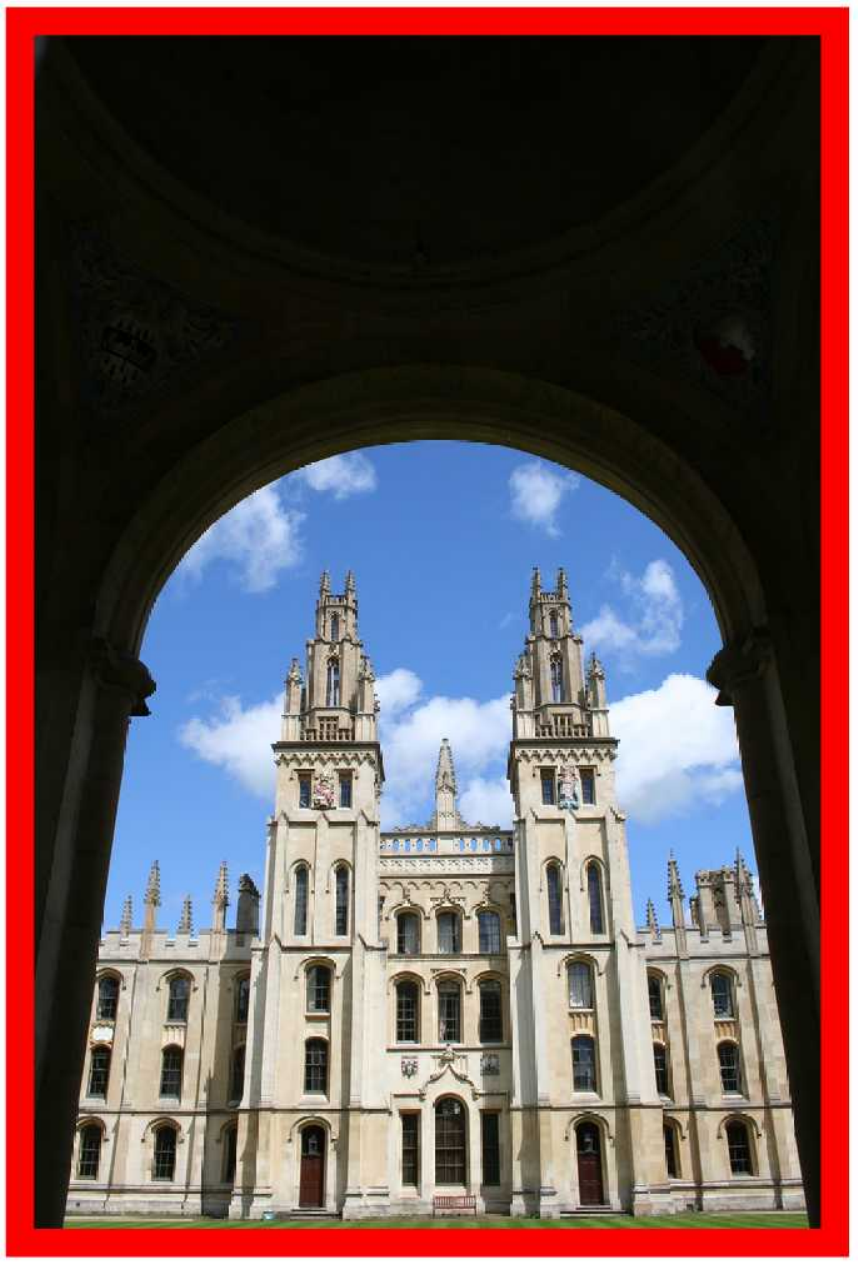}\end{tabular}
\begin{tabular}{@{\sssp}c@{\sssp}}\includegraphics[height=\figh]{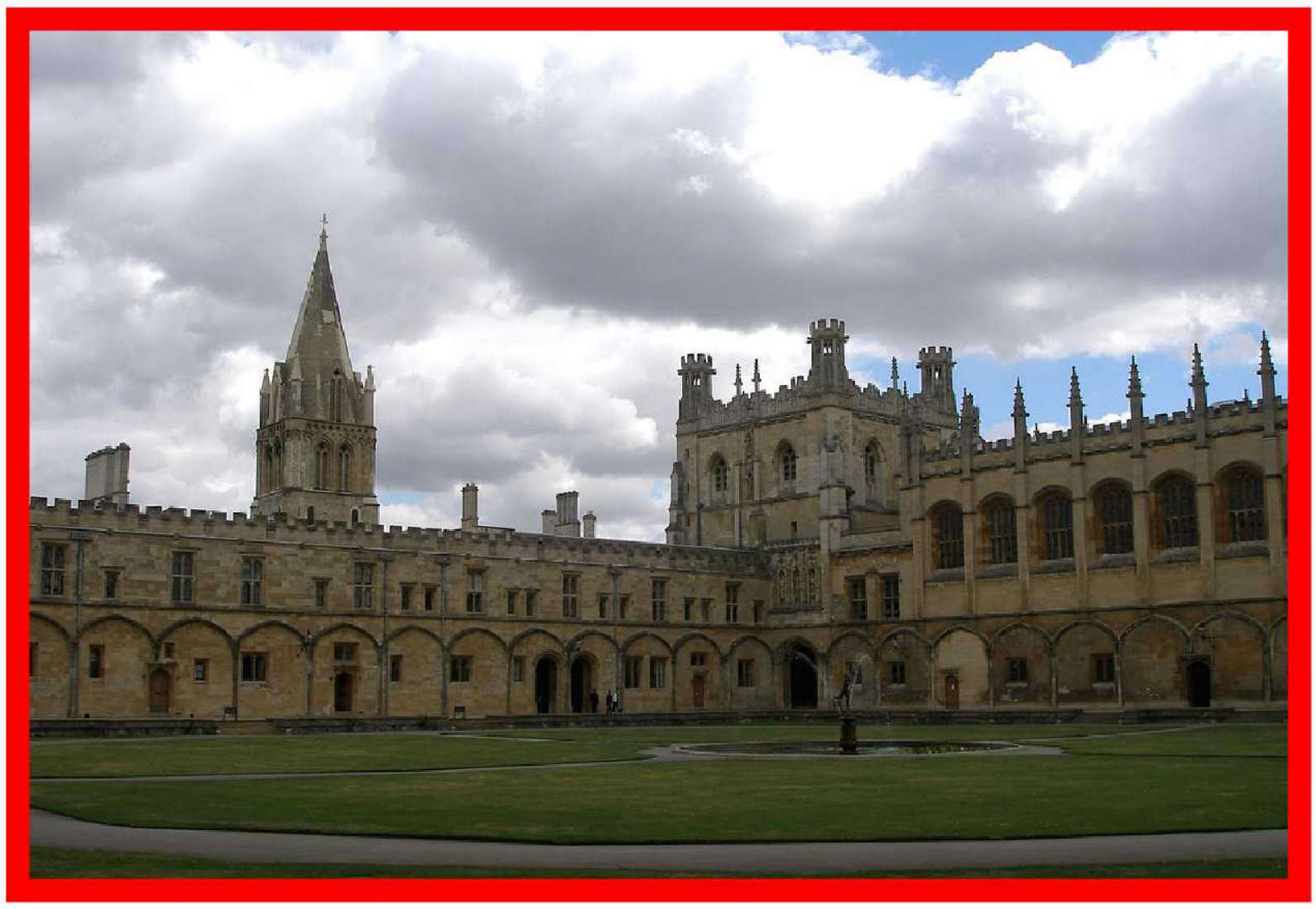}\end{tabular}
\begin{tabular}{@{\sssp}c@{\sssp}}\includegraphics[height=\figh]{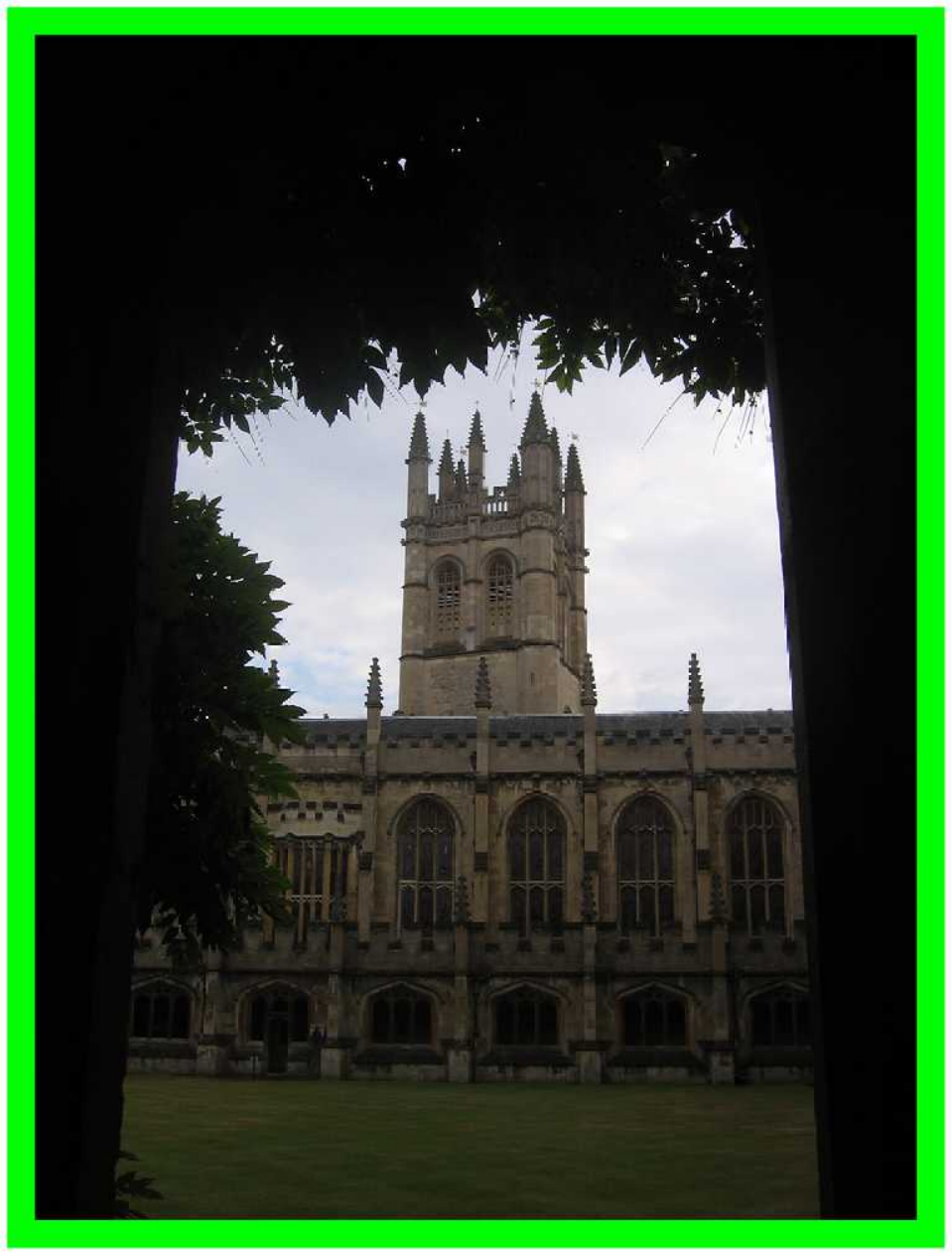}\end{tabular}
\begin{tabular}{@{\sssp}c@{\sssp}}\includegraphics[height=\figh]{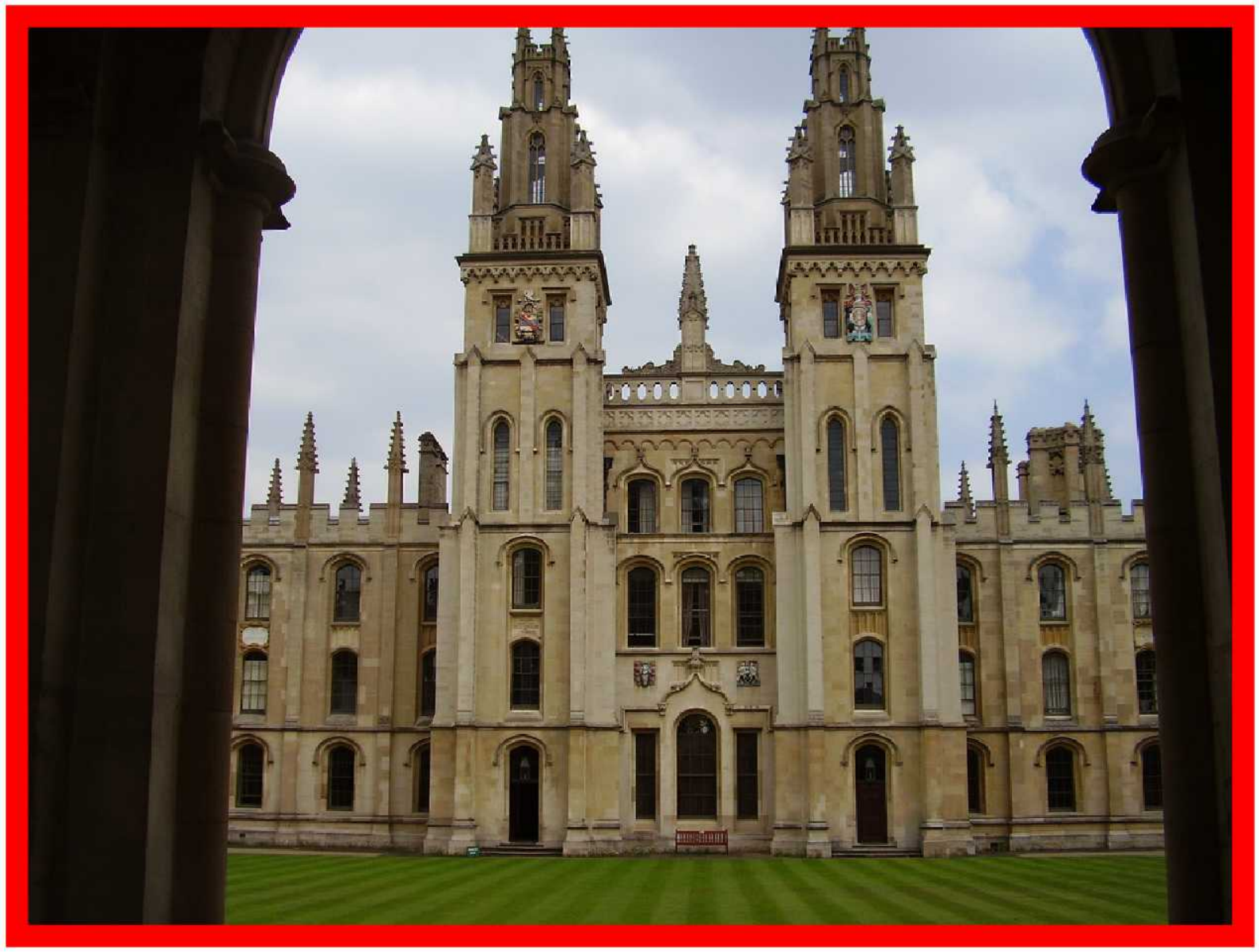}\end{tabular}

\begin{tabular}{@{\sssp}c@{\sssp}}\includegraphics[height=\figh]{figs/rerank/42//q42_0932_2625.pdf}\\Query\\ \end{tabular} 
\begin{tabular}{@{\sssp}c@{\sssp}}\includegraphics[height=\figh]{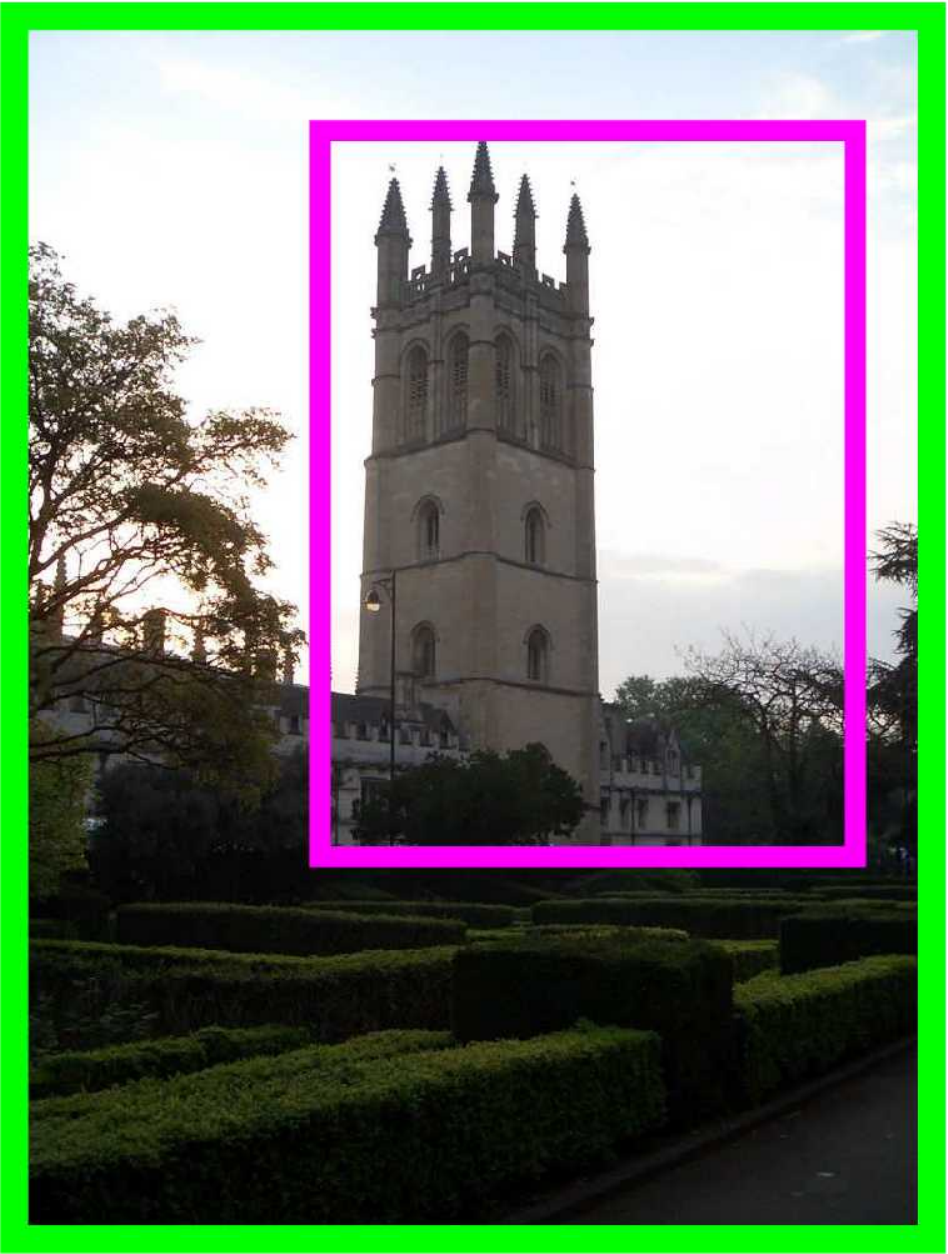}\\1 $\rightarrow$ 1\\ \end{tabular} 
\begin{tabular}{@{\sssp}c@{\sssp}}\includegraphics[height=\figh]{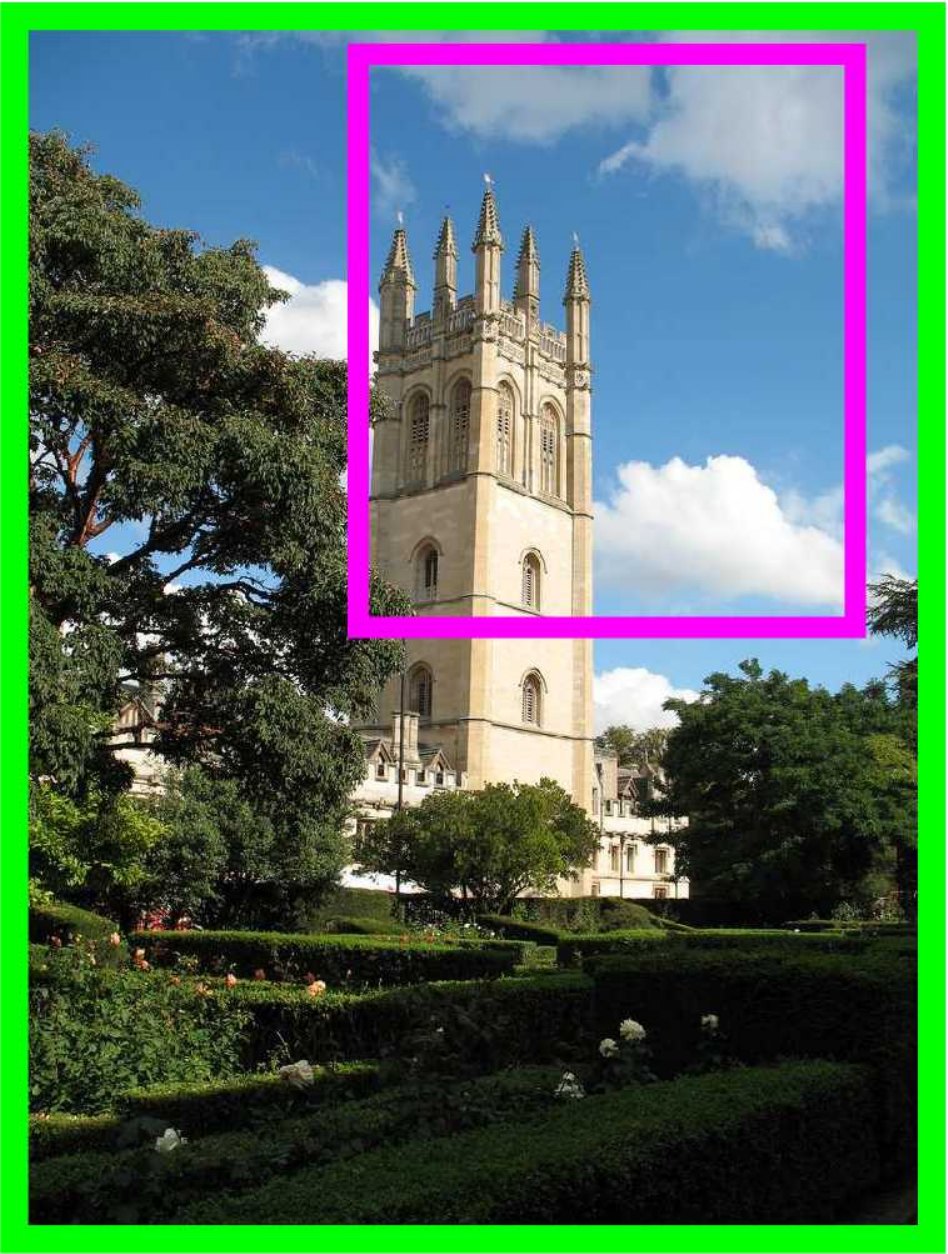}\\4 $\rightarrow$ 2\\ \end{tabular} 
\begin{tabular}{@{\sssp}c@{\sssp}}\includegraphics[height=\figh]{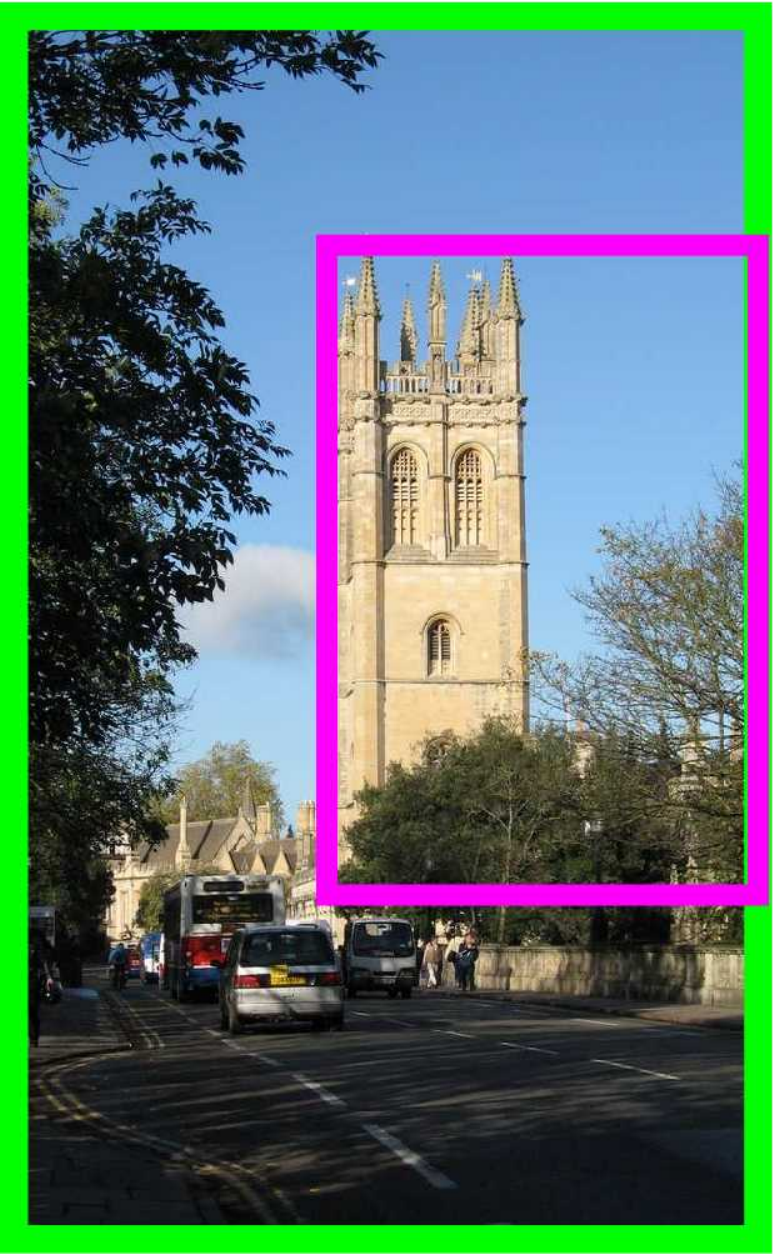}\\220 $\rightarrow$ 3\\ \end{tabular} 
\begin{tabular}{@{\sssp}c@{\sssp}}\includegraphics[height=\figh]{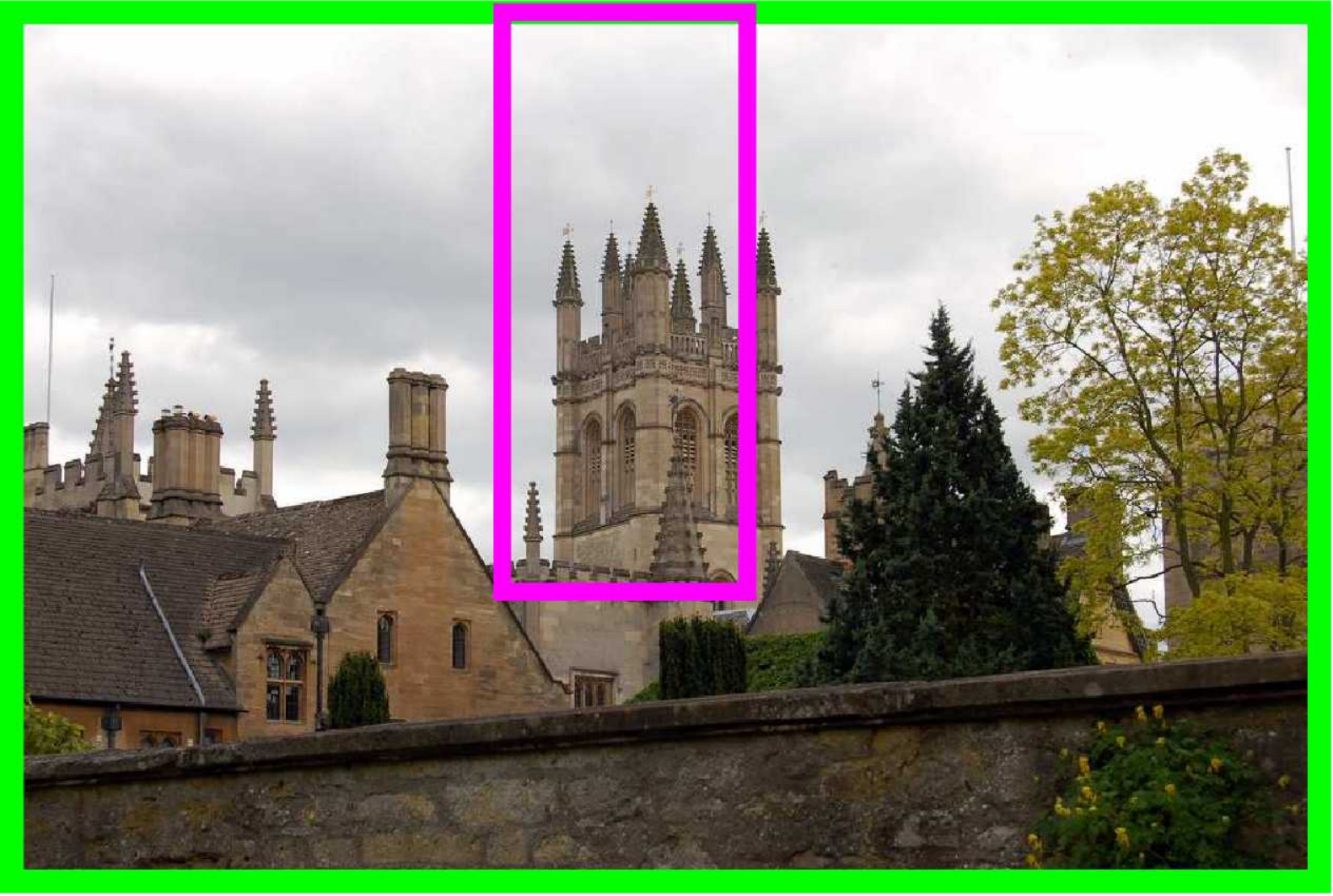}\\52 $\rightarrow$ 4\\ \end{tabular} 
\begin{tabular}{@{\sssp}c@{\sssp}}\includegraphics[height=\figh]{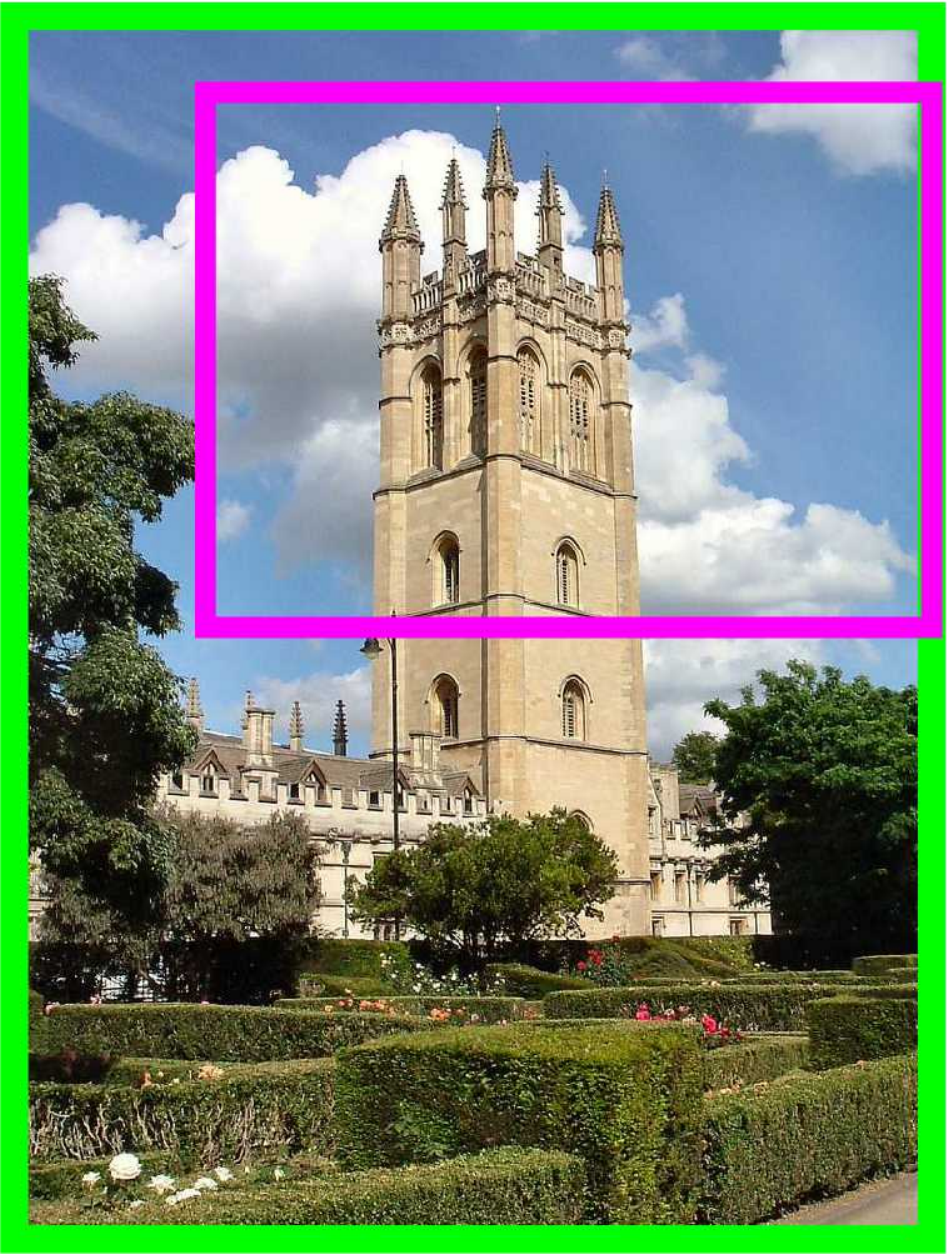}\\15 $\rightarrow$ 5\\ \end{tabular} 
\begin{tabular}{@{\sssp}c@{\sssp}}\includegraphics[height=\figh]{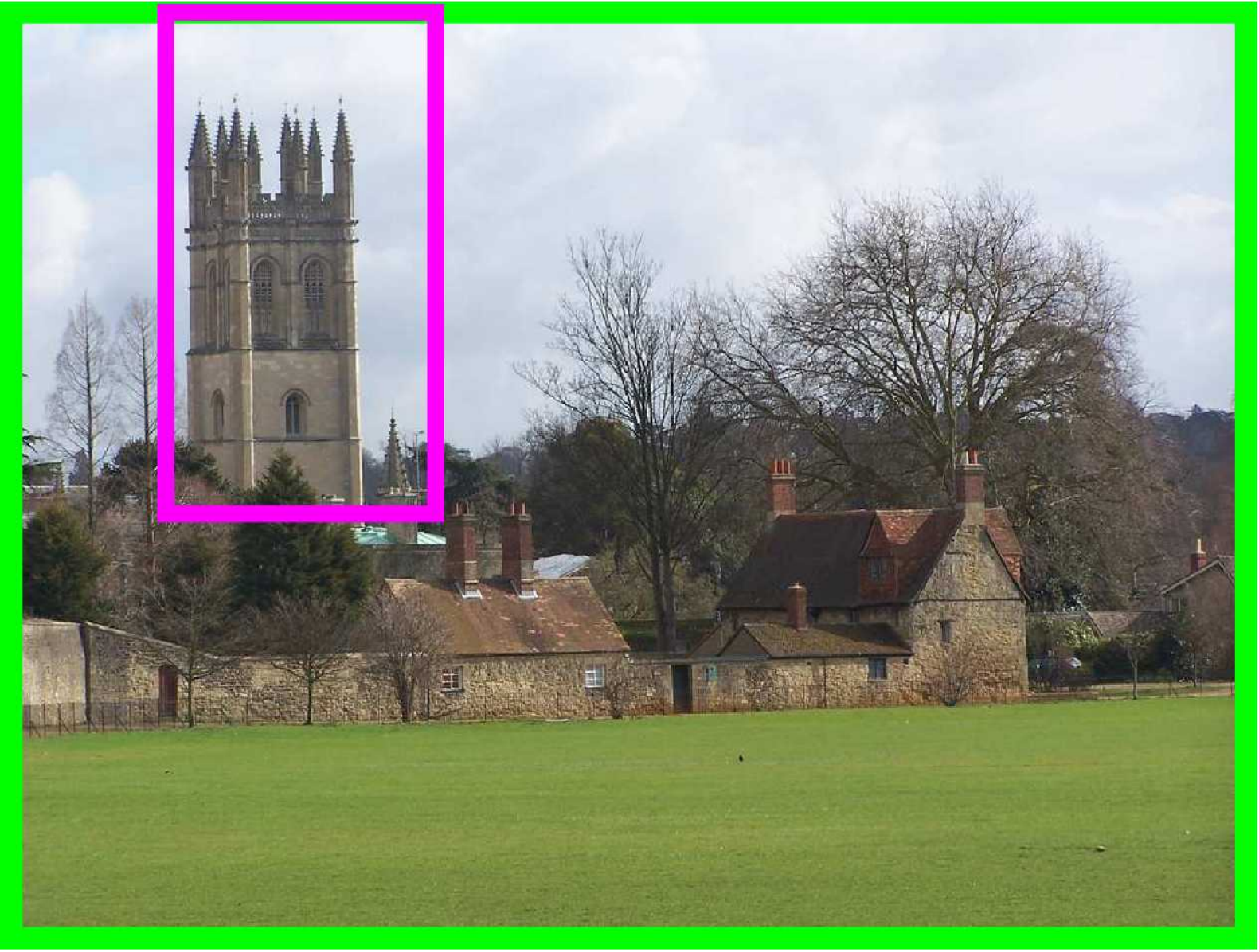}\\212 $\rightarrow$ 6\\ \end{tabular} 
\begin{tabular}{@{\sssp}c@{\sssp}}\includegraphics[height=\figh]{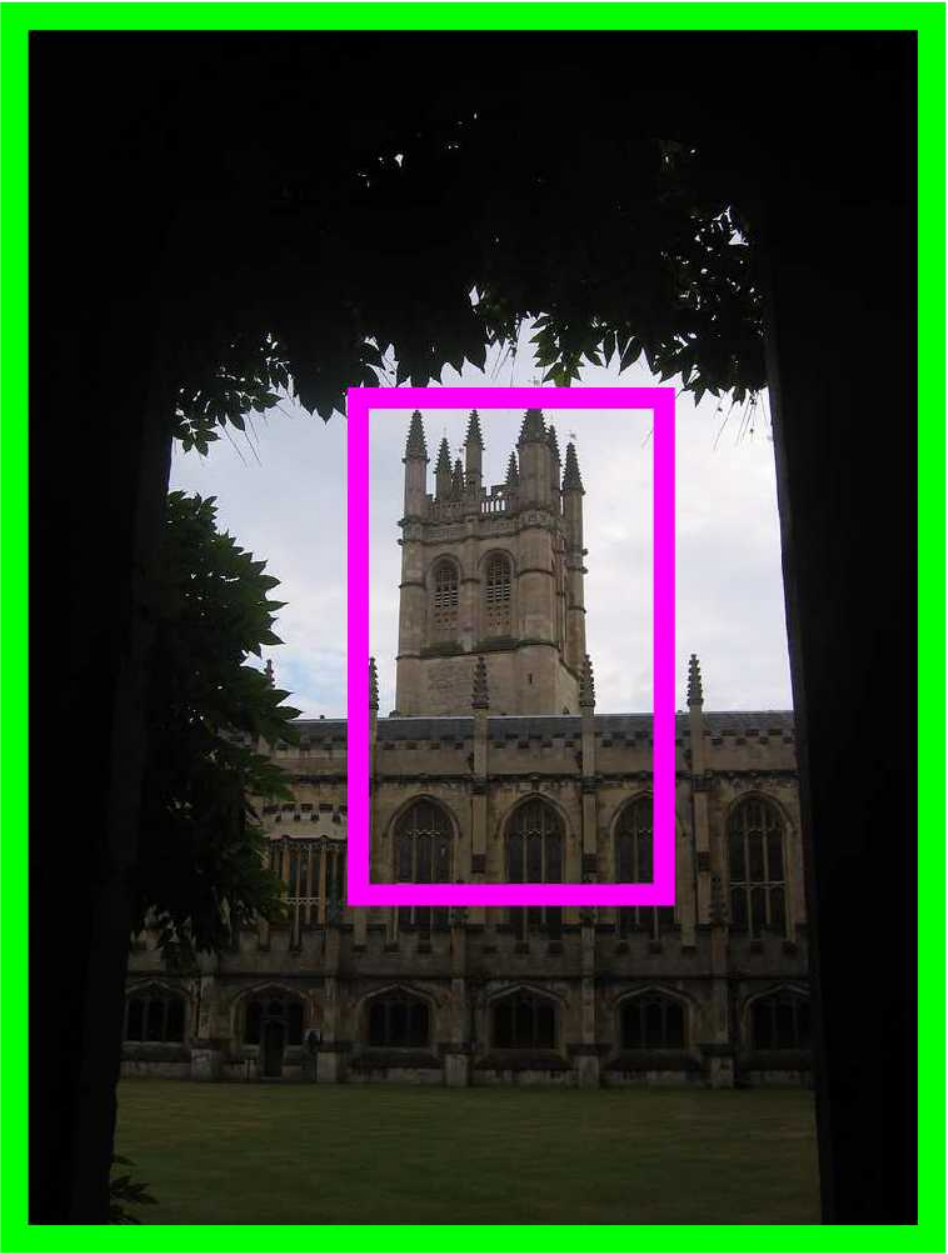}\\10 $\rightarrow$ 7\\ \end{tabular} 
\begin{tabular}{@{\sssp}c@{\sssp}}\includegraphics[height=\figh]{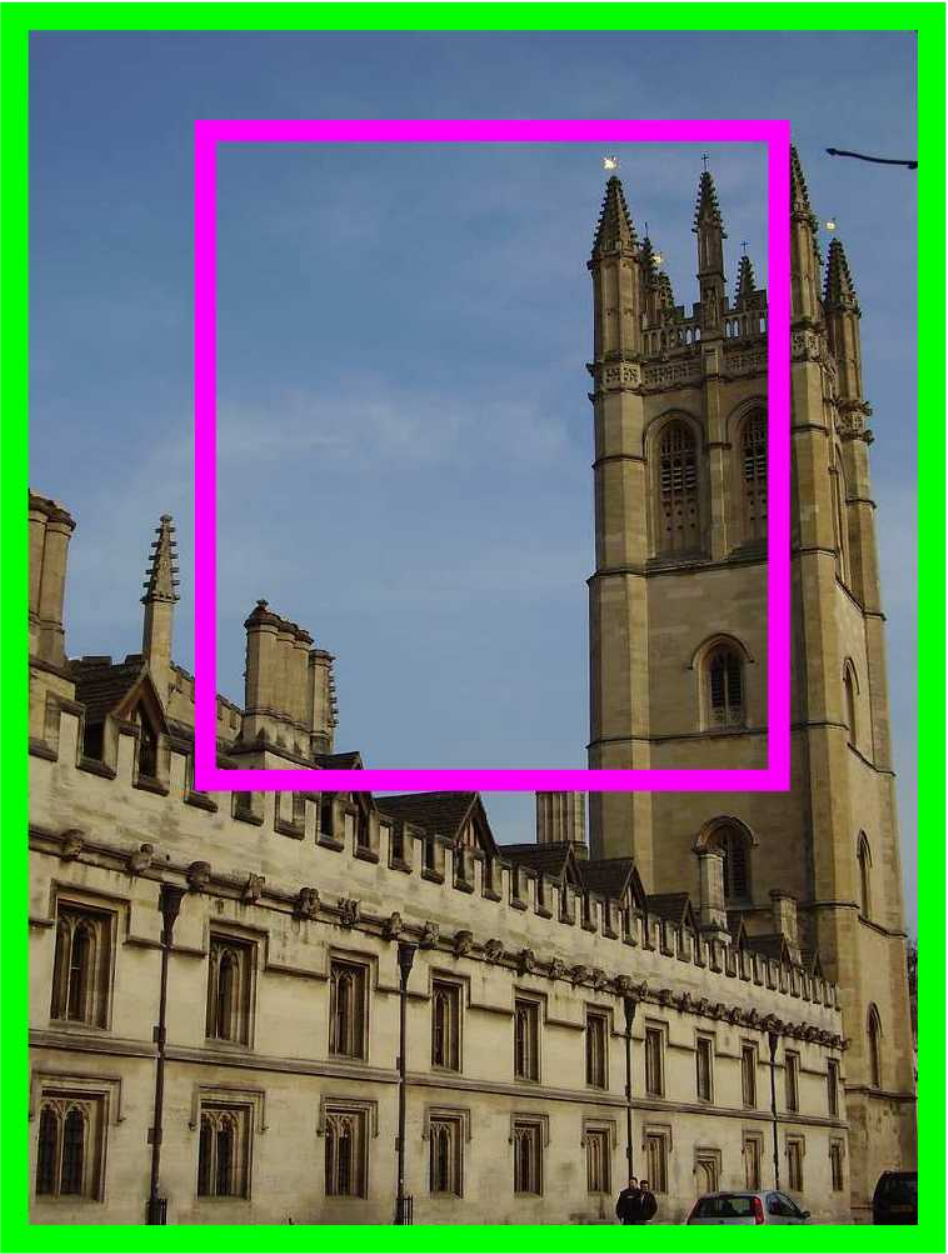}\\159 $\rightarrow$ 8\\ \end{tabular} 
\begin{tabular}{@{\sssp}c@{\sssp}}\includegraphics[height=\figh]{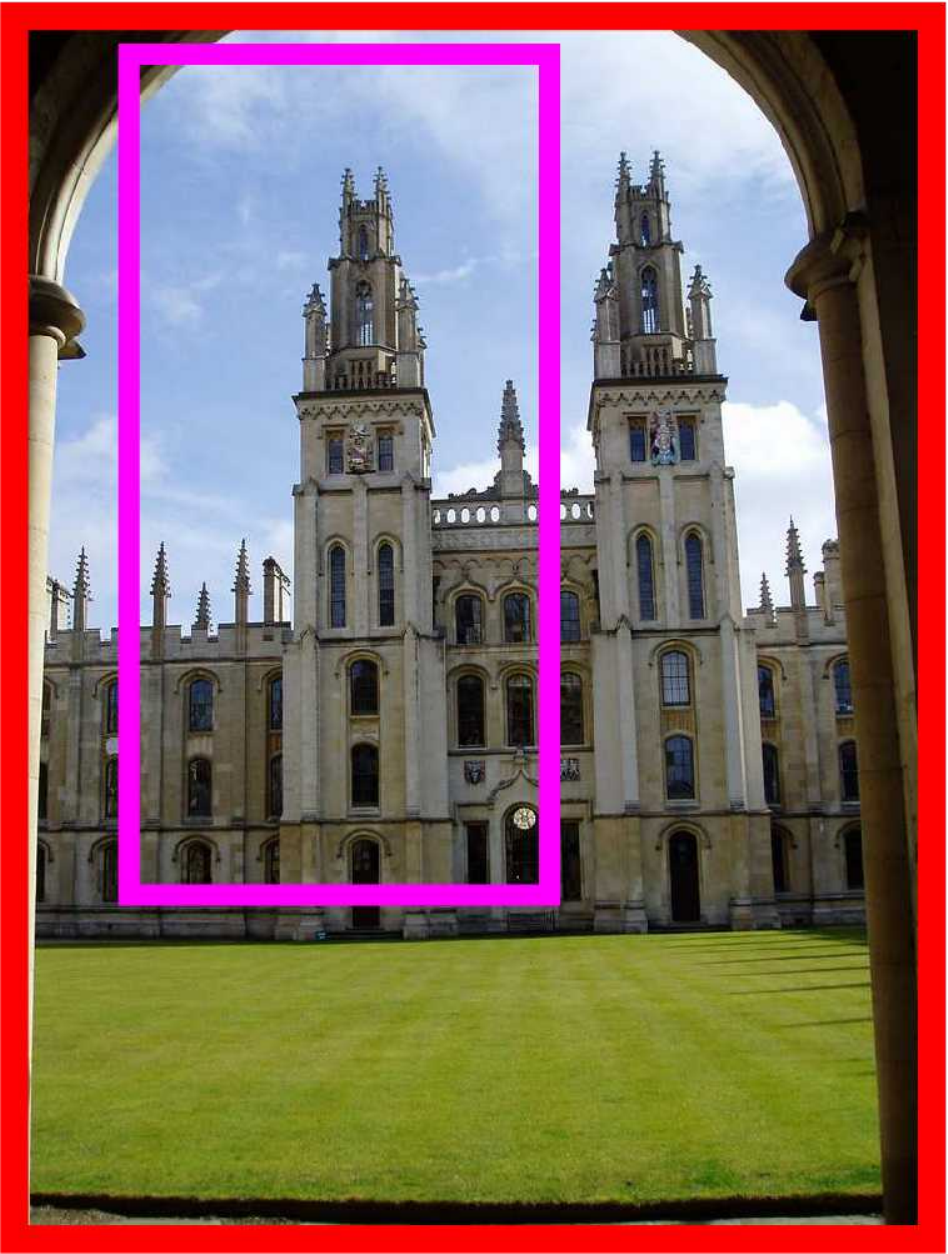}\\26 $\rightarrow$ 9\\ \end{tabular} 
\begin{tabular}{@{\sssp}c@{\sssp}}\includegraphics[height=\figh]{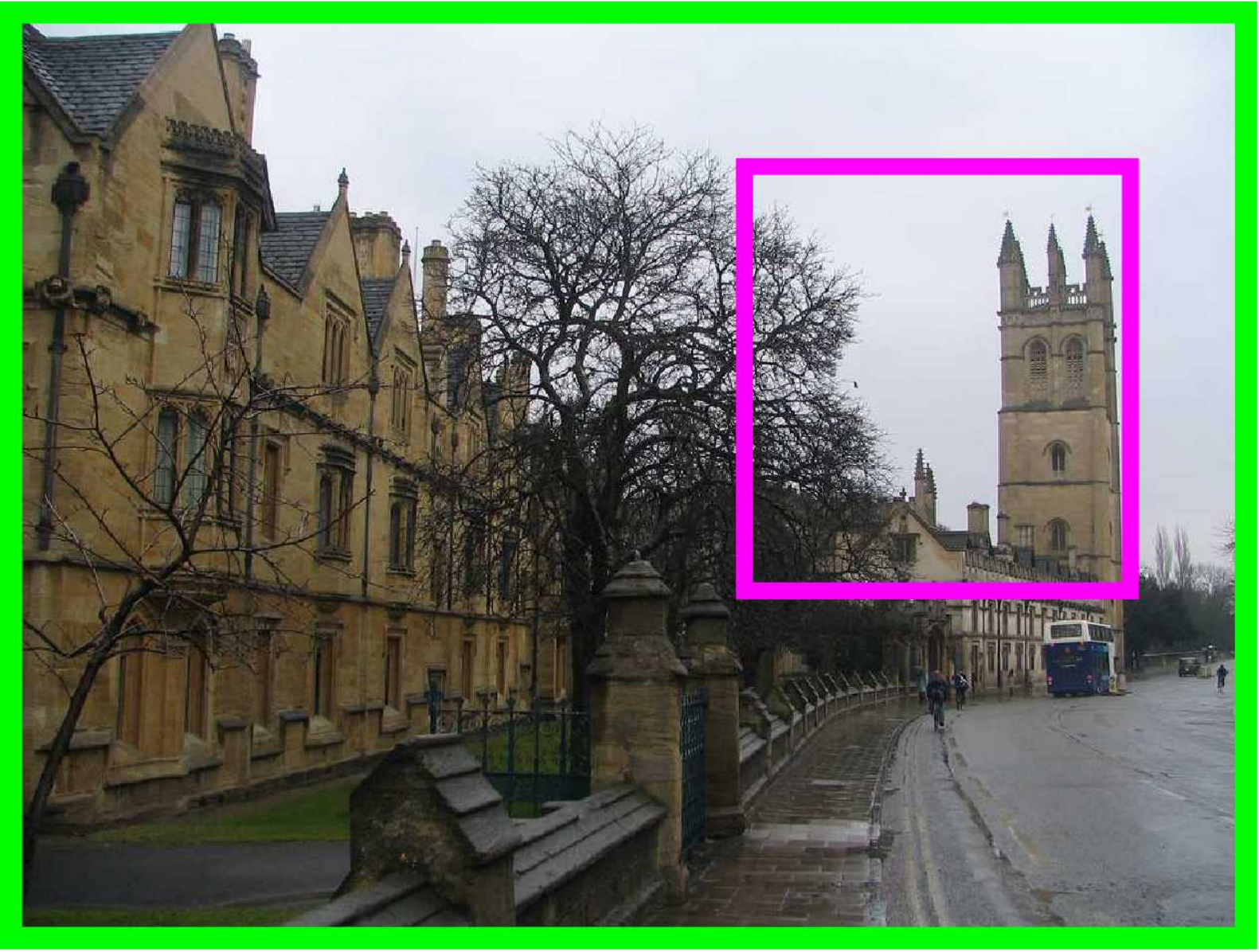}\\860 $\rightarrow$ 10\\ \end{tabular}  

\vspace{2ex}

\begin{tabular}{@{\sssp}c@{\sssp}}\includegraphics[height=\figh]{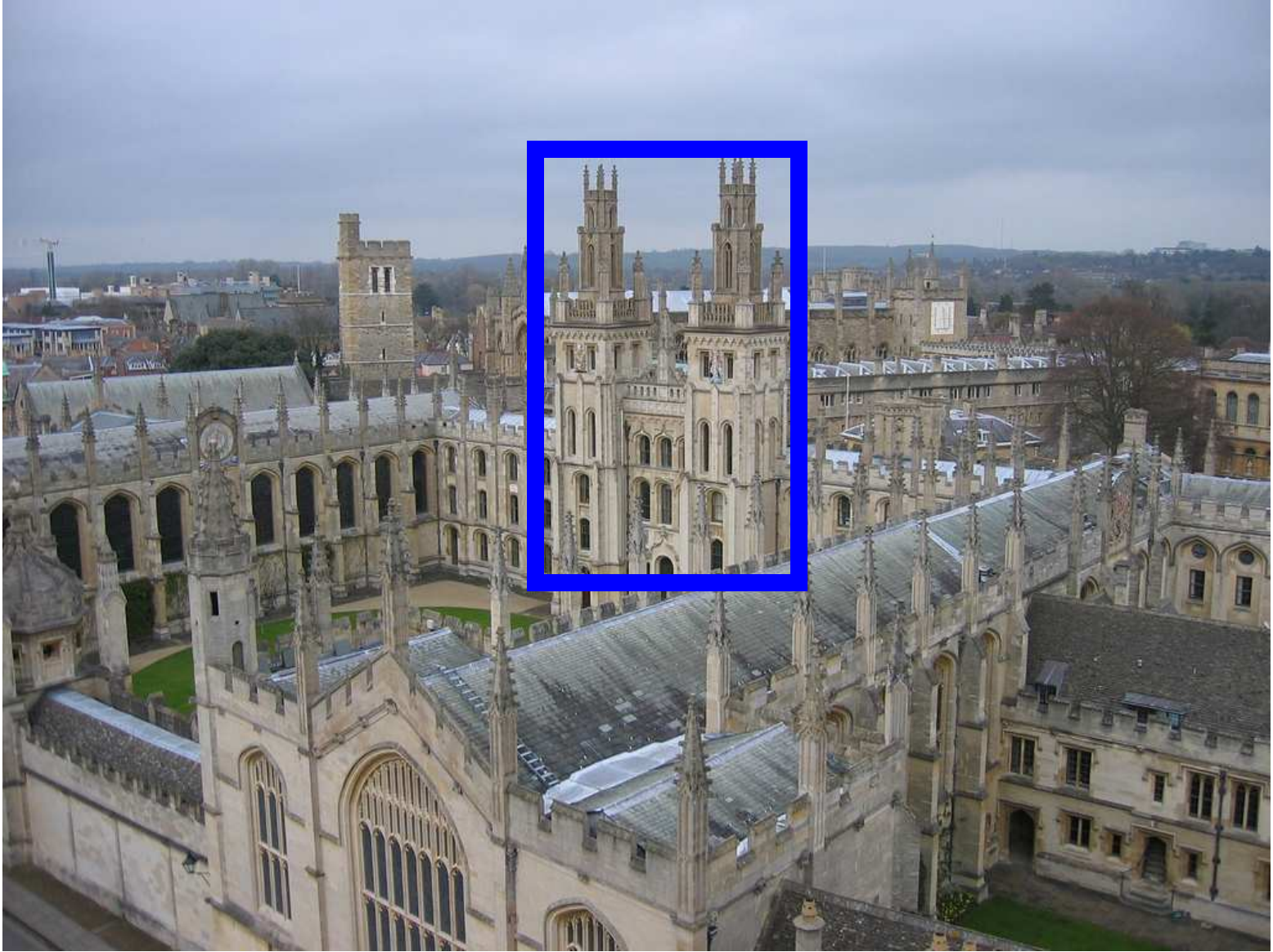}\end{tabular} 
\begin{tabular}{@{\sssp}c@{\sssp}}\includegraphics[height=\figh]{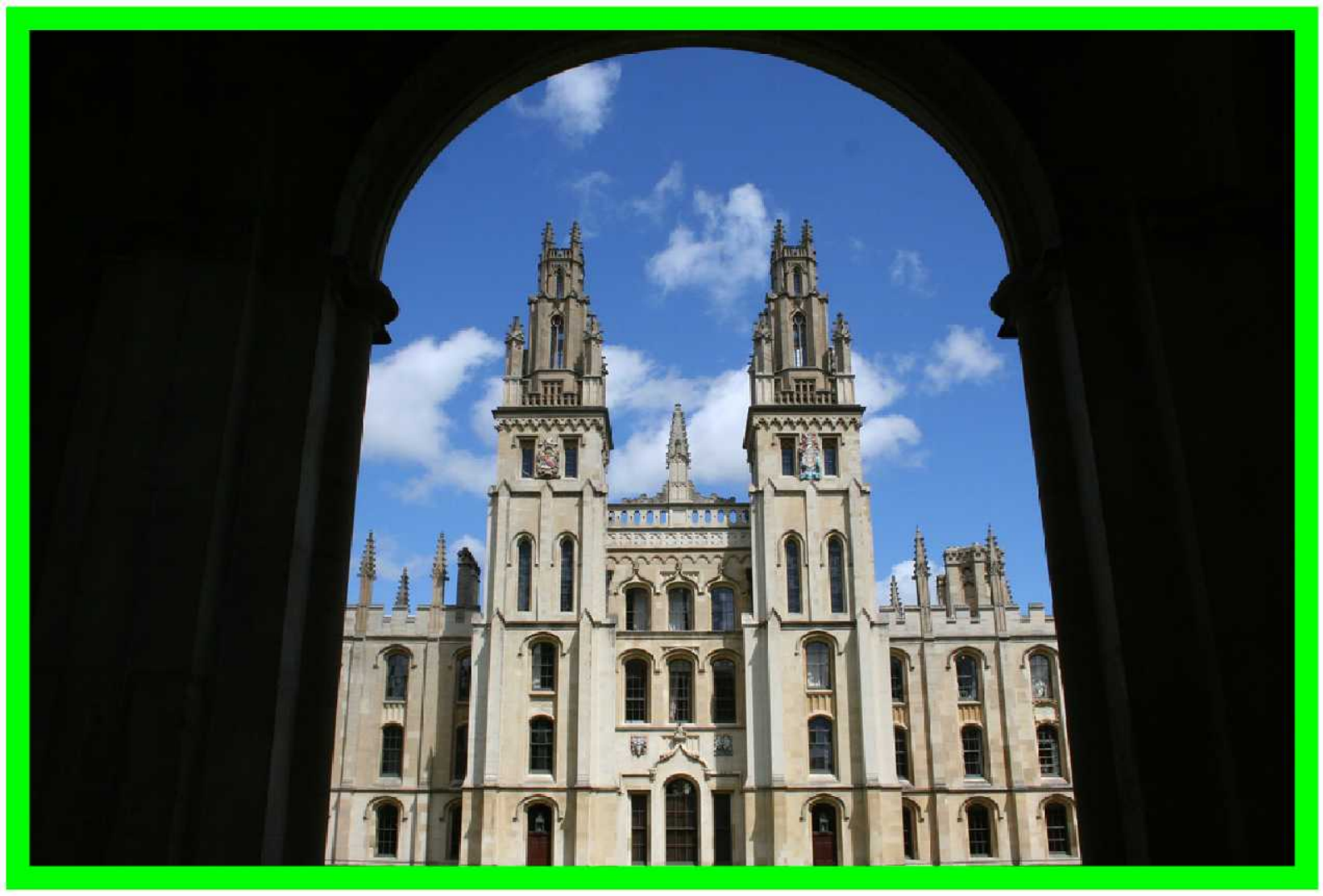}\end{tabular}
\begin{tabular}{@{\sssp}c@{\sssp}}\includegraphics[height=\figh]{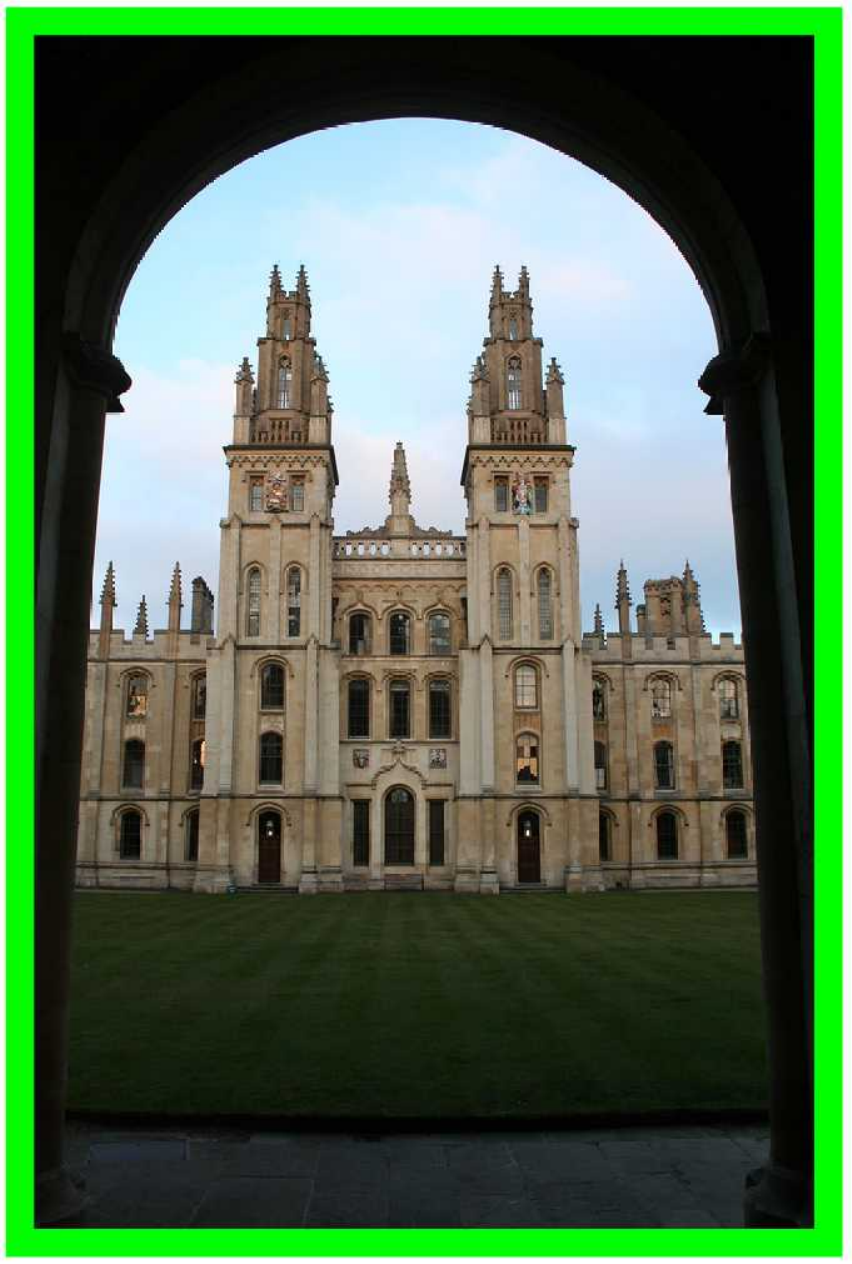}\end{tabular}
\begin{tabular}{@{\sssp}c@{\sssp}}\includegraphics[height=\figh]{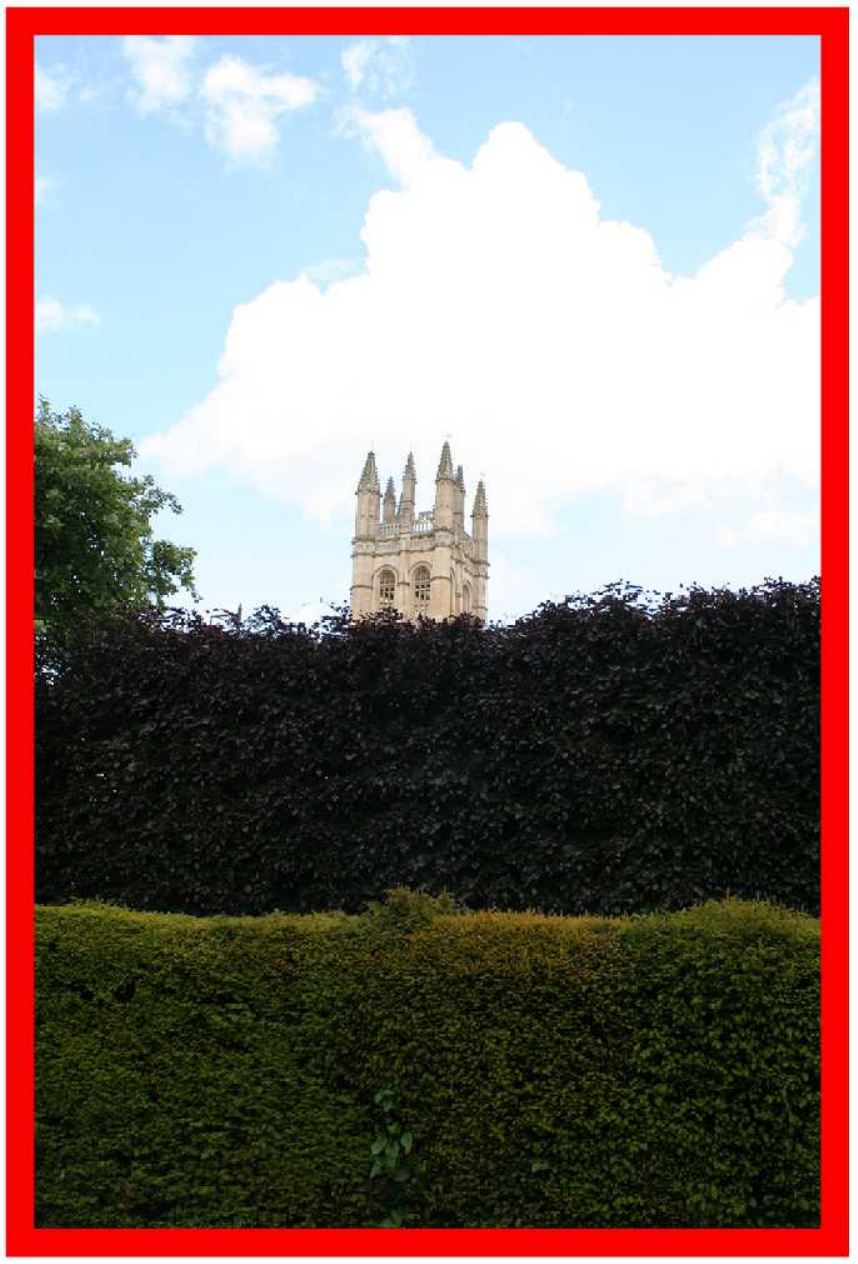}\end{tabular}
\begin{tabular}{@{\sssp}c@{\sssp}}\includegraphics[height=\figh]{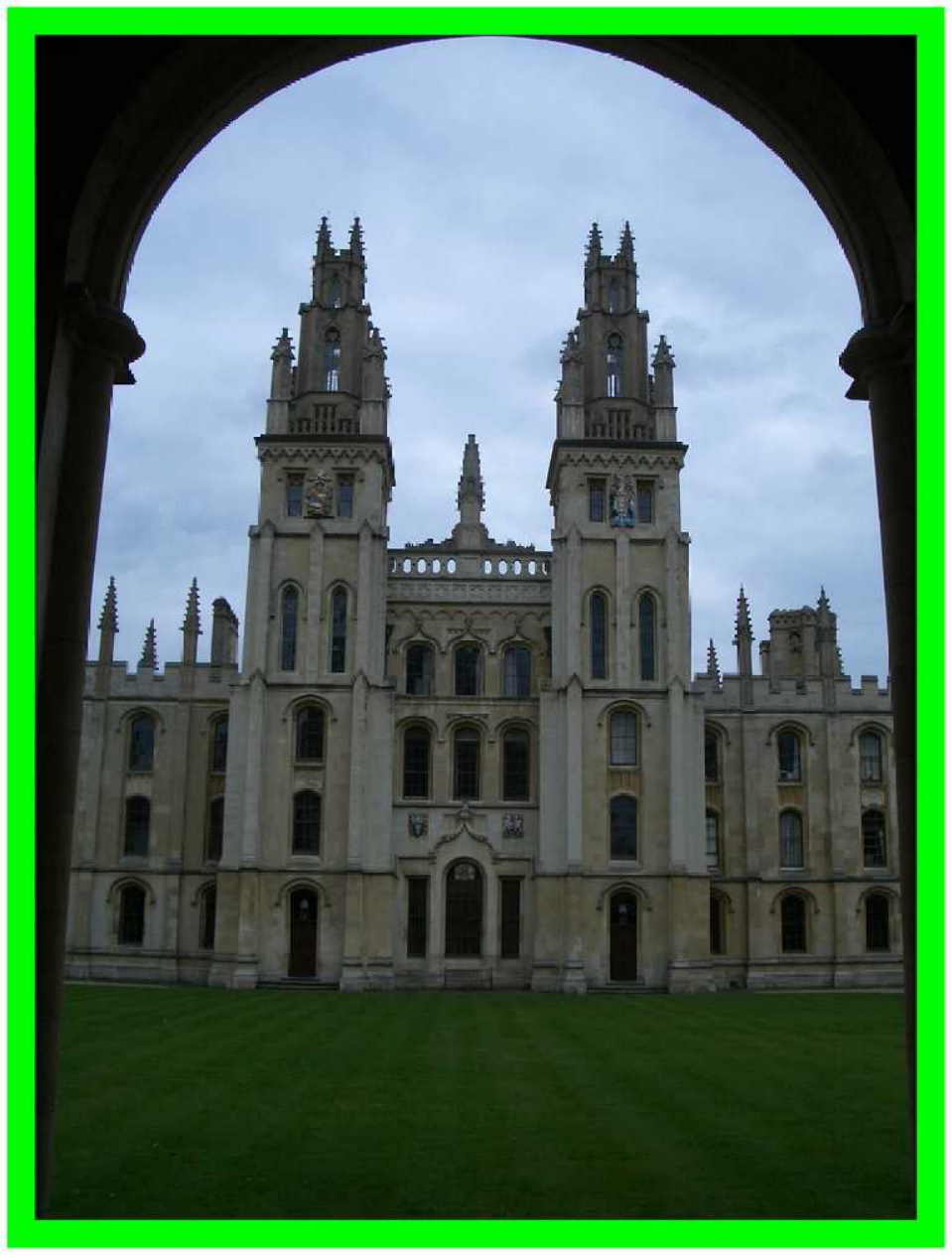}\end{tabular}
\begin{tabular}{@{\sssp}c@{\sssp}}\includegraphics[height=\figh]{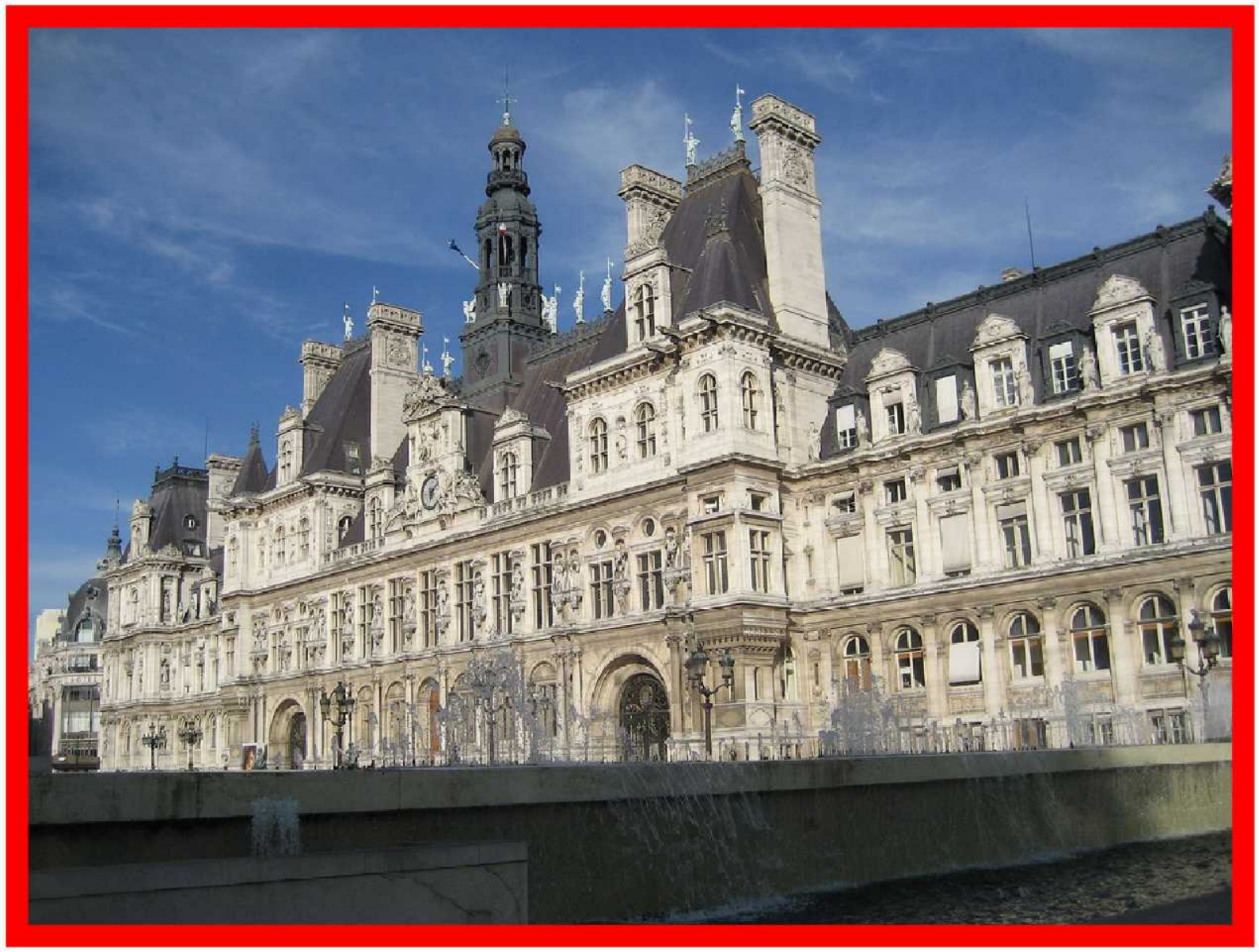}\end{tabular}
\begin{tabular}{@{\sssp}c@{\sssp}}\includegraphics[height=\figh]{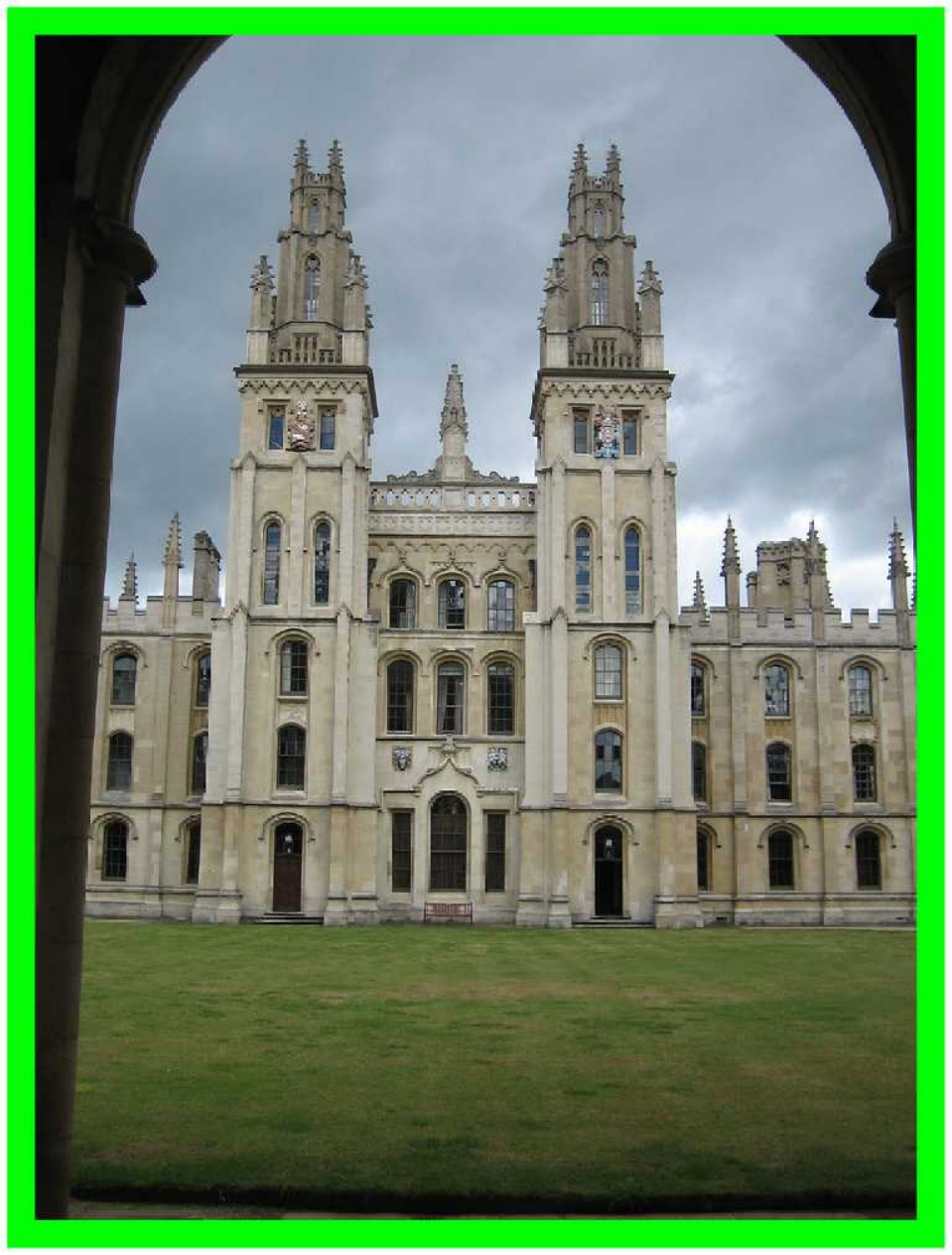}\end{tabular}
\begin{tabular}{@{\sssp}c@{\sssp}}\includegraphics[height=\figh]{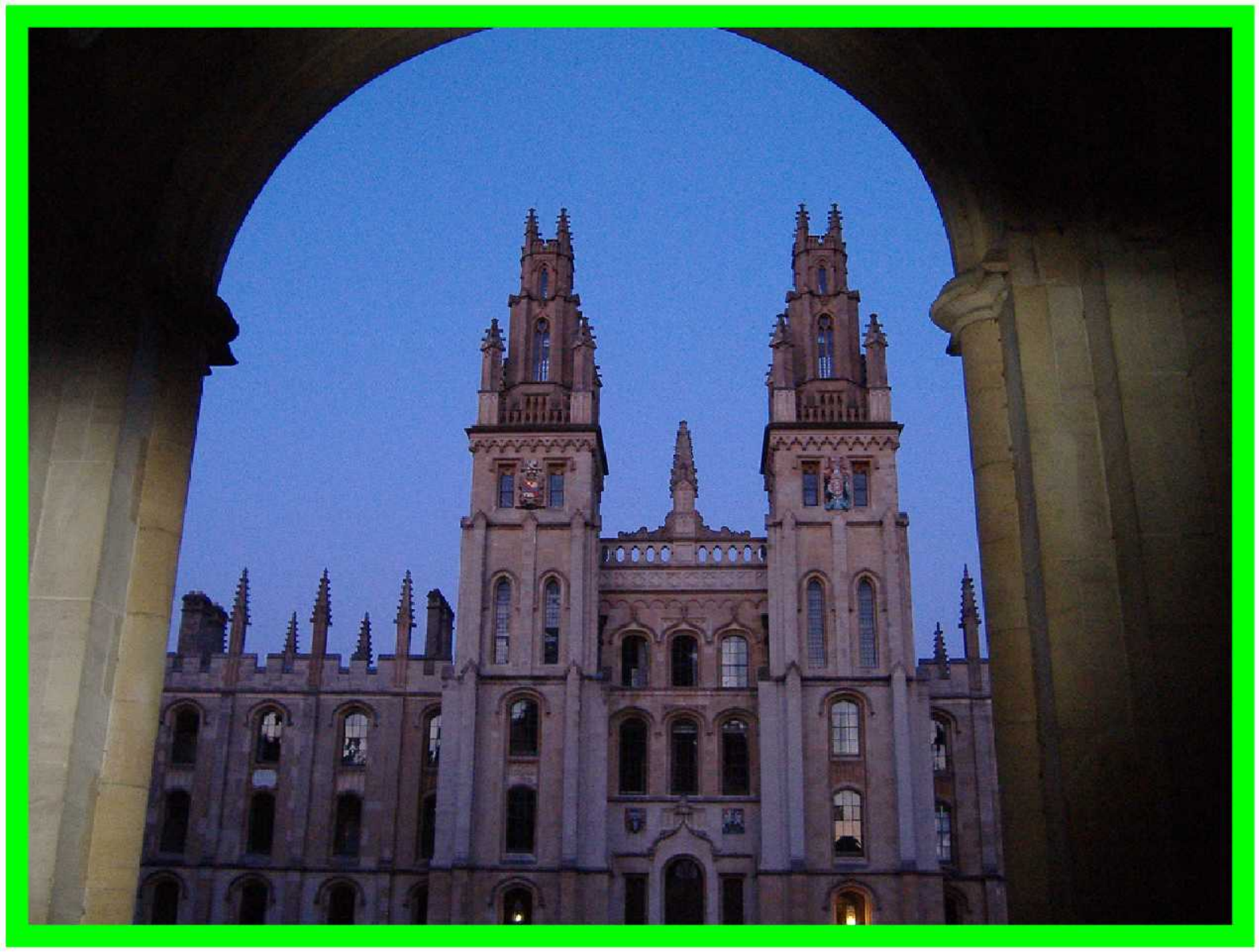}\end{tabular}
\begin{tabular}{@{\sssp}c@{\sssp}}\includegraphics[height=\figh]{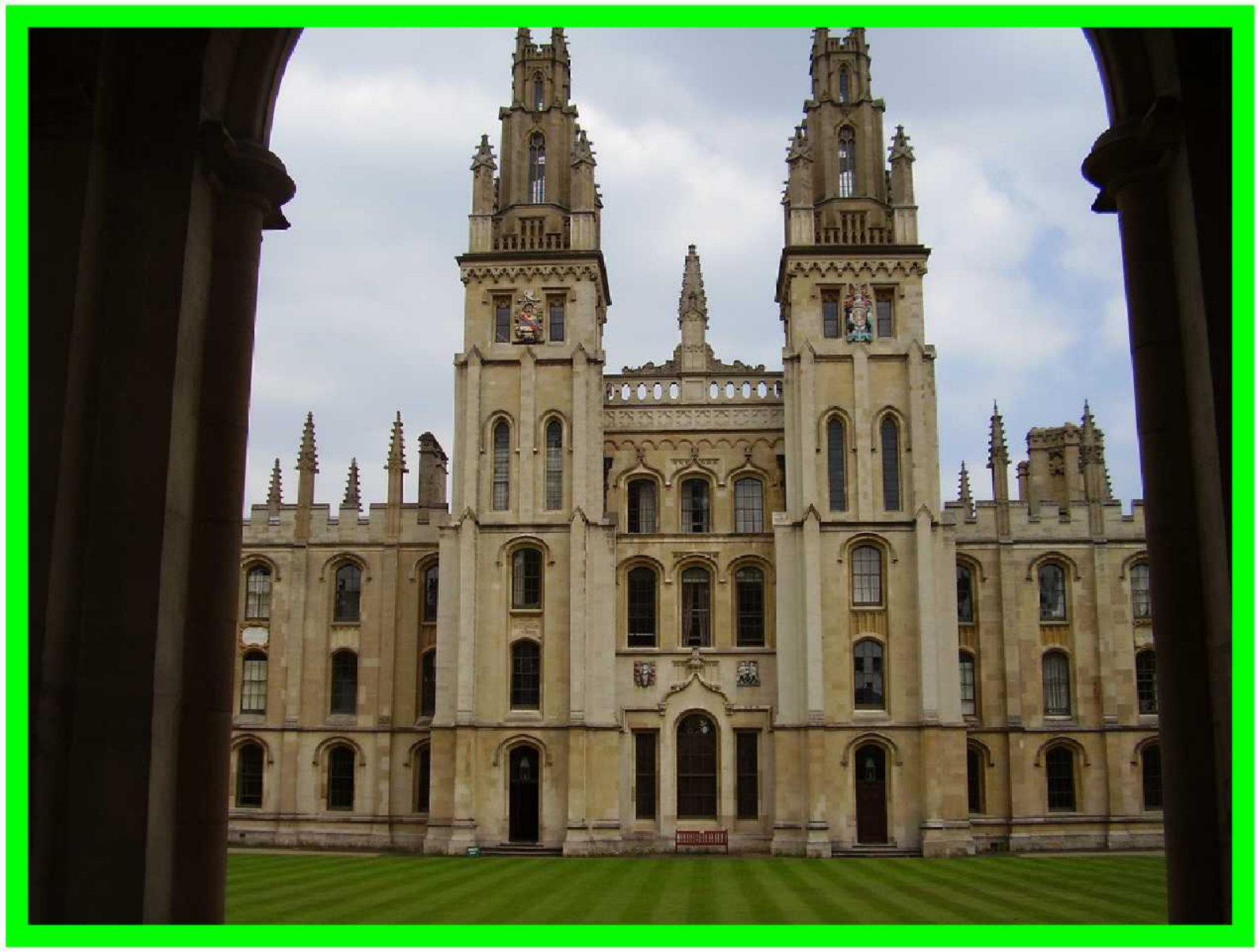}\end{tabular}
\begin{tabular}{@{\sssp}c@{\sssp}}\includegraphics[height=\figh]{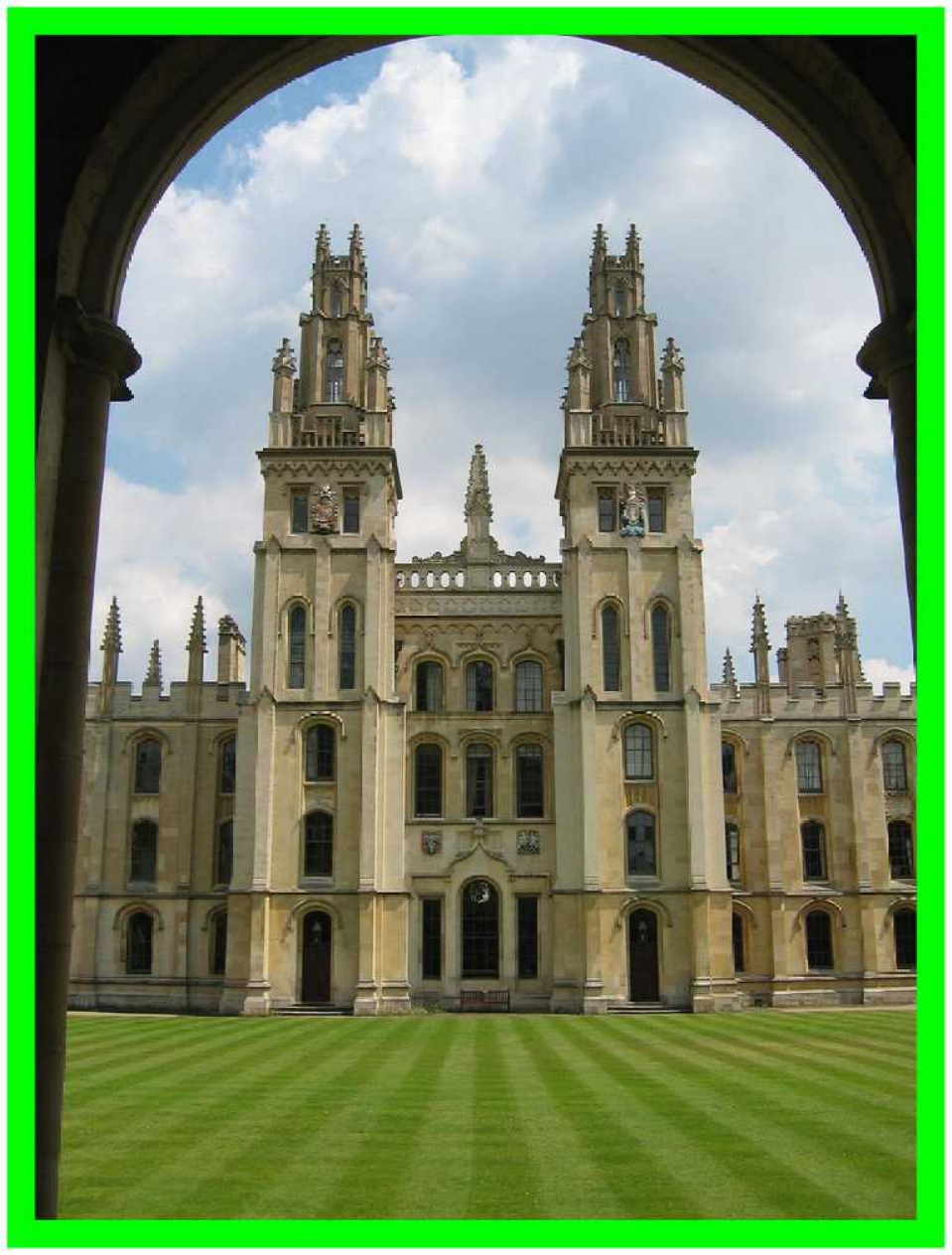}\end{tabular}
\begin{tabular}{@{\sssp}c@{\sssp}}\includegraphics[height=\figh]{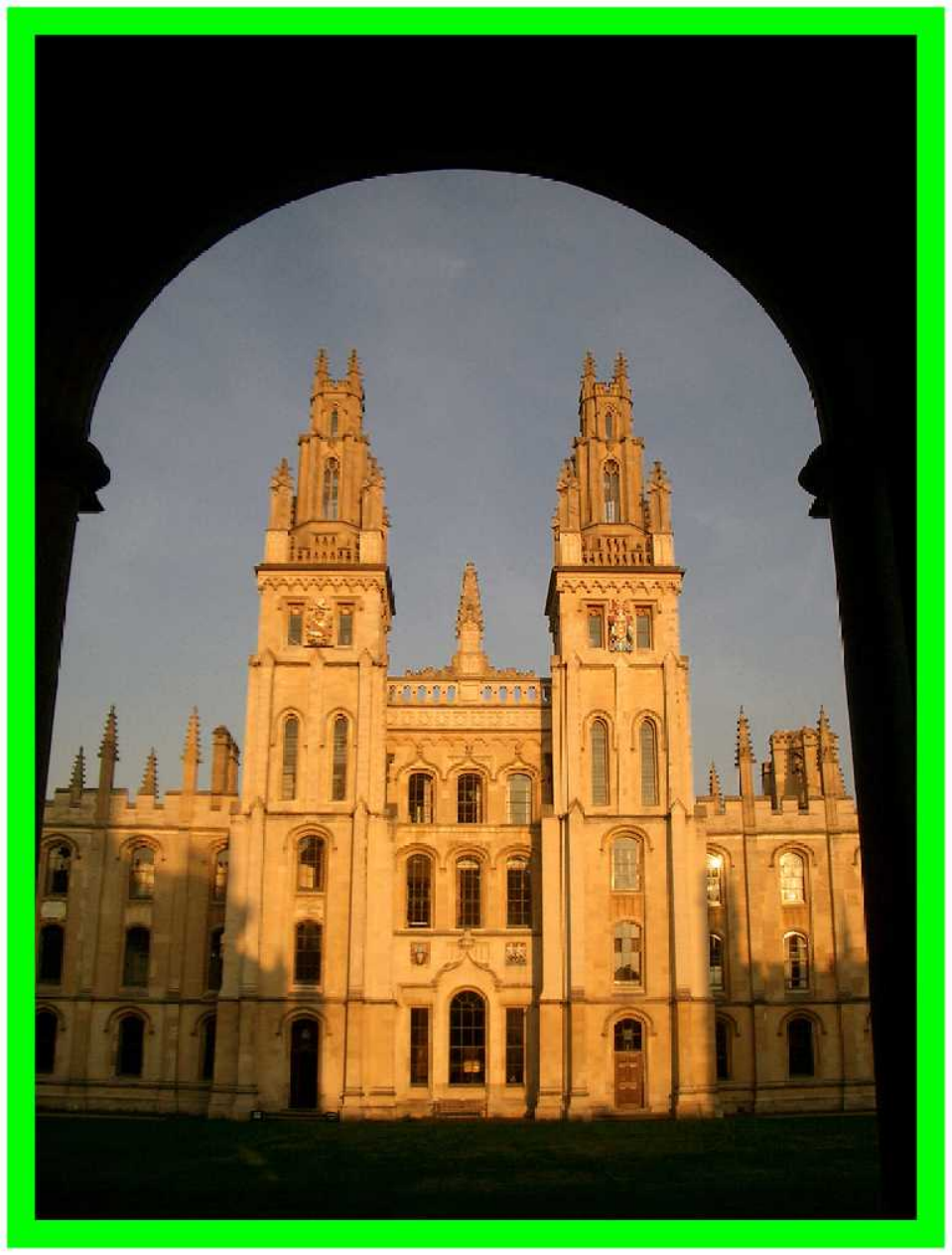}\end{tabular}

\begin{tabular}{@{\sssp}c@{\sssp}}\includegraphics[height=\figh]{figs/rerank/4//q4_3004_7522.pdf}\\Query\\ \end{tabular} 
\begin{tabular}{@{\sssp}c@{\sssp}}\includegraphics[height=\figh]{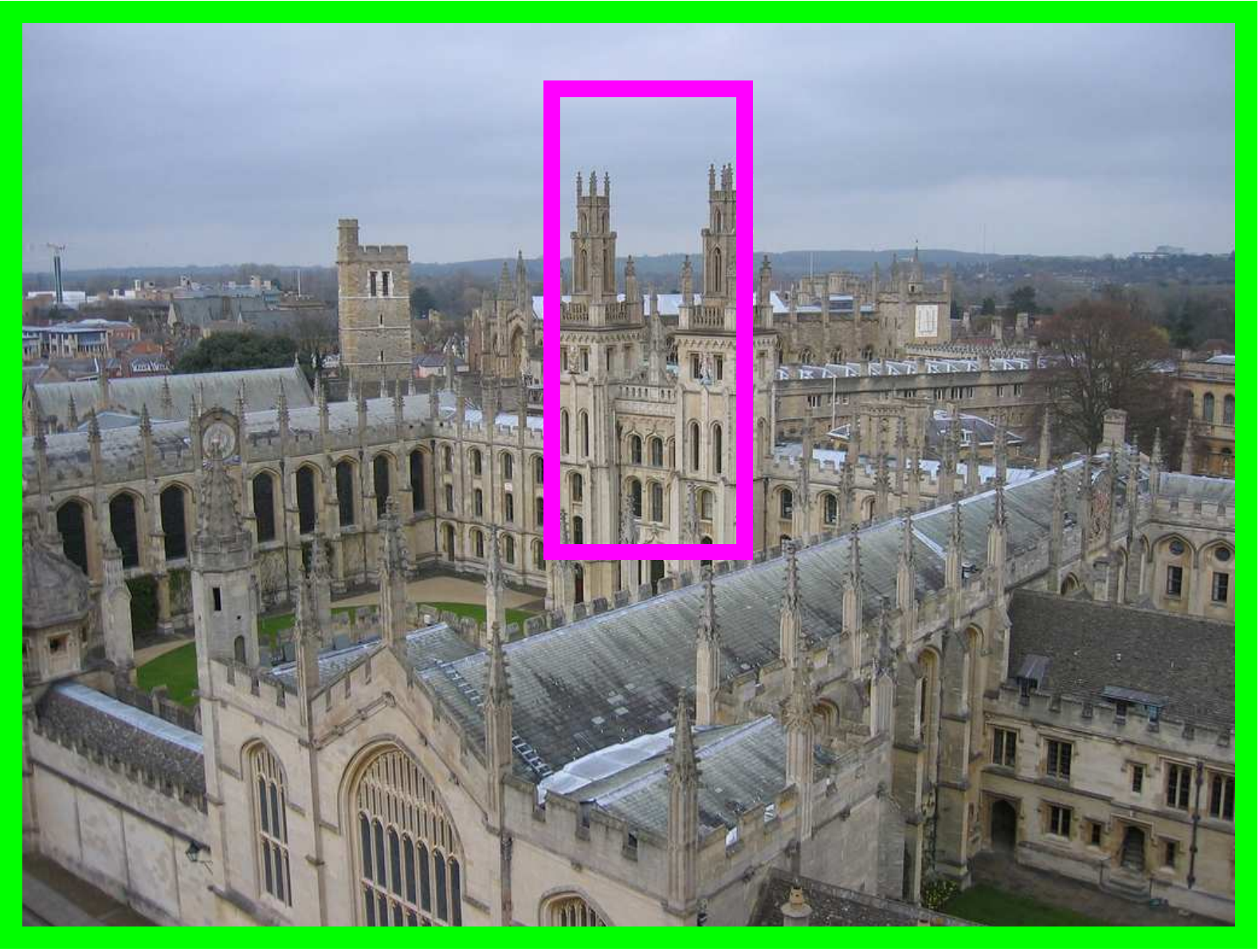}\\120 $\rightarrow$ 1\\ \end{tabular} 
\begin{tabular}{@{\sssp}c@{\sssp}}\includegraphics[height=\figh]{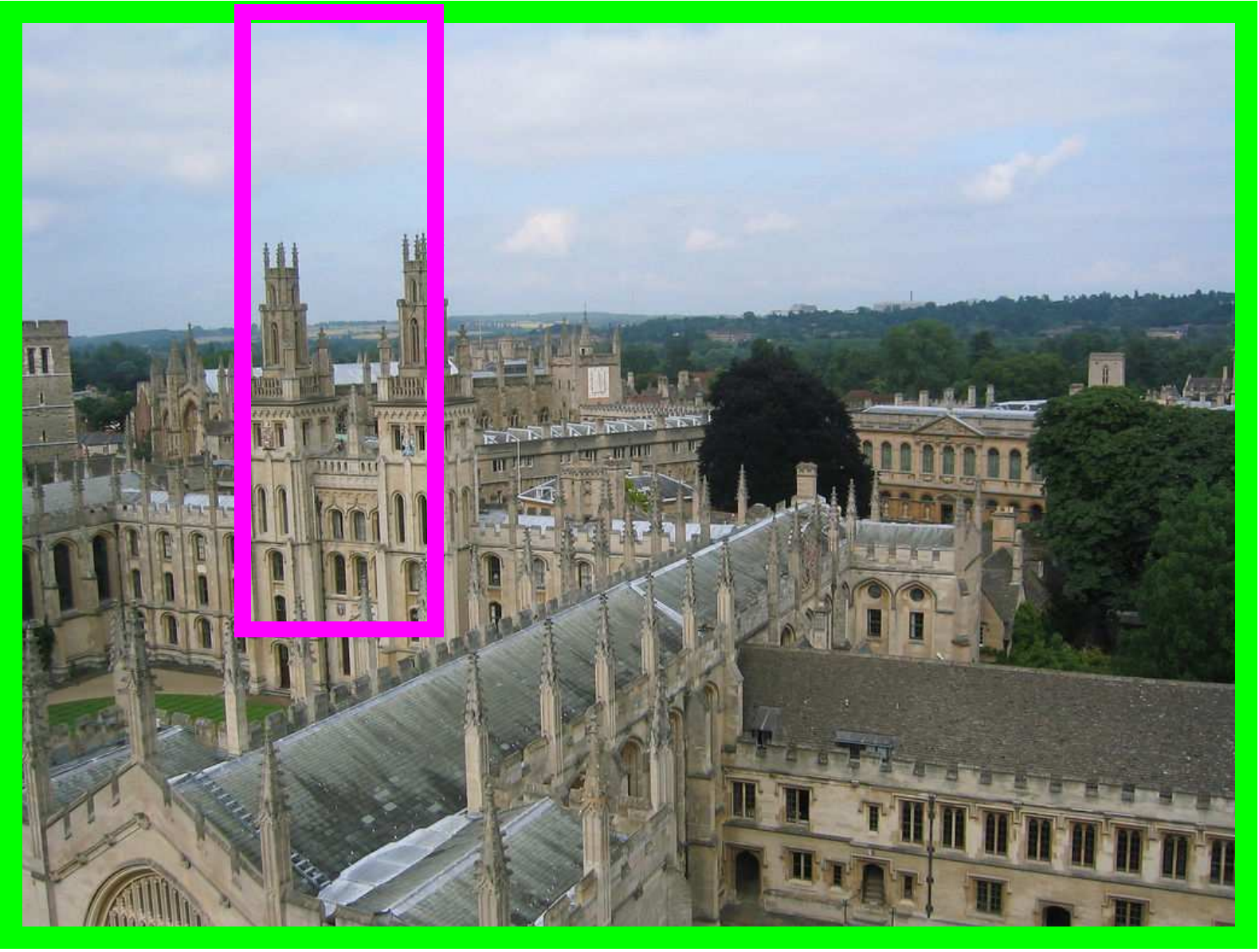}\\118 $\rightarrow$ 2\\ \end{tabular} 
\begin{tabular}{@{\sssp}c@{\sssp}}\includegraphics[height=\figh]{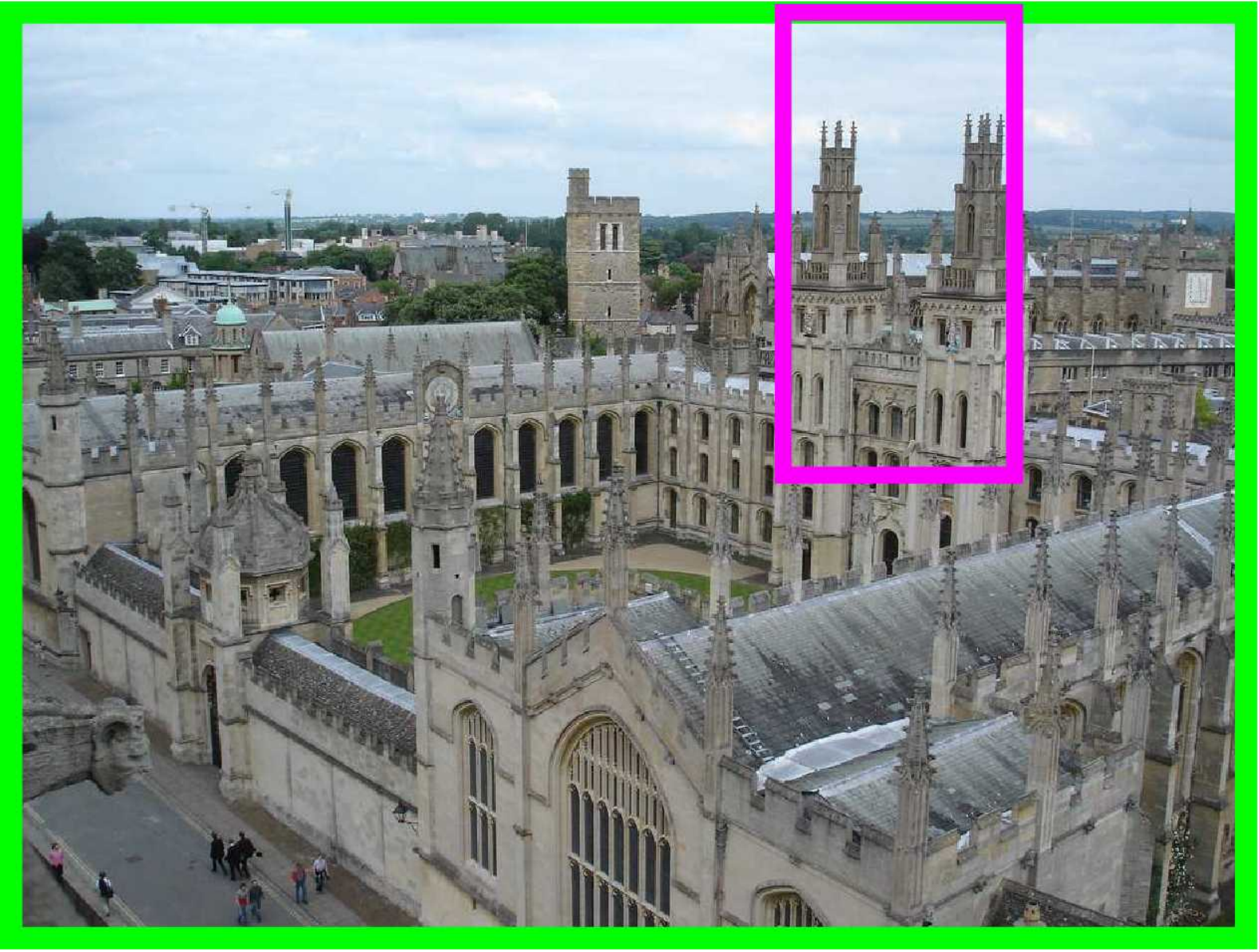}\\753 $\rightarrow$ 3\\ \end{tabular} 
\begin{tabular}{@{\sssp}c@{\sssp}}\includegraphics[height=\figh]{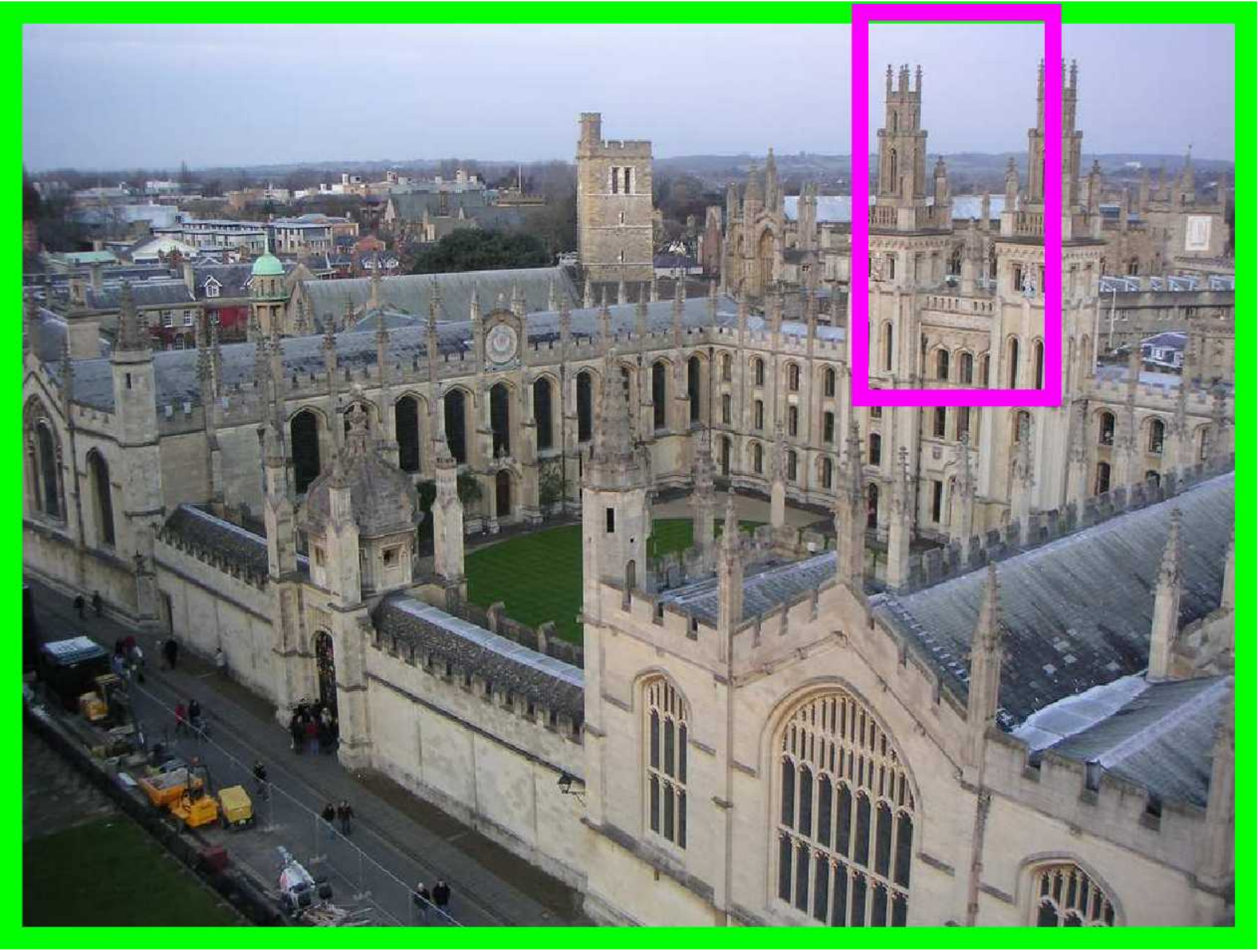}\\467 $\rightarrow$ 4\\ \end{tabular} 
\begin{tabular}{@{\sssp}c@{\sssp}}\includegraphics[height=\figh]{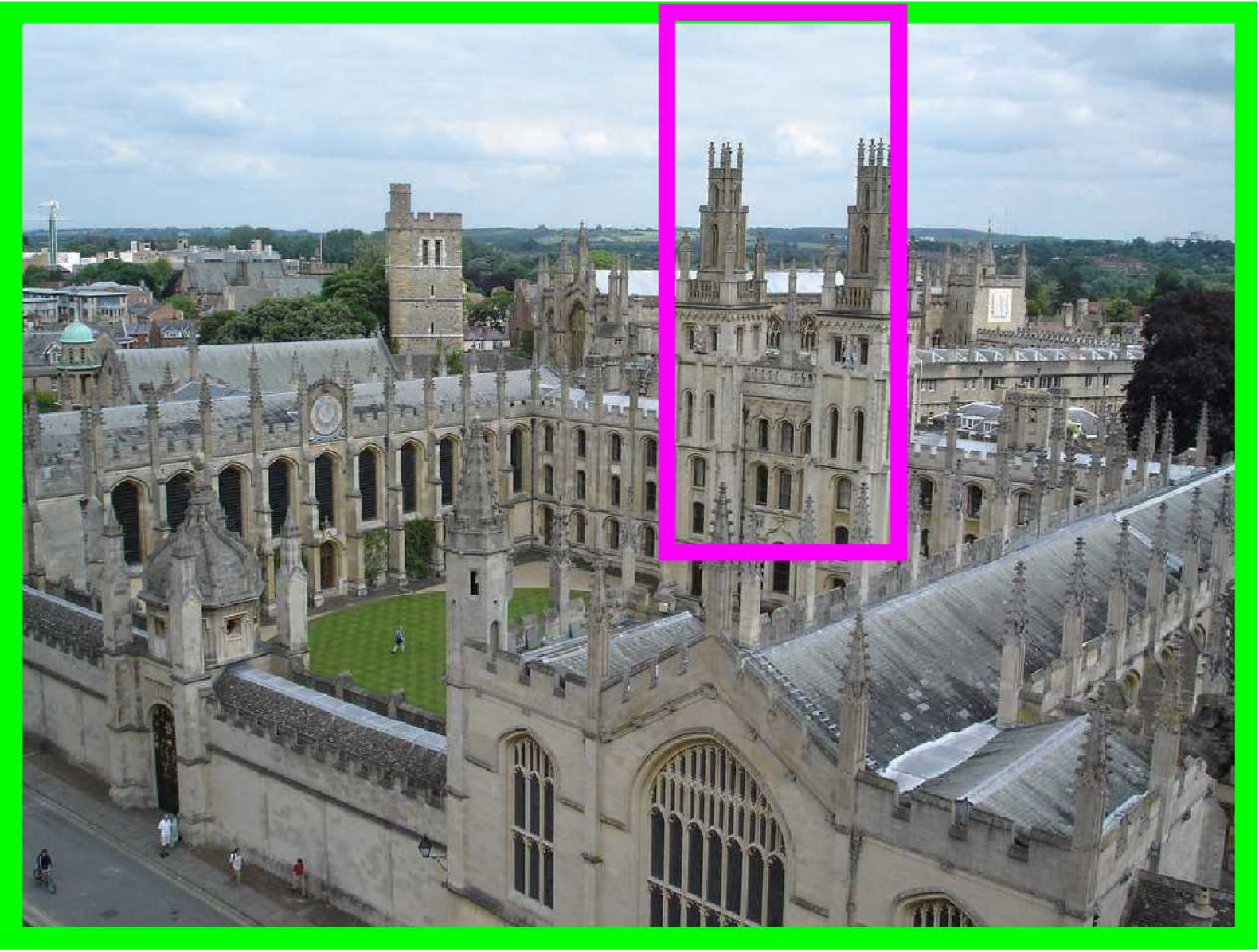}\\631 $\rightarrow$ 5\\ \end{tabular} 
\begin{tabular}{@{\sssp}c@{\sssp}}\includegraphics[height=\figh]{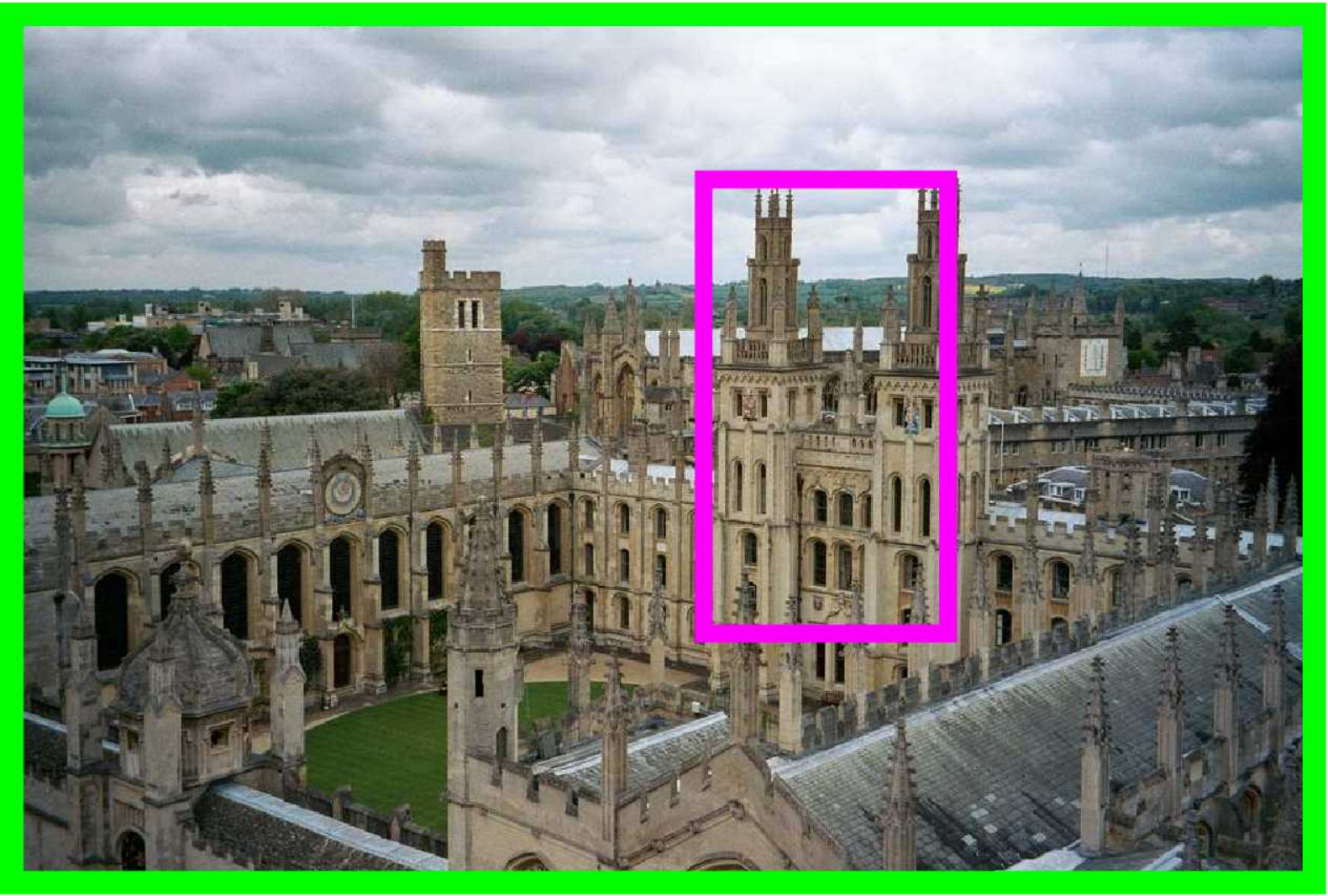}\\82 $\rightarrow$ 6\\ \end{tabular} 
\begin{tabular}{@{\sssp}c@{\sssp}}\includegraphics[height=\figh]{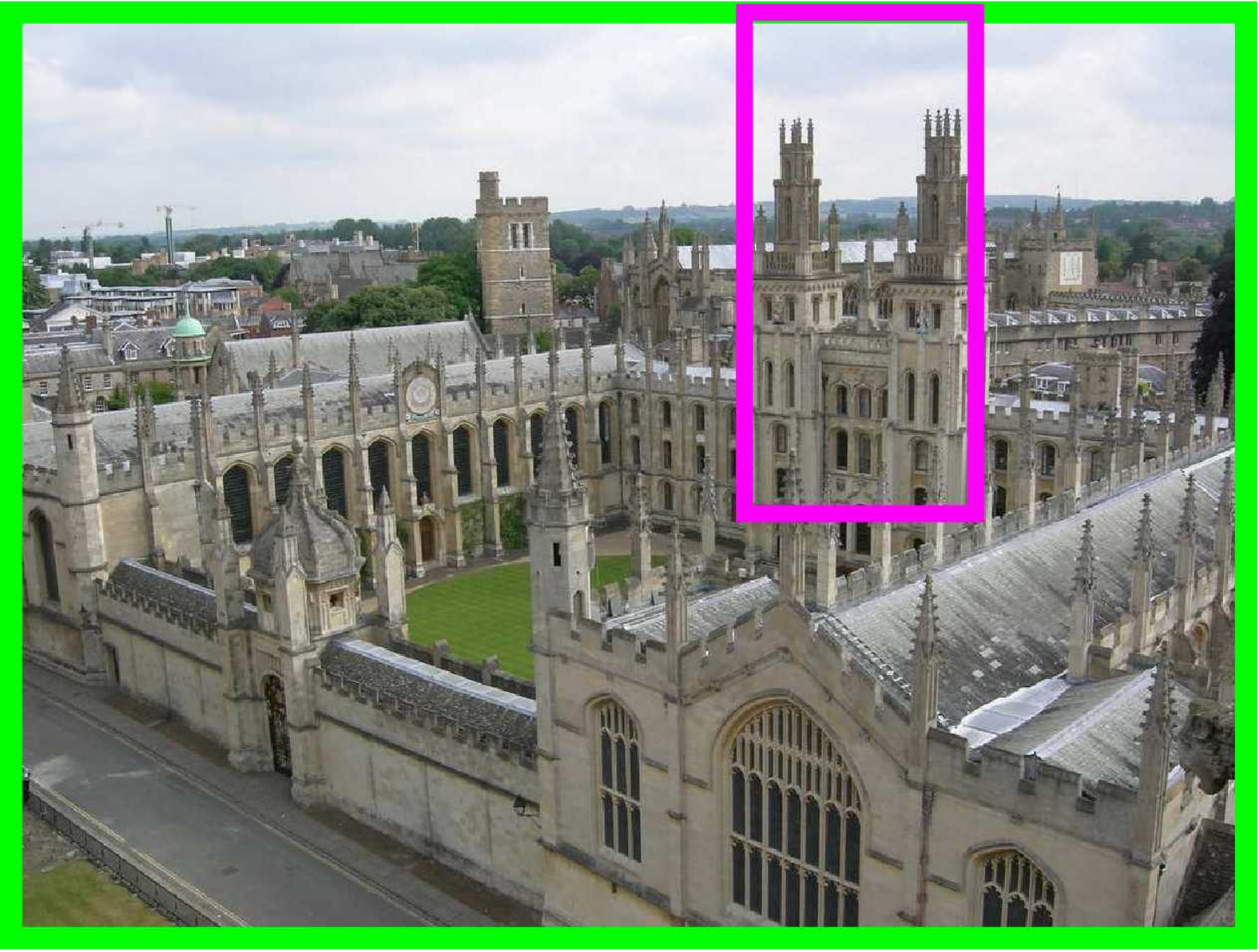}\\594 $\rightarrow$ 7\\ \end{tabular} 

  \vspace{2ex}

\begin{tabular}{@{\sssp}c@{\sssp}}\includegraphics[height=\figh]{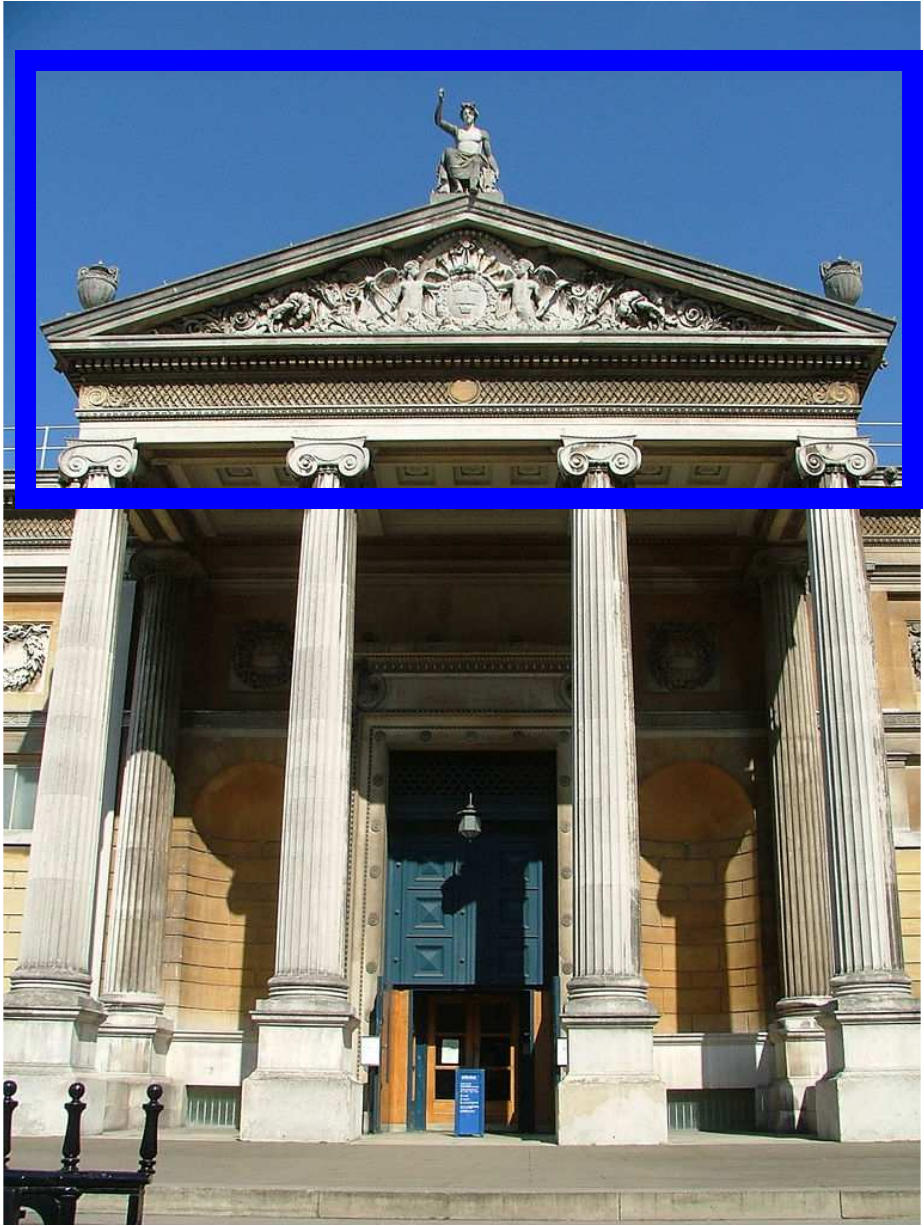}\end{tabular} 
\begin{tabular}{@{\sssp}c@{\sssp}}\includegraphics[height=\figh]{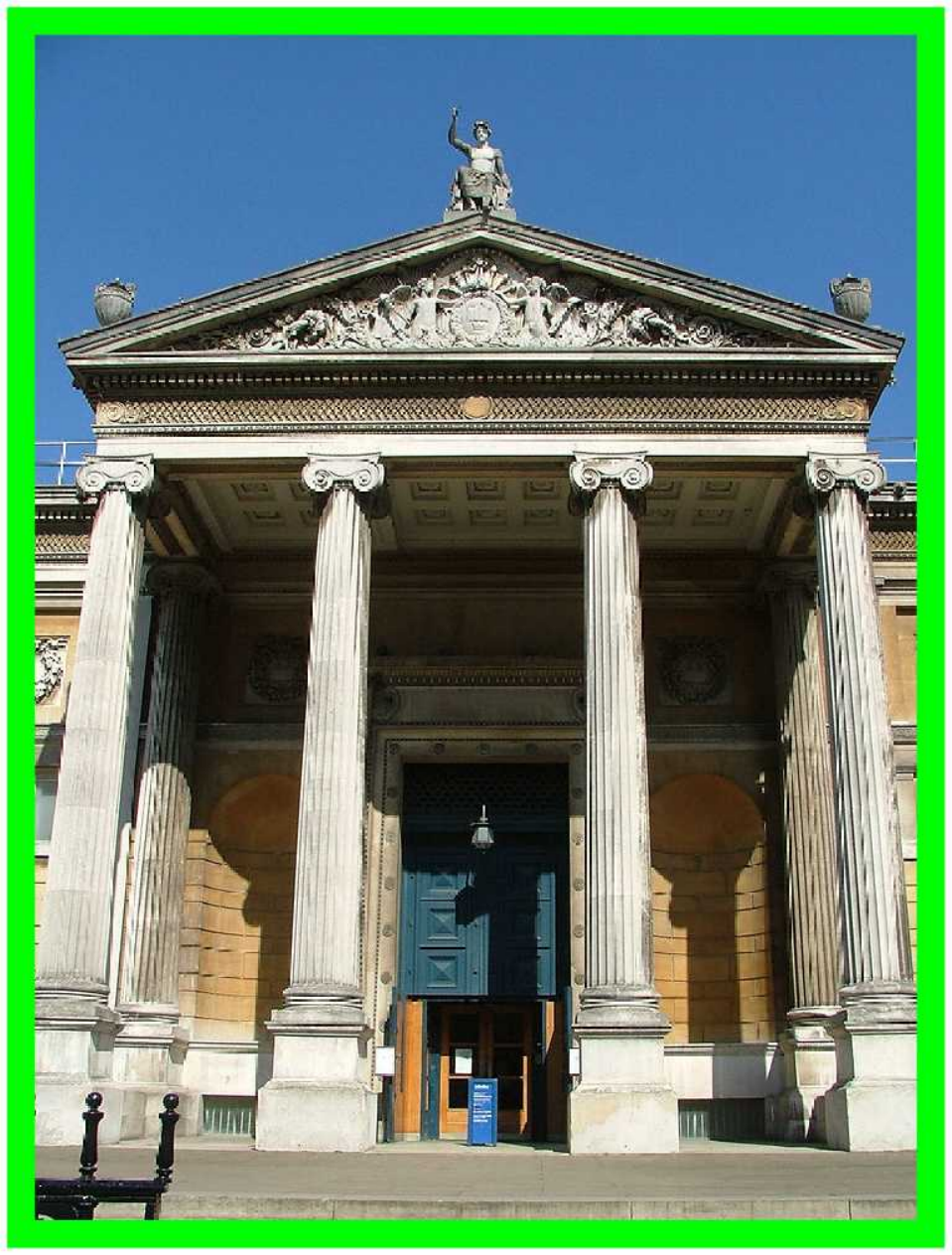}\end{tabular}
\begin{tabular}{@{\sssp}c@{\sssp}}\includegraphics[height=\figh]{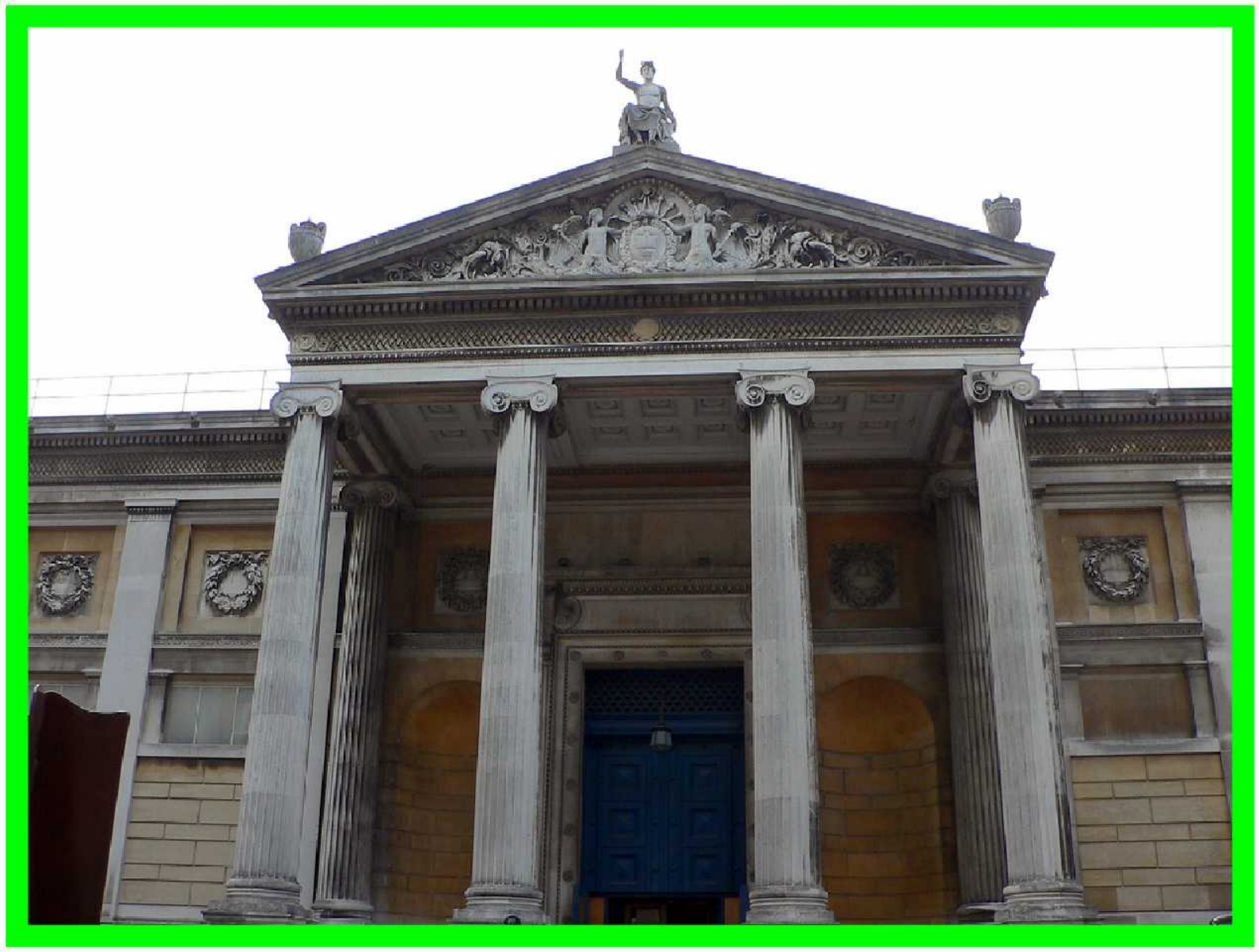}\end{tabular}
\begin{tabular}{@{\sssp}c@{\sssp}}\includegraphics[height=\figh]{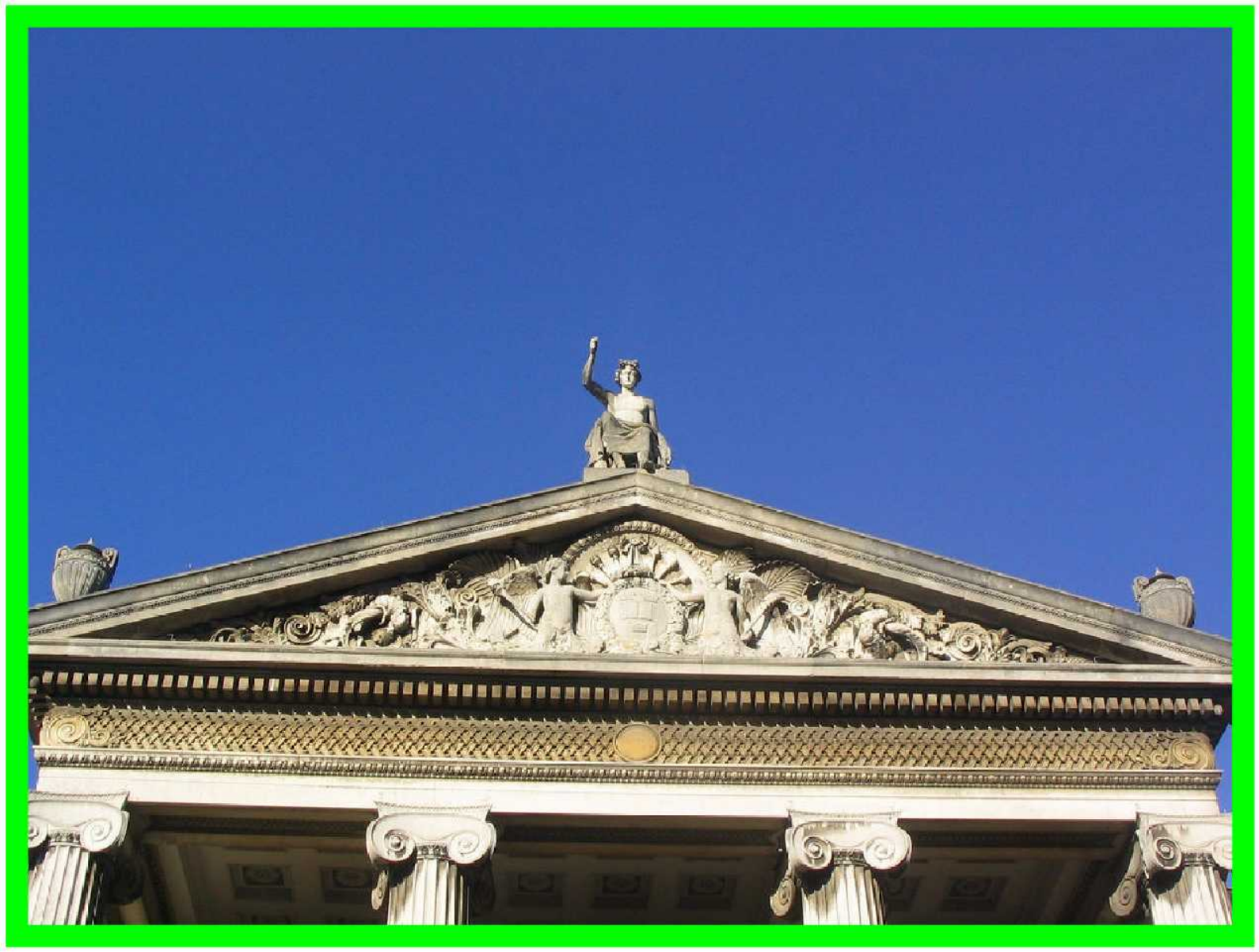}\end{tabular}
\begin{tabular}{@{\sssp}c@{\sssp}}\includegraphics[height=\figh]{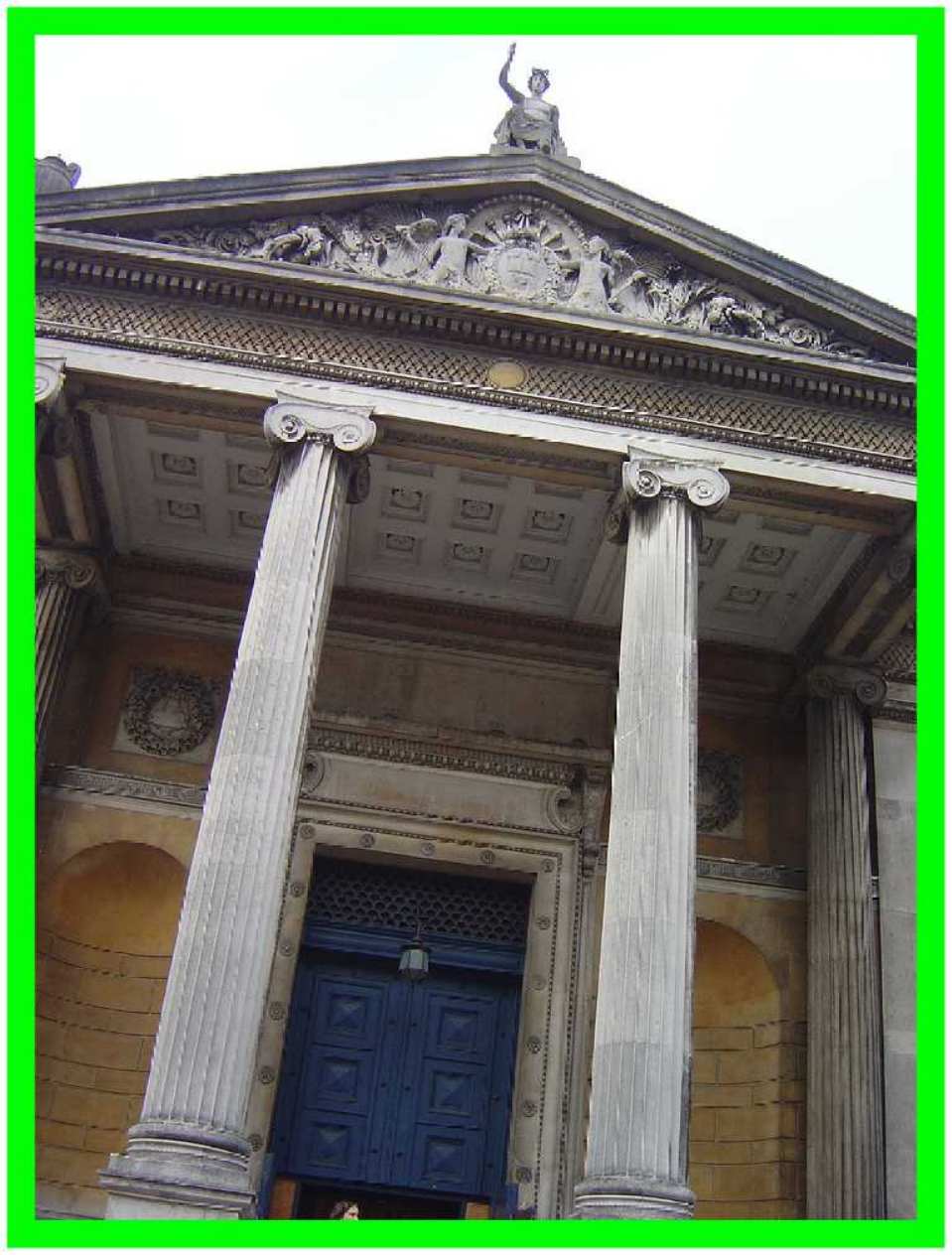}\end{tabular}
\begin{tabular}{@{\sssp}c@{\sssp}}\includegraphics[height=\figh]{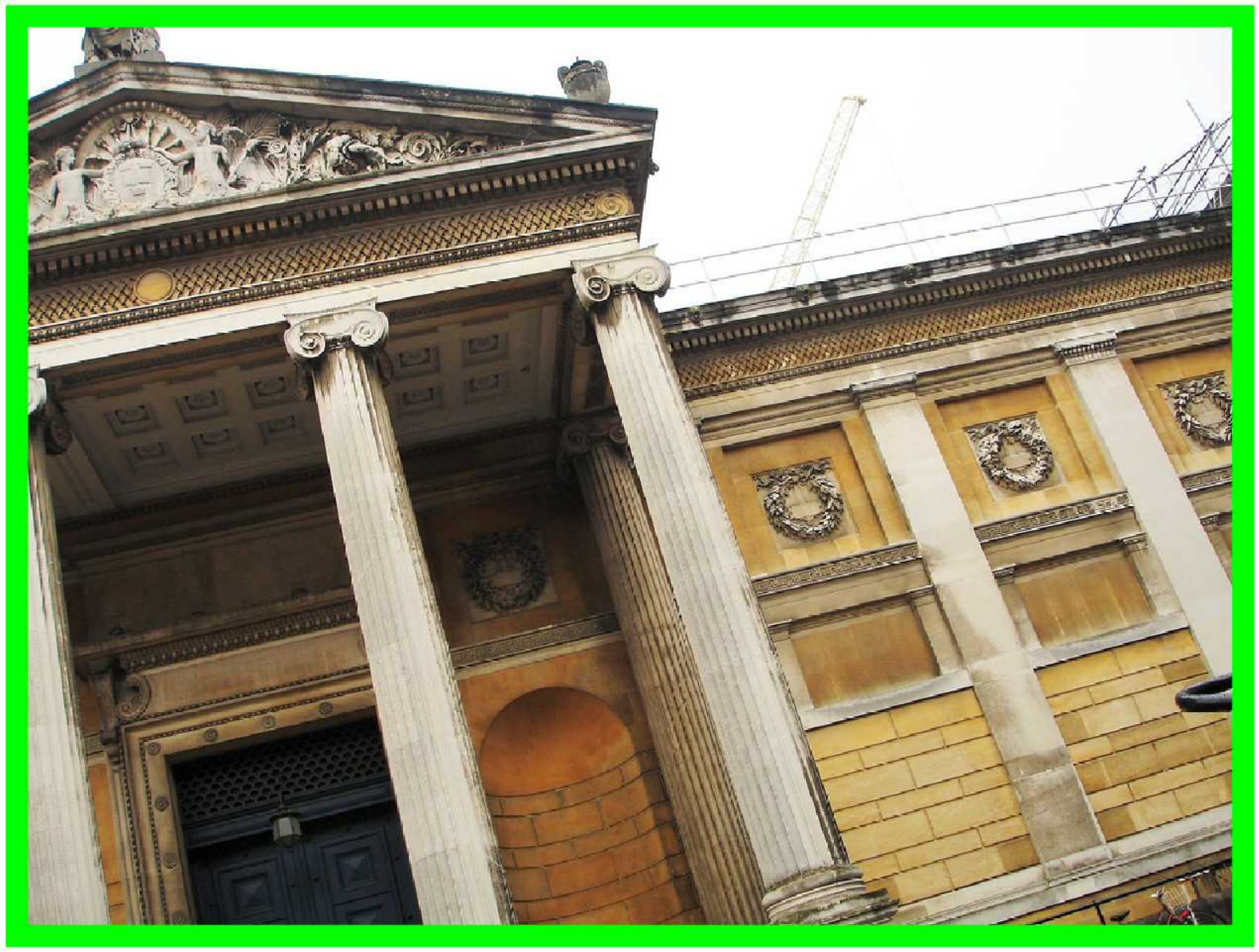}\end{tabular}
\begin{tabular}{@{\sssp}c@{\sssp}}\includegraphics[height=\figh]{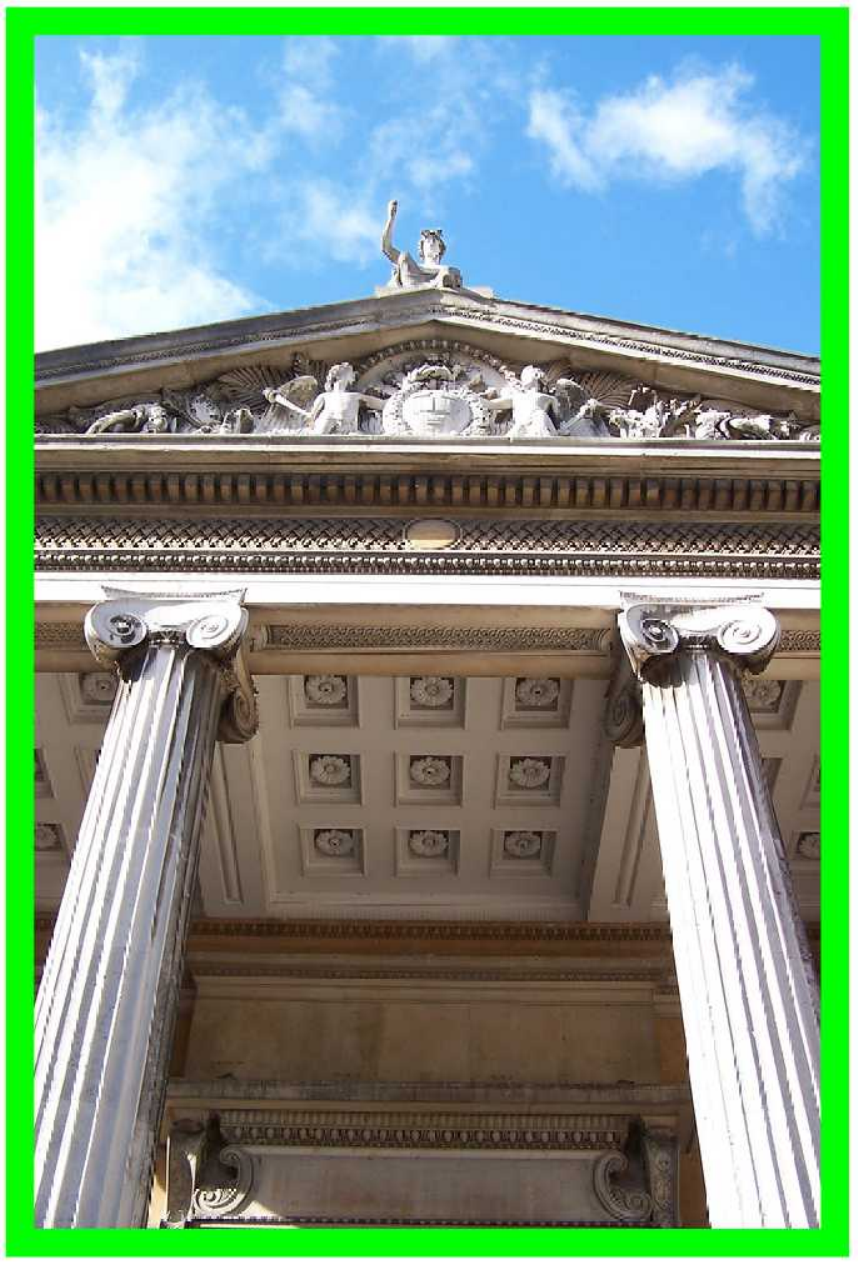}\end{tabular}
\begin{tabular}{@{\sssp}c@{\sssp}}\includegraphics[height=\figh]{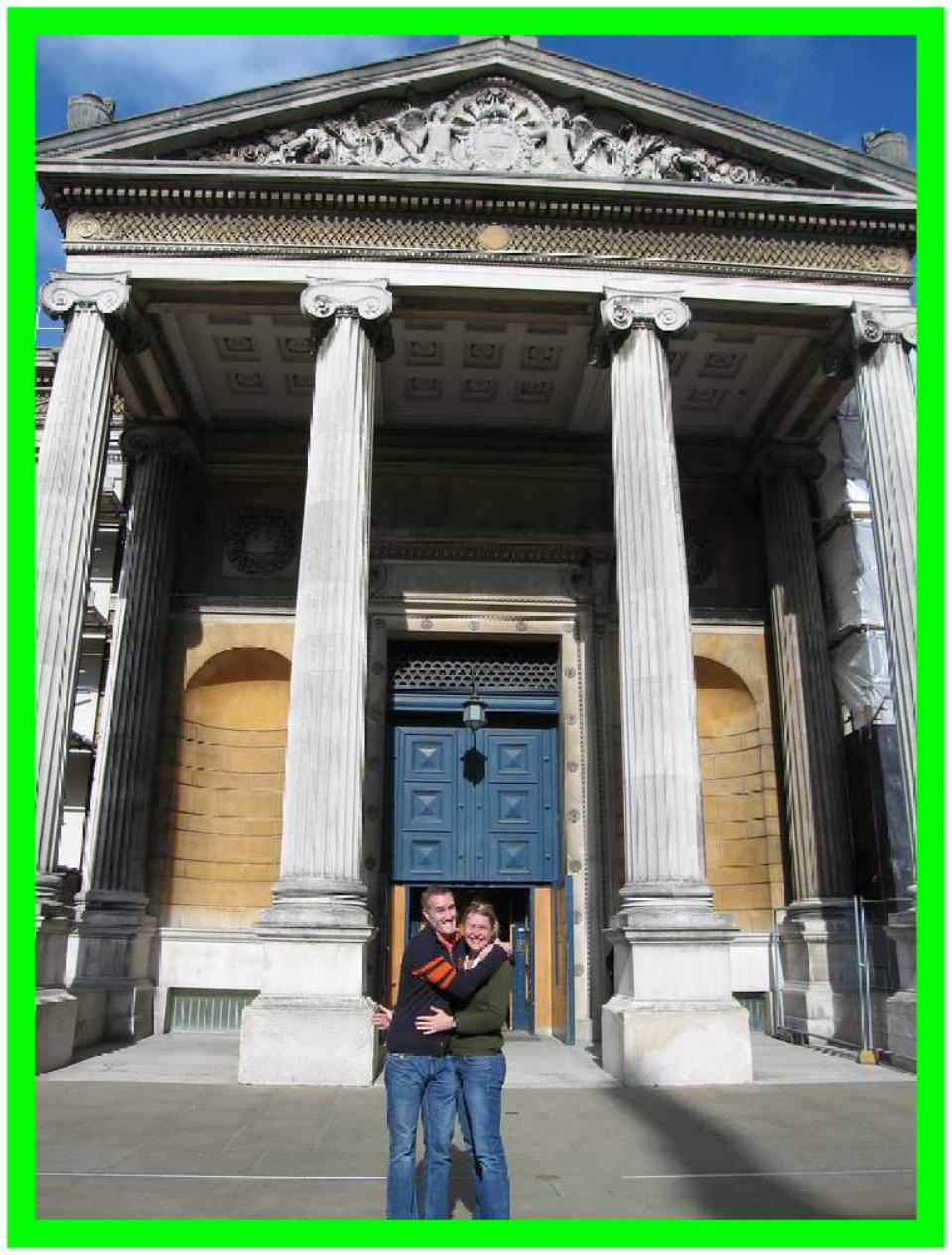}\end{tabular}
\begin{tabular}{@{\sssp}c@{\sssp}}\includegraphics[height=\figh]{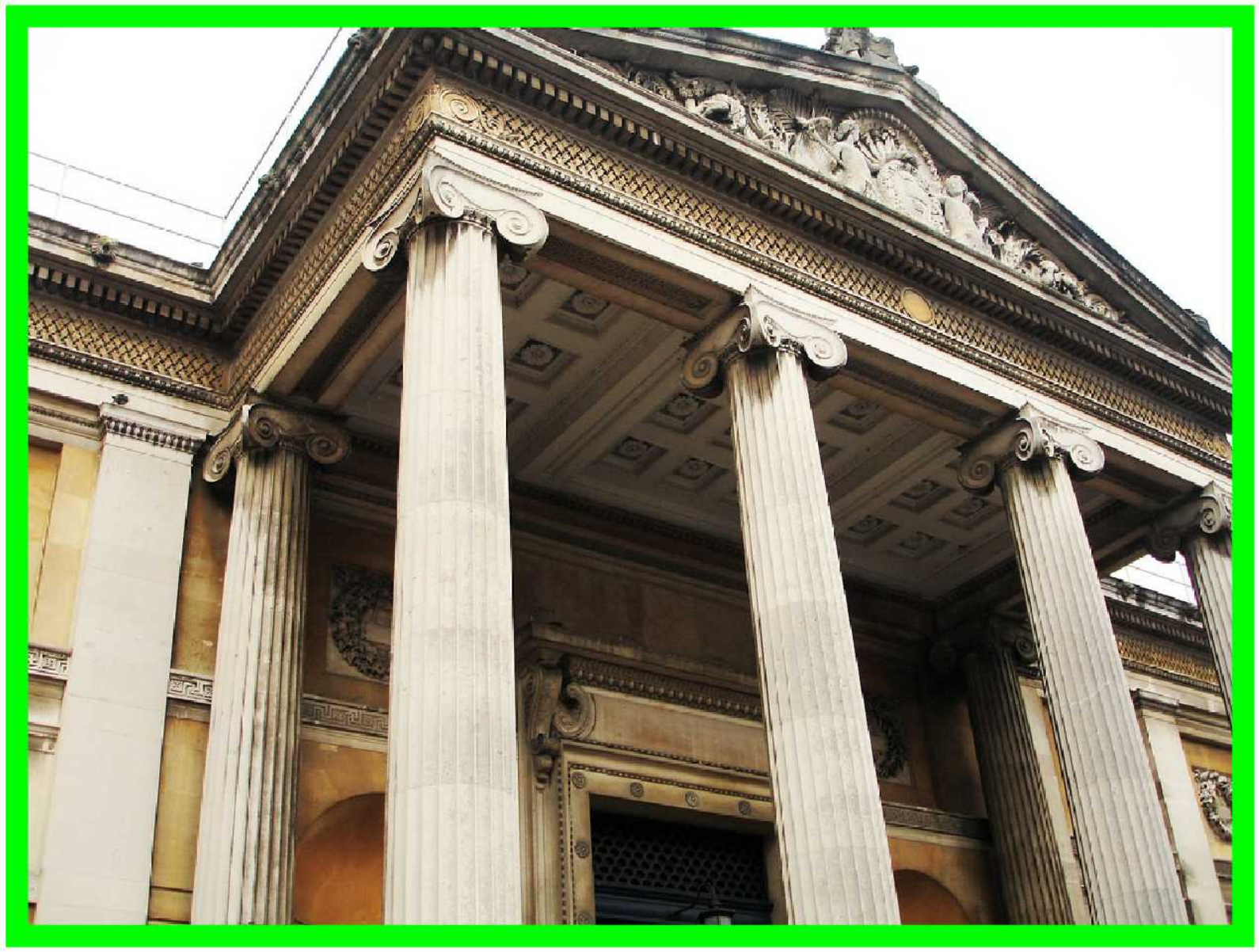}\end{tabular}
\begin{tabular}{@{\sssp}c@{\sssp}}\includegraphics[height=\figh]{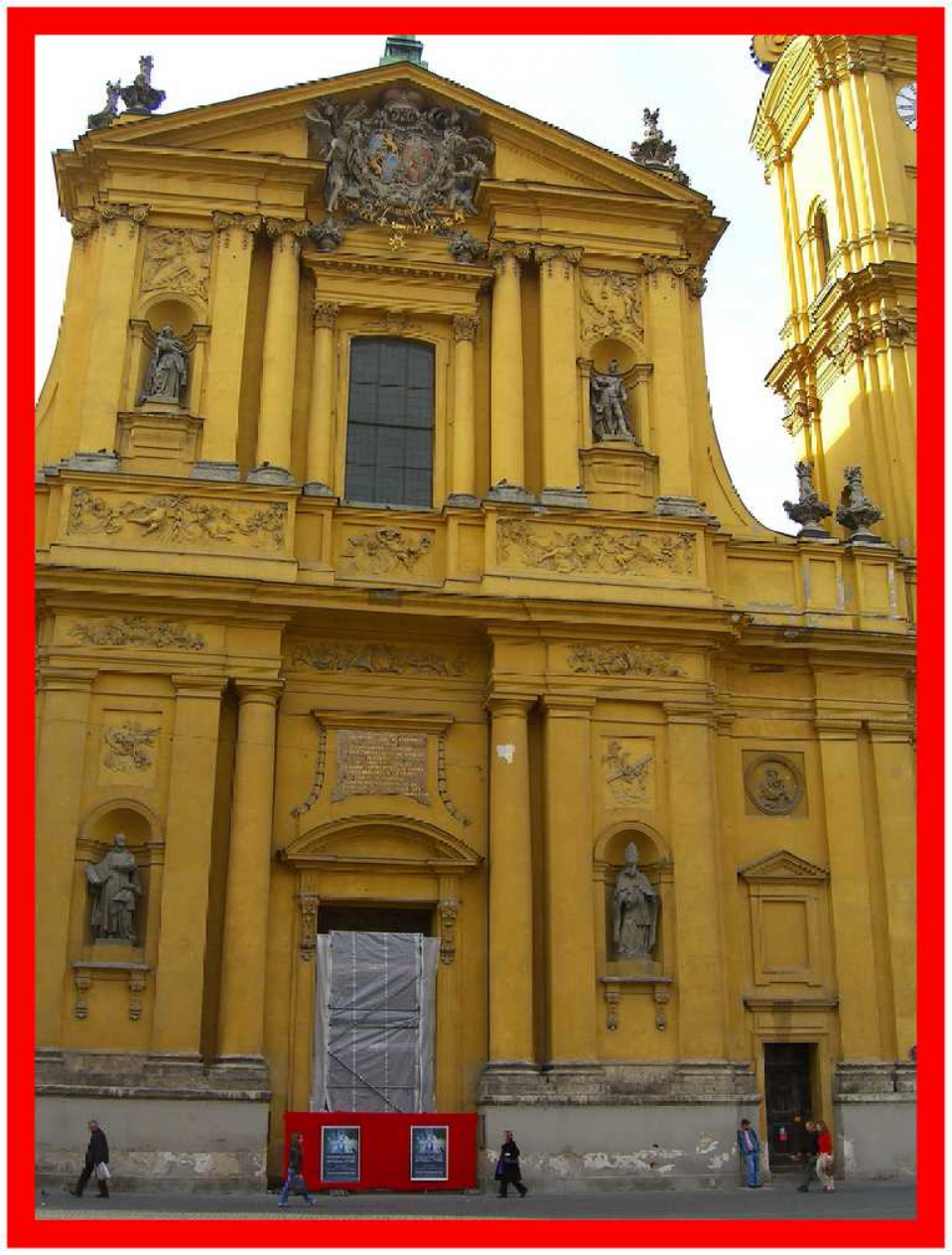}\end{tabular}
\begin{tabular}{@{\sssp}c@{\sssp}}\includegraphics[height=\figh]{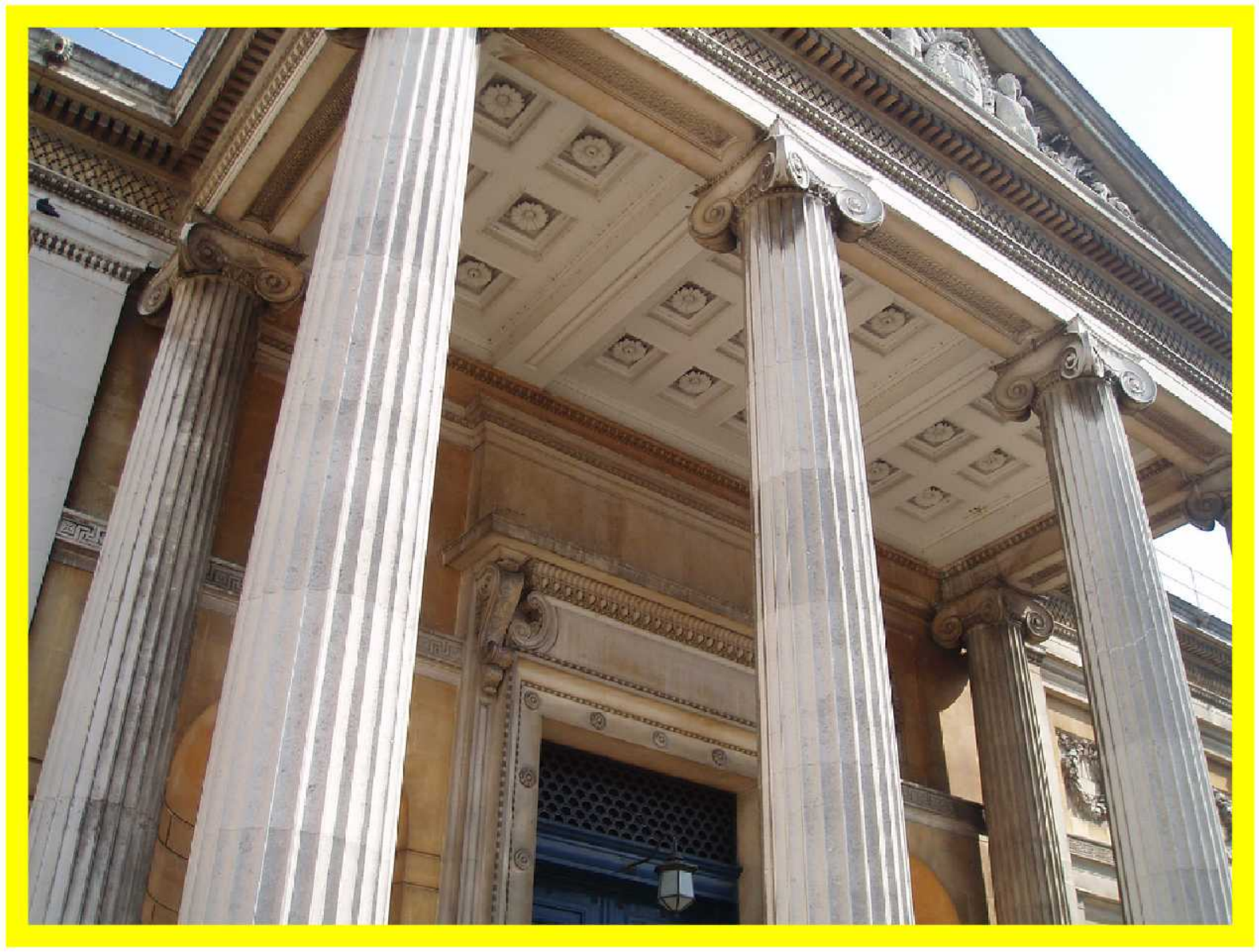}\end{tabular}

 \begin{tabular}{@{\sssp}c@{\sssp}}\includegraphics[height=\figh]{figs/rerank/9//q9_4081_5270.pdf}\\Query\\ \end{tabular} 
 \begin{tabular}{@{\sssp}c@{\sssp}}\includegraphics[height=\figh]{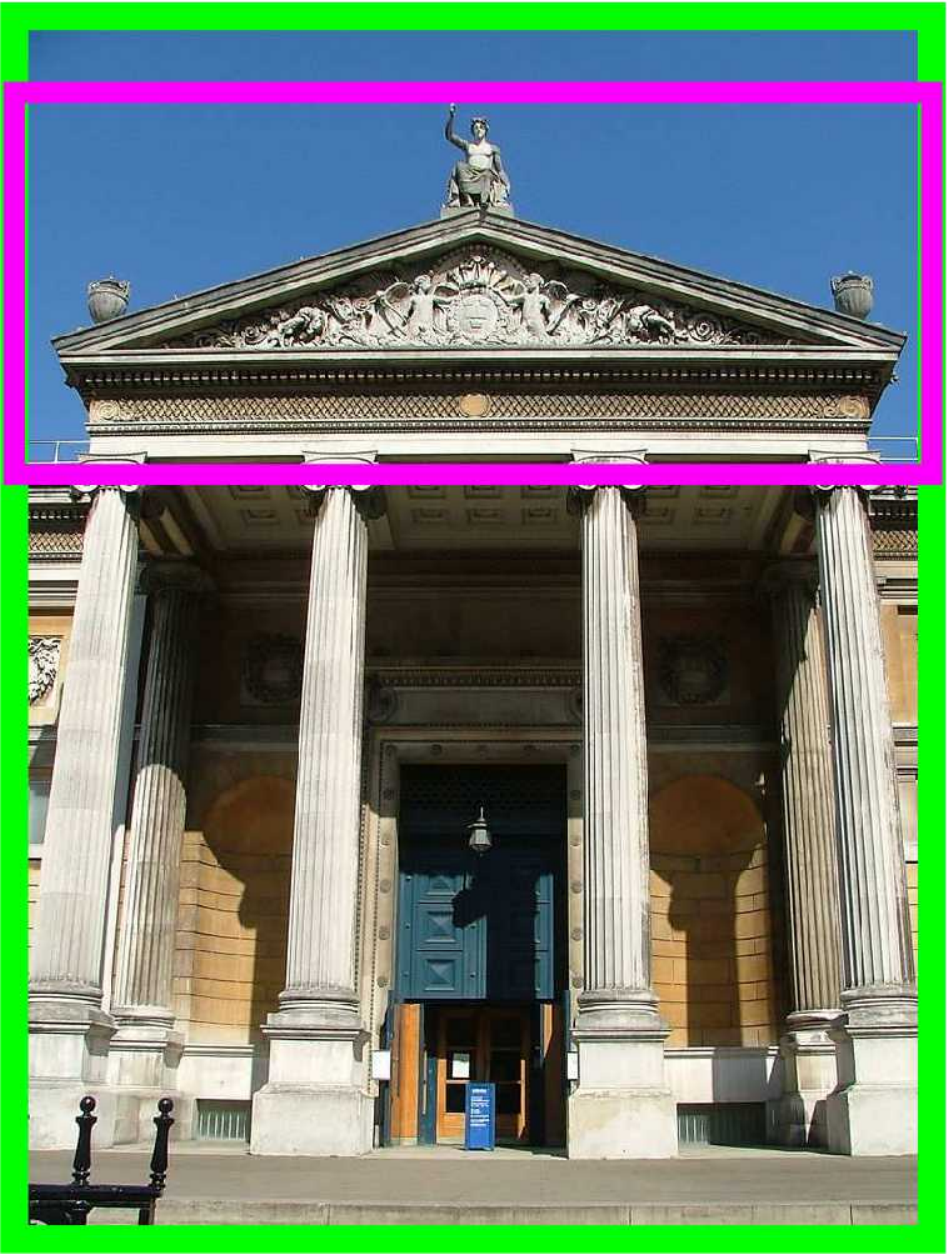}\\1 $\rightarrow$ 1\\ \end{tabular} 
 \begin{tabular}{@{\sssp}c@{\sssp}}\includegraphics[height=\figh]{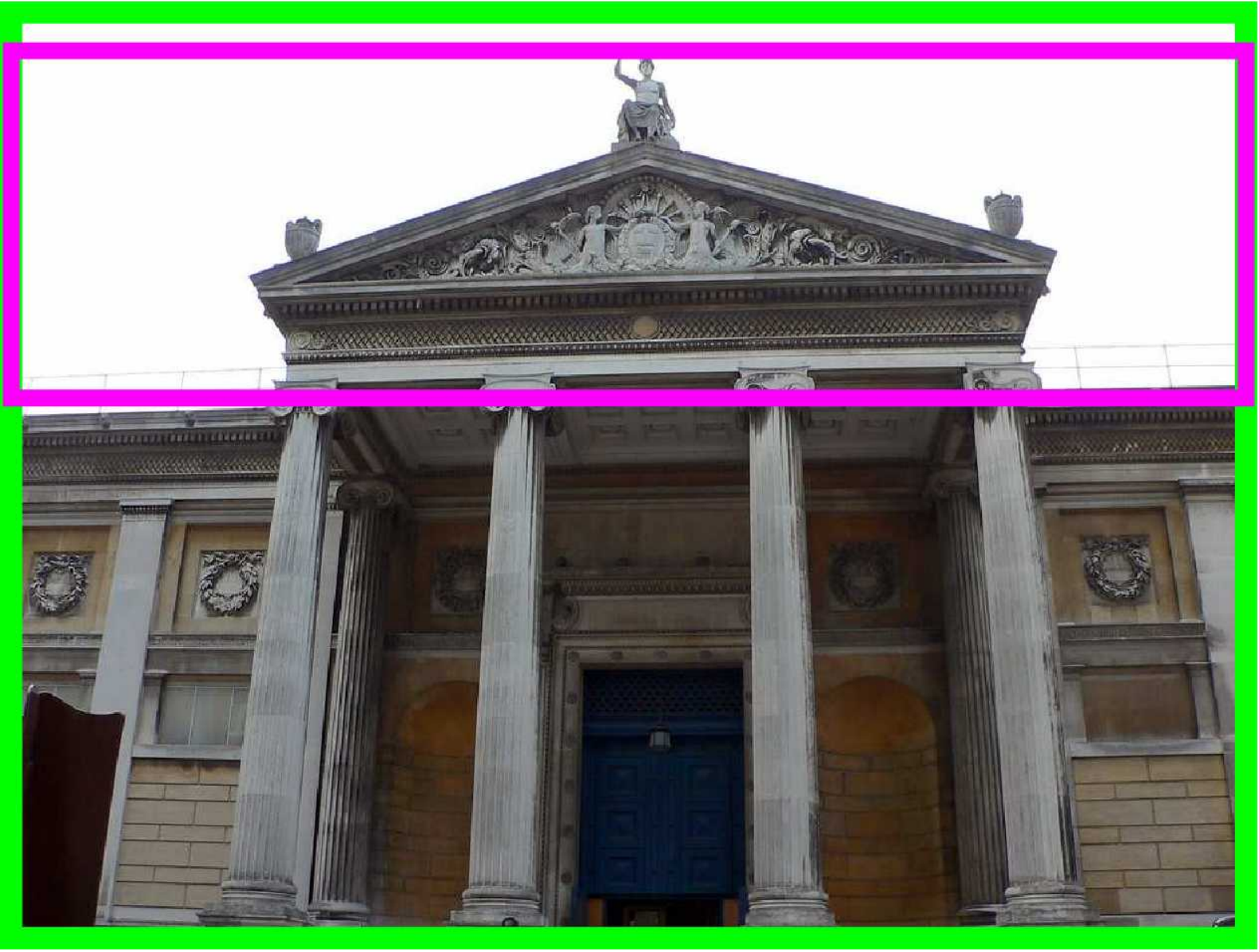}\\2 $\rightarrow$ 2\\ \end{tabular} 
 \begin{tabular}{@{\sssp}c@{\sssp}}\includegraphics[height=\figh]{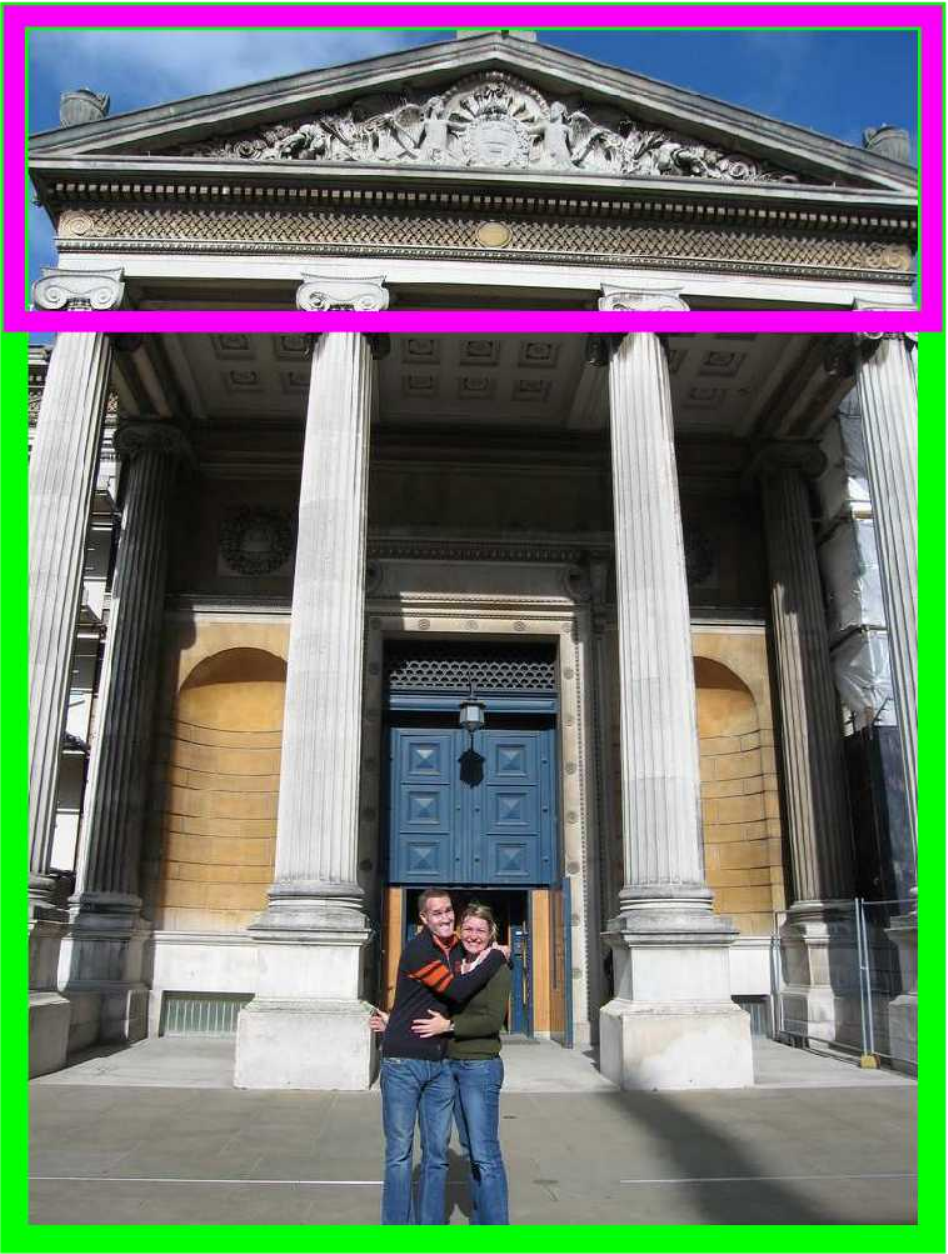}\\7 $\rightarrow$ 3\\ \end{tabular} 
 \begin{tabular}{@{\sssp}c@{\sssp}}\includegraphics[height=\figh]{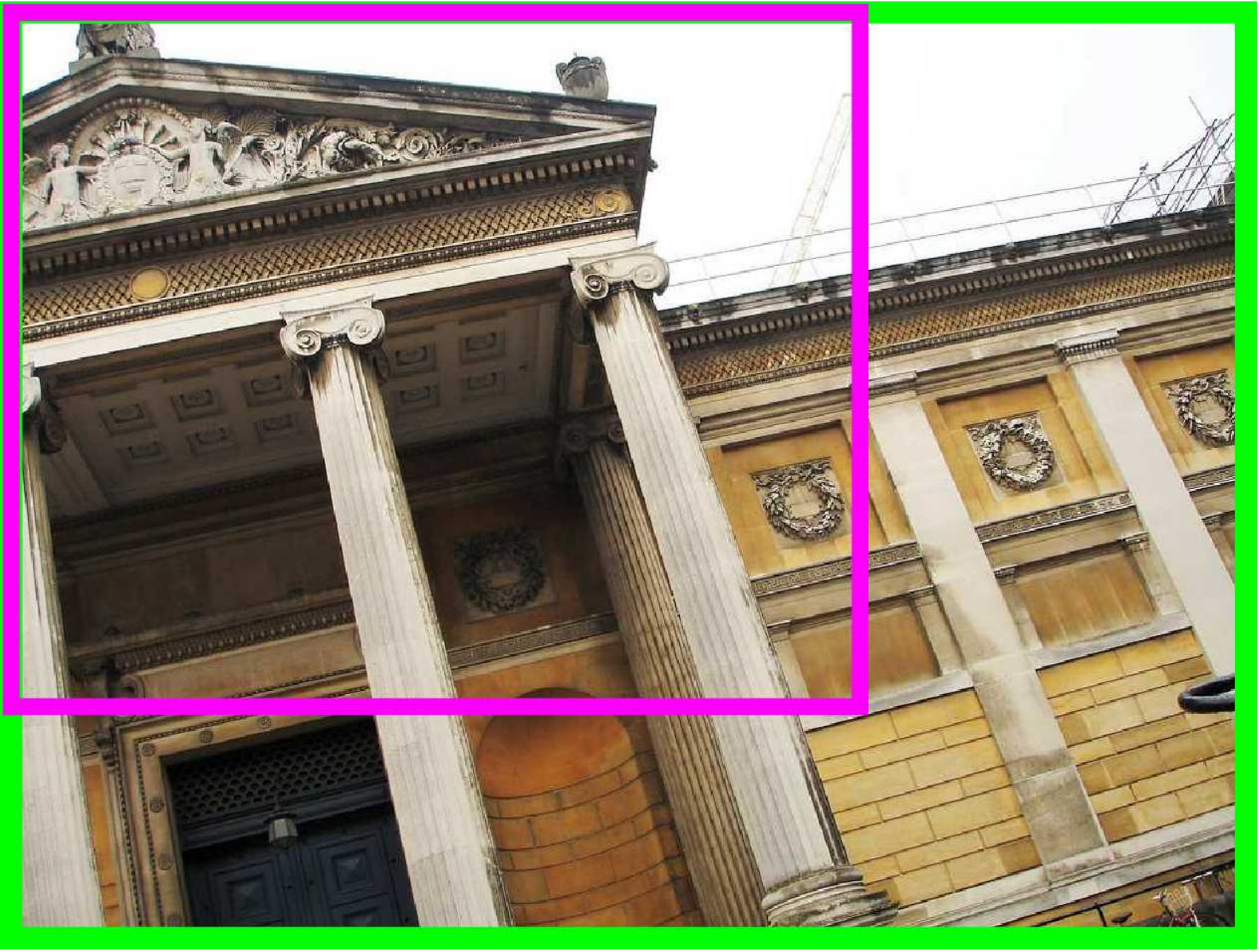}\\5 $\rightarrow$ 4\\ \end{tabular} 
 \begin{tabular}{@{\sssp}c@{\sssp}}\includegraphics[height=\figh]{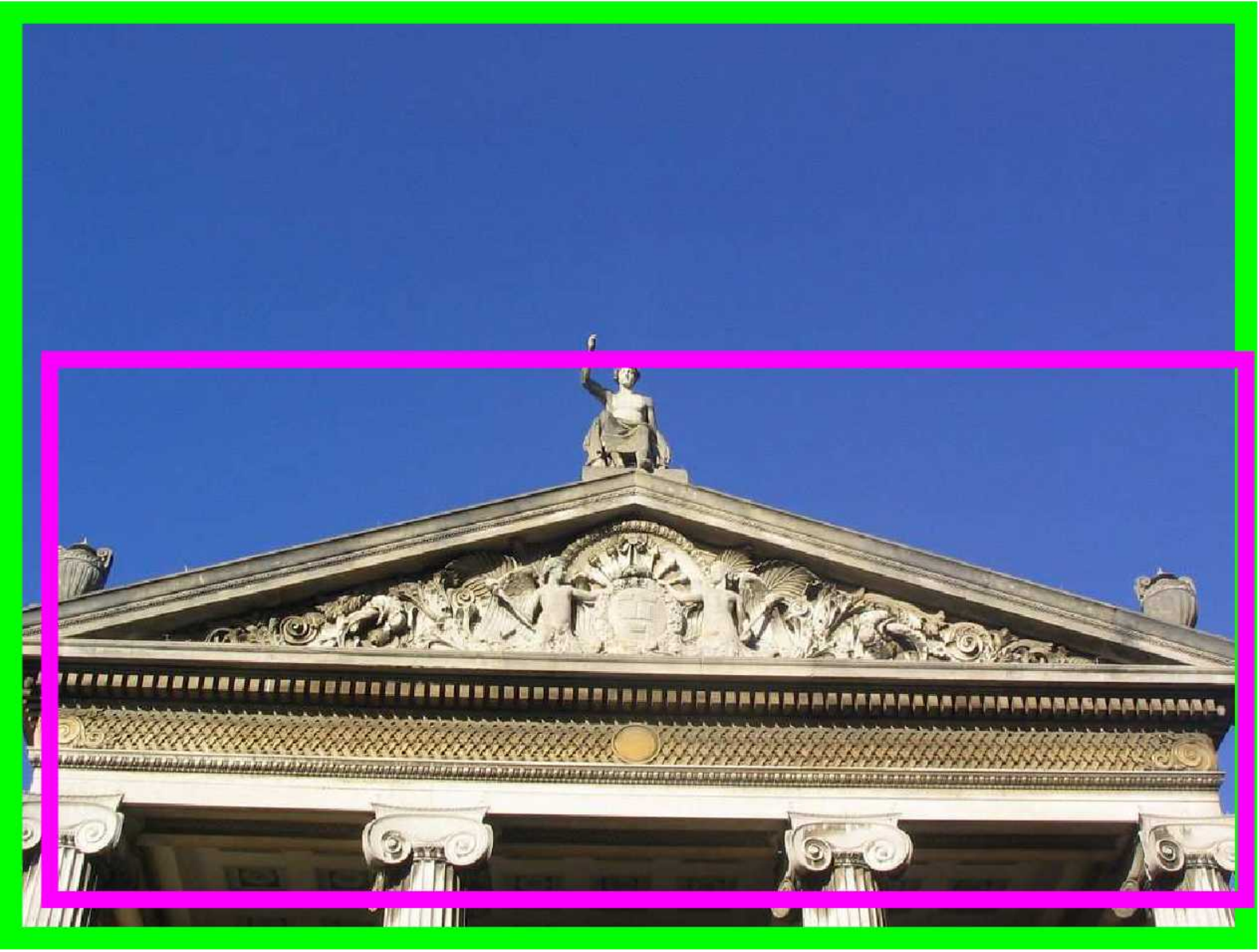}\\3 $\rightarrow$ 5\\ \end{tabular} 
 \begin{tabular}{@{\sssp}c@{\sssp}}\includegraphics[height=\figh]{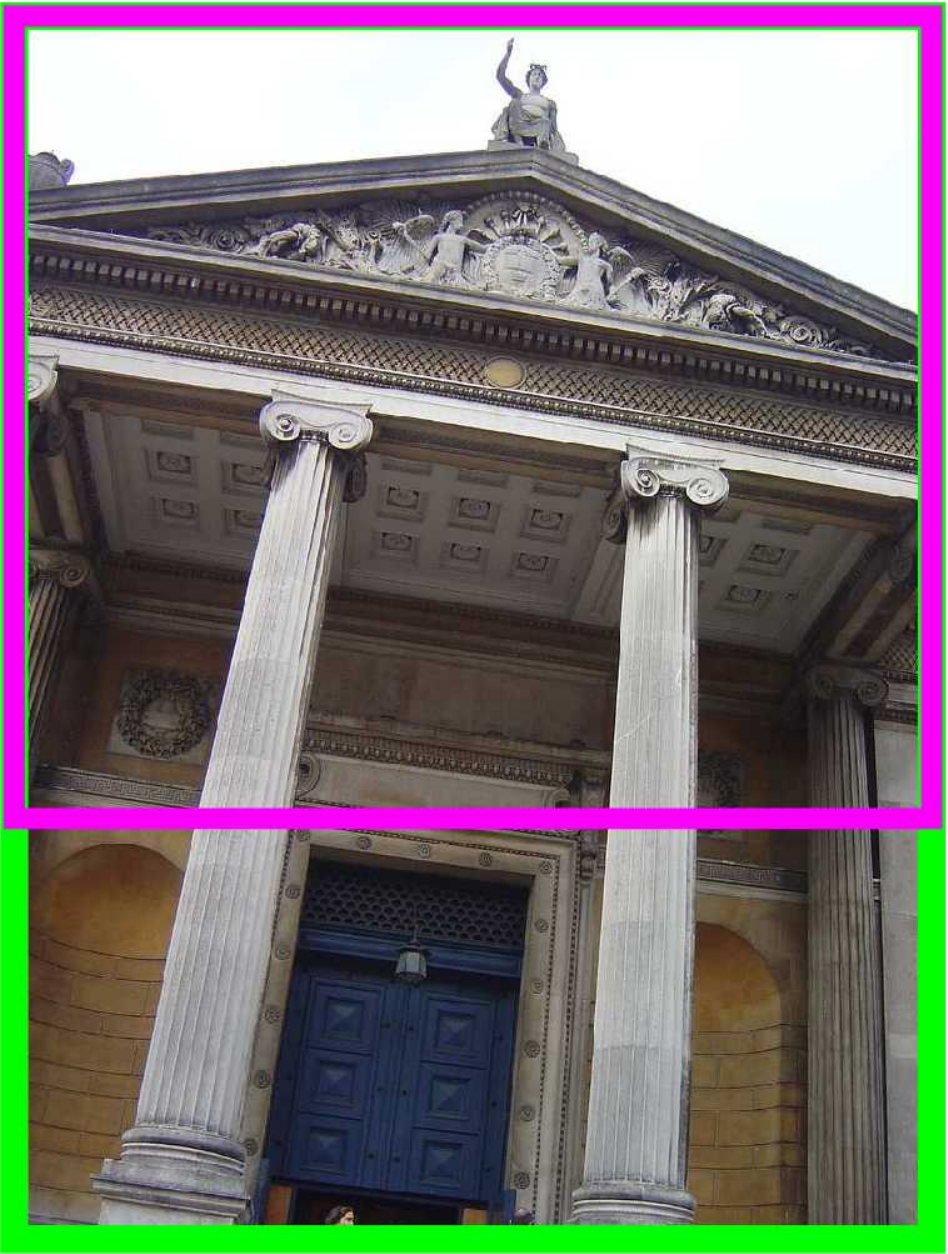}\\4 $\rightarrow$ 6\\ \end{tabular} 
 \begin{tabular}{@{\sssp}c@{\sssp}}\includegraphics[height=\figh]{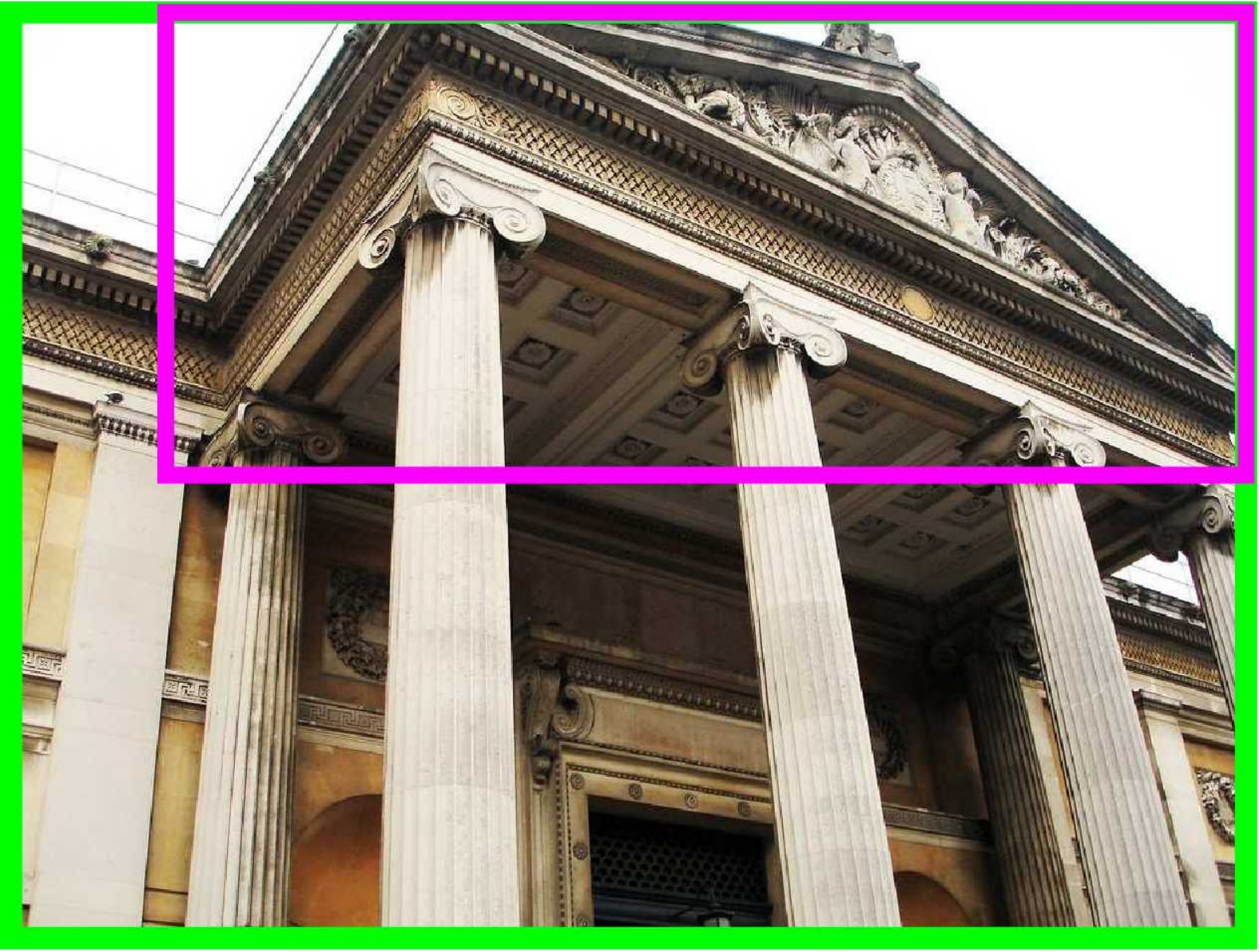}\\8 $\rightarrow$ 7\\ \end{tabular} 
 \begin{tabular}{@{\sssp}c@{\sssp}}\includegraphics[height=\figh]{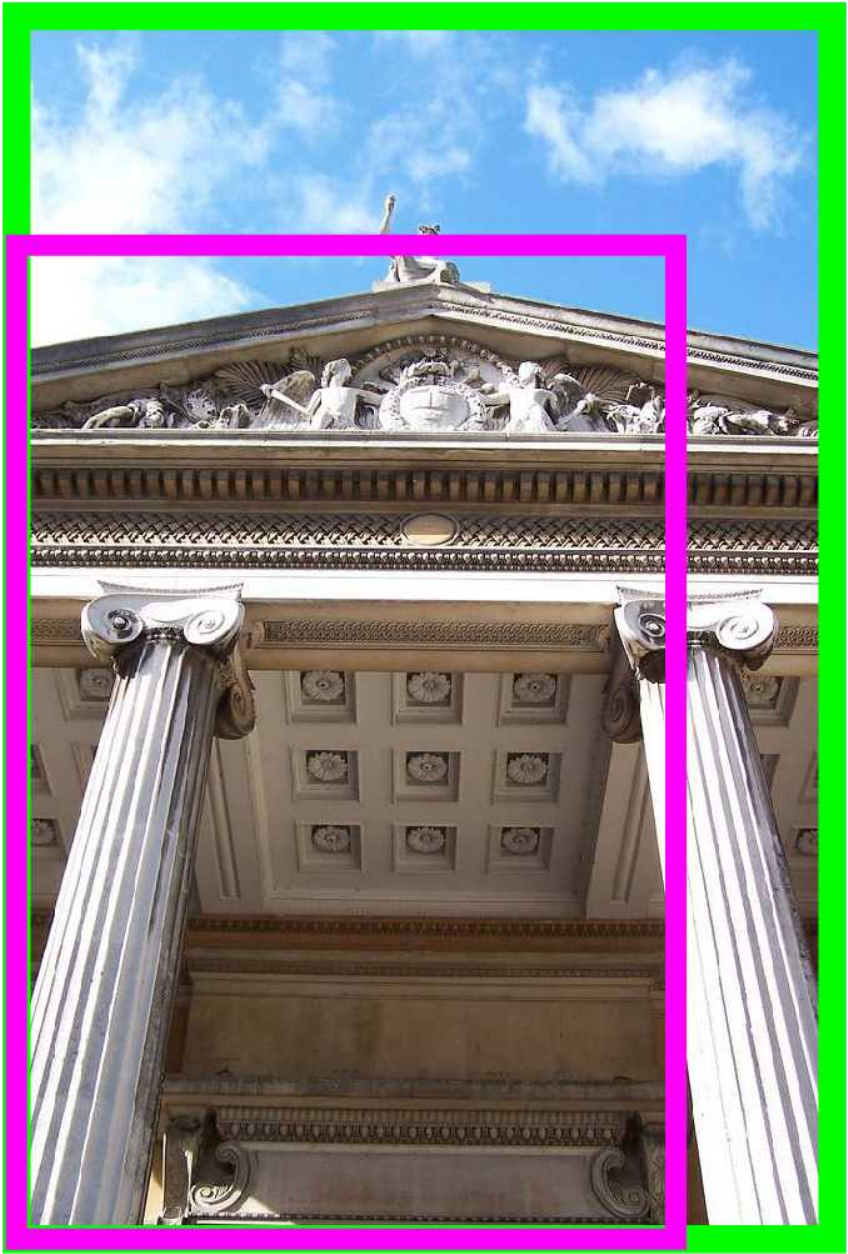}\\6 $\rightarrow$ 8\\ \end{tabular} 
 \begin{tabular}{@{\sssp}c@{\sssp}}\includegraphics[height=\figh]{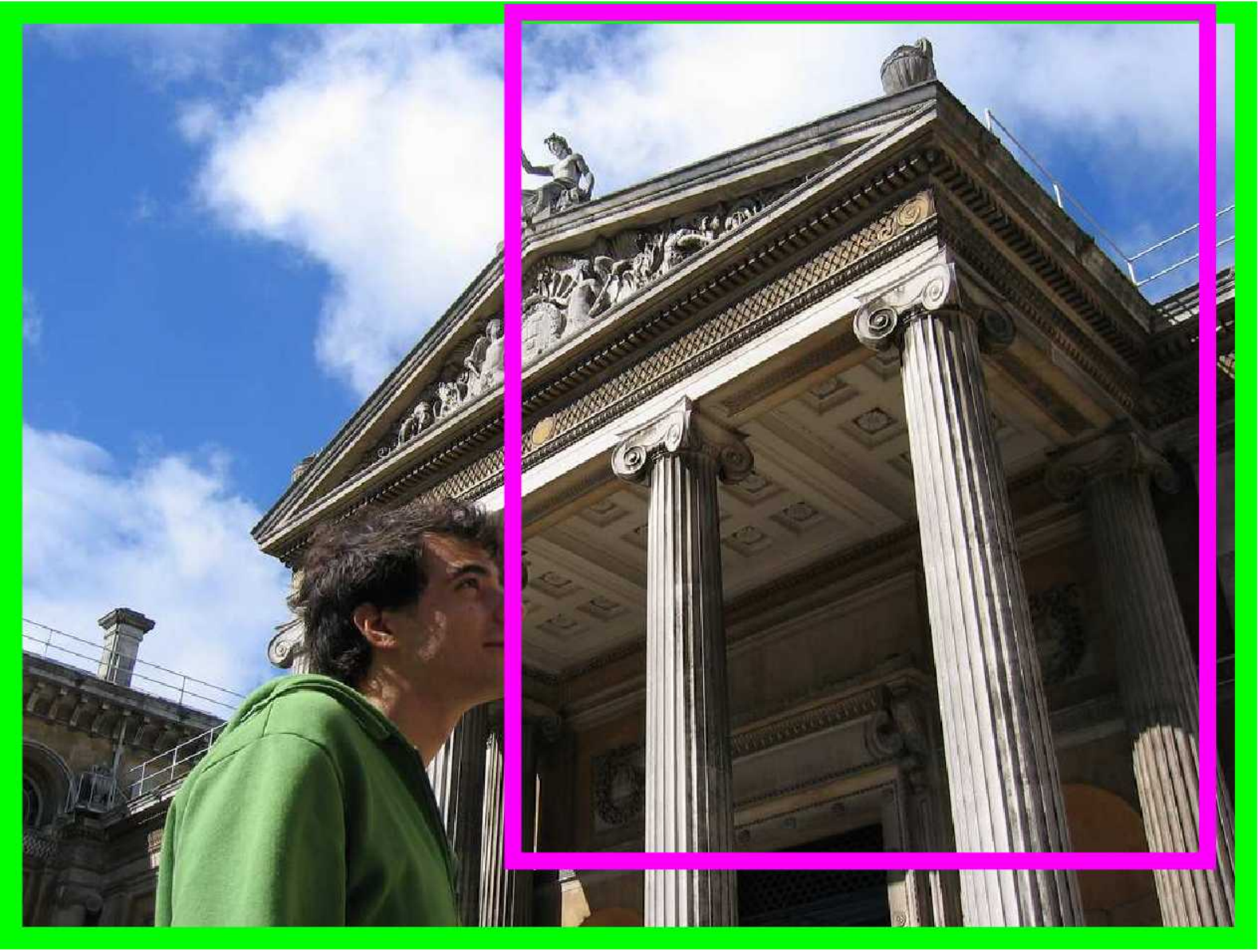}\\43 $\rightarrow$ 9\\ \end{tabular} 
 \begin{tabular}{@{\sssp}c@{\sssp}}\includegraphics[height=\figh]{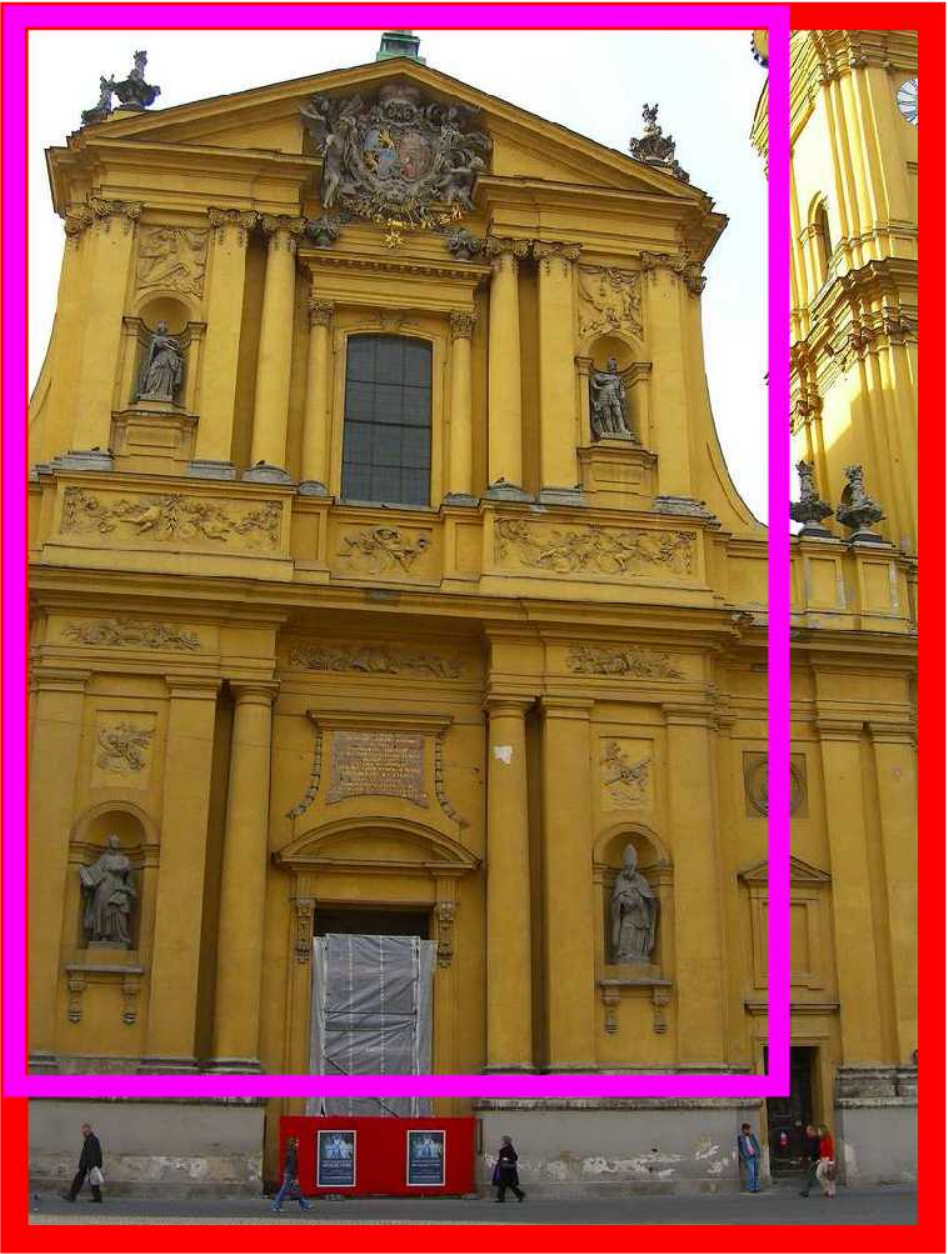}\\9 $\rightarrow$ 10\\ \end{tabular} 

\caption{Examples of top retrieved images before (top) and after (bottom) re-ranking with \deeploc. 
On the left we show the query image and depict the bounding box in blue color.
When re-ranking is used, we present the top ranked images and report for each image its initial and final ranking.
The localization window is shown in magenta, while positive/negative/junk images are depicted with green/red/yellow border.
\label{fig:rerankexample}}
\vspace{0pt}
\end{figure*}

Examples of ranking using \gfv and re-ranking using \deeploc are presented in Figure~\ref{fig:rerankexample}.
Recall that we only provide a rough object localization, since our main goal is to obtain improved image similarity.
Furthermore, the provided localization is accurate enough for re-ranking.
%

\begin{table}[t]
\caption{Performance comparison with state of the art. We report results for compact vector representations (left) and for retrieval approaches employing geometry, re-ranking, query expansion, or vector approximations (right). D = dimensionality.
Our approaches are identified with bullets $\bullet$.
\label{tab:soa}}
\vspace{2ex}
\footnotesize
\centering
\begin{tabular}{|@{\sssp}l@{\sssp}|@{\sssp}l@{\sssp}|@{\sssp}c@{\sssp}|@{\sssp}c@{\sssp}|@{\sssp}c@{\sssp}|@{\sssp}c@{\sssp}|@{\sssp}l@{\sssp}|@{\sssp}c@{\sssp}|@{\sssp}c@{\sssp}|@{\sssp}c@{\sssp}|@{\sssp}c@{\sssp}|} \hline
Method	       			   & D 	  & Oxf5k & Par6k & Oxf105k & Par106k &  Method	      	           & 		Oxf5k & 		Par6k & 		Oxf105k &  Par106k    \\ \hline \hline
{\scriptsize\cite{JZ14}}   & 1024 & 56.0  &	-    &  50.2    & -        & {\scriptsize\cite{CMPM11}} &       82.7   & 		80.5  &     76.7     &    71.0  	  \\
{\scriptsize\cite{JZ14}}   & 128  & 43.3  & -    &  35.3    & -        & {\scriptsize\cite{DGBQG11}}& 	   81.4   & 	    80.3  &		76.7    &  	  - 	      \\
{\scriptsize\cite{BSCL14}} & 128  & 55.7  & -    &  52.3    & -        & {\scriptsize\cite{MPCM13}} &\textbf{84.9} &	    82.4  &\textbf{79.5} &    77.3      \\
{\scriptsize\cite{RSMC14}} & 256  & 53.3  & 67.0 &  48.9    & -        & {\scriptsize\cite{SLBW14}} &        75.2  &      74.1  &     72.9     &     -        \\
{\scriptsize\cite{BL15}}   & 256  & 53.1  & -    &  50.1    & -        & {\scriptsize\cite{TGSS14}} &        77.8  &       -    &       -      &       -       \\
{\scriptsize\rfv} $\bullet$		   & 256  & 56.1  & 72.9 &  47.0    & 60.1     & {\scriptsize\cite{TAJ15}}  & 	    80.4  & 	    77.0  &		75.0     & 	  -	   	   \\
{\scriptsize\rfv} $\bullet$		   & 512  &\textbf{66.9}&\textbf{83.0}&\textbf{61.6} &\textbf{75.7} & {\scriptsize\rfv\hspace{-2pt}+\deeploc\hspace{-2pt}+QE} $\bullet$& 		77.3     &\textbf{86.5} &   73.2    &\textbf{79.8}  \\ \hline
\end{tabular}
\vspace{1ex}
\end{table}

\textbf{Comparison to the state of the art.}
We compare the proposed methods to state-of-the-art performance of compact representations and approaches based on local features that perform precise descriptor matching, re-ranking or query expansion.
Results are shown in Table~\ref{tab:soa}\footnote{Small differences of scores compared to the first version of the manuscript on arxiv are due to a slightly different evaluation protocol used before. Now, the evaluation protocol is the standard one for these datasets.}. AlexNex and VGG16 are used to produce the 256D and 512D vectors for \rfv, respectively.
Regarding the compact representations, our short-sized \rfv outperforms all other approaches.
The better performance on Paris is inherited by the nature of the pre-trained networks; the baseline \gfv with VGG achieves 55.2 on Oxford5k and 74.7 on Paris6k.

Unlike previous description schemes derived from CNN layers, our approach compete with the best
approaches based on local features for geometric matching and query expansion.
Our 
\deeploc can even outperform them: while our results are lower
on Oxford, we achieve the best performance on Paris and, to the best of our knowledge, outperform all published results on this benchmark.
Higher scores on Paris6k are reported by~\cite{AZ12} (91.0) and by~\cite{ZJS15} (91.5). These are achieved by learning the codebook on Paris6k itself and by performing pre-processing of the indexed dataset.

\textbf{Discussion about other CNN-based approaches.}
\cite{RSMC14} propose to perform region cross-matching and accumulate the maximum similarity per query region.
We evaluate this cross-matching process on the collection of regional vectors used in \rfv; we simply skip the final aggregation process and keep the regional vectors individually.
The cross-matching achieves 75.2\% mAP on Oxford5k as a filtering stage, while re-ranking with \deeploc on top of this acts in a complementary way and increases the performance up to 78.1\%. 
However, cross-matching has two drawbacks.
Firstly, the region vectors have to be stored individually and increase the memory requirements by a factor of $|R|$, where $|R|$ is the number of extracted regions.
Secondly, the complexity cost is linear in the number of indexed images and quite high since it requires to compute $|R|^2$ (\eg 1024~\citep{RSMC14}) inner products per image.
The work of~\cite{RSMC14} follows a non-standard evaluation protocol by enlarging the provided query bounding boxes.
In addition, the cost of their feature extraction is extremely high since they feed 32 images of resolution $576\times 576$ to the CNN.
The recent work of \cite{XTHZ15} is quite similar to theirs and is applied on both retrieval and classification.

\cite{BL15} show that global sum-pooling on convolutional layer activations is better than max-pooling when the final image vectors are PCA-whitened. 
When whitening is not employed, then the latter is better.
In the context of object localization we efficiently evaluate a large number of candidate regions on query time with \deeploc.
Performing whitening on each candidate region vector significantly increases the cost and is prohibitive for this task.
We switch max-pooling to sum-pooling for both our proposed \rfv and \deeploc and test performance. Note that sum-pooling is a special case of our integral max-pooling with $\alpha=1$.
Switching to sum-pooling makes \rfv perform $69.8$ and \rfv+\deeploc+QE perform $76.9$ on Paris106k. These scores are directly comparable to our scores in Table~\ref{tab:soa} and reveal that our choice is consistently better in all cases within our pipeline.

\section{Conclusions}
In this work, we re-visit both filtering and re-ranking retrieval stages by employing CNN activations of convolutional layers.
Our compact vector representation encodes several image regions with simple aggregation method and is shown to outperform state-of-the-art competitors.
Our localization increases the performance of the retrieval system that is initially based on a compact representation.
The same CNN information adopted during the filtering stage is employed for re-ranking as well.
Our approach competes with state-of-the-art methods that employ costly geometric matching or query expansion and we achieve the highest performance on Paris dataset, and provided a much better performance than existing approaches built upon CNN features. 
A very recent work \citep{AGTPS15} shows how \gfv performance is improved by end-to-end fine tunning where the objective is based on MAC similarity.

{
\bibliography{egbib}
\bibliographystyle{iclr2016_conference}
}

\end{document}